\pdfoutput=1

\documentclass[11pt]{article}

\usepackage{acl}

\usepackage[ruled,vlined,linesnumbered]{algorithm2e}

\usepackage{times}
\usepackage{latexsym}

\usepackage[T1]{fontenc}

\usepackage[utf8]{inputenc}

\usepackage{microtype}

\usepackage{inconsolata}

\usepackage{graphicx}

\usepackage{subcaption}
\usepackage{multirow}
\usepackage[inline]{enumitem}
\usepackage{amssymb}
\usepackage{amsthm}
\usepackage{amsmath} 
\usepackage{xcolor,colortbl}
\usepackage[capitalize,noabbrev]{cleveref}

\newtheorem{theorem}{Theorem}

\usepackage{mathtools}

\usepackage{mdframed}

\theoremstyle{plain}

\newmdtheoremenv[%
linecolor=gray,leftmargin=10,%
rightmargin=10,
backgroundcolor=gray!40,
nobreak=true,
]{myprop}{Template}[section]

\newcommand{\expsec}[1]{\vspace{+0.2cm} \noindent \textcolor{blue}{\underline{\textbf{#1}}}} 

\title{Explaining and Improving Contrastive Decoding \\ by Extrapolating the Probabilities of a Huge and Hypothetical LM}

\definecolor{colorloss3}{RGB}{203, 41, 123}
\definecolor{colorloss2}{RGB}{29, 177, 0}
\definecolor{colorloss1}{RGB}{0, 162, 255}

\author{
  Haw-Shiuan Chang\textsuperscript{1}\thanks{The work was mostly done at Amazon.} \, \, Nanyun Peng\textsuperscript{2} \, \, Mohit Bansal\textsuperscript{2}  \\ \textbf{Anil Ramakrishna\textsuperscript{2}} \, \, \textbf{Tagyoung Chung\textsuperscript{2}} \\
  \textsuperscript{1}UMass Amherst CICS \, \, \textsuperscript{2}Amazon AGI Foundations \\
  \texttt{ hschang@cs.umass.edu, \{pengnany,mobansal,aniramak,tagyoung\}@amazon.com} 
}

\begin{document}
\maketitle
\begin{abstract}

Contrastive decoding (CD)~\citep{li2022contrastive} improves the next-token distribution of a large expert language model (LM) using a small amateur LM. Although CD is applied to various LMs and domains to enhance open-ended text generation, it is still unclear why CD often works well, when it could fail, and how we can make it better. To deepen our understanding of CD, we first theoretically prove that CD could be viewed as linearly extrapolating the next-token logits from a huge and hypothetical LM.
We also highlight that the linear extrapolation could make CD unable to output the most obvious answers that have already been assigned high probabilities by the amateur LM.

To overcome CD’s limitation, we propose a new unsupervised decoding method called \textbf{A}symptotic \textbf{P}robability \textbf{D}ecoding (APD).\footnote{The code will be released at \url{https://github.com/amazon-science/llm-asymptotic-decoding}} APD explicitly extrapolates the probability curves from the LMs of different sizes to infer the asymptotic probabilities from an infinitely large LM without inducing more inference costs than CD. In \textsc{FactualityPrompts}, an open-ended text generation benchmark, sampling using APD significantly boosts factuality in comparison to the CD sampling and its variants, and achieves state-of-the-art results for Pythia 6.9B and OPT 6.7B. Furthermore, in five commonsense QA datasets, APD is often significantly better than CD and achieves a similar effect of using a larger LLM. For example, the perplexity of APD on top of Pythia 6.9B is even lower than the perplexity of Pythia 12B in CommonsenseQA and LAMBADA.

\end{abstract}

\section{Introduction}

\begin{figure}[t!]
\centering
\includegraphics[width=1\linewidth]{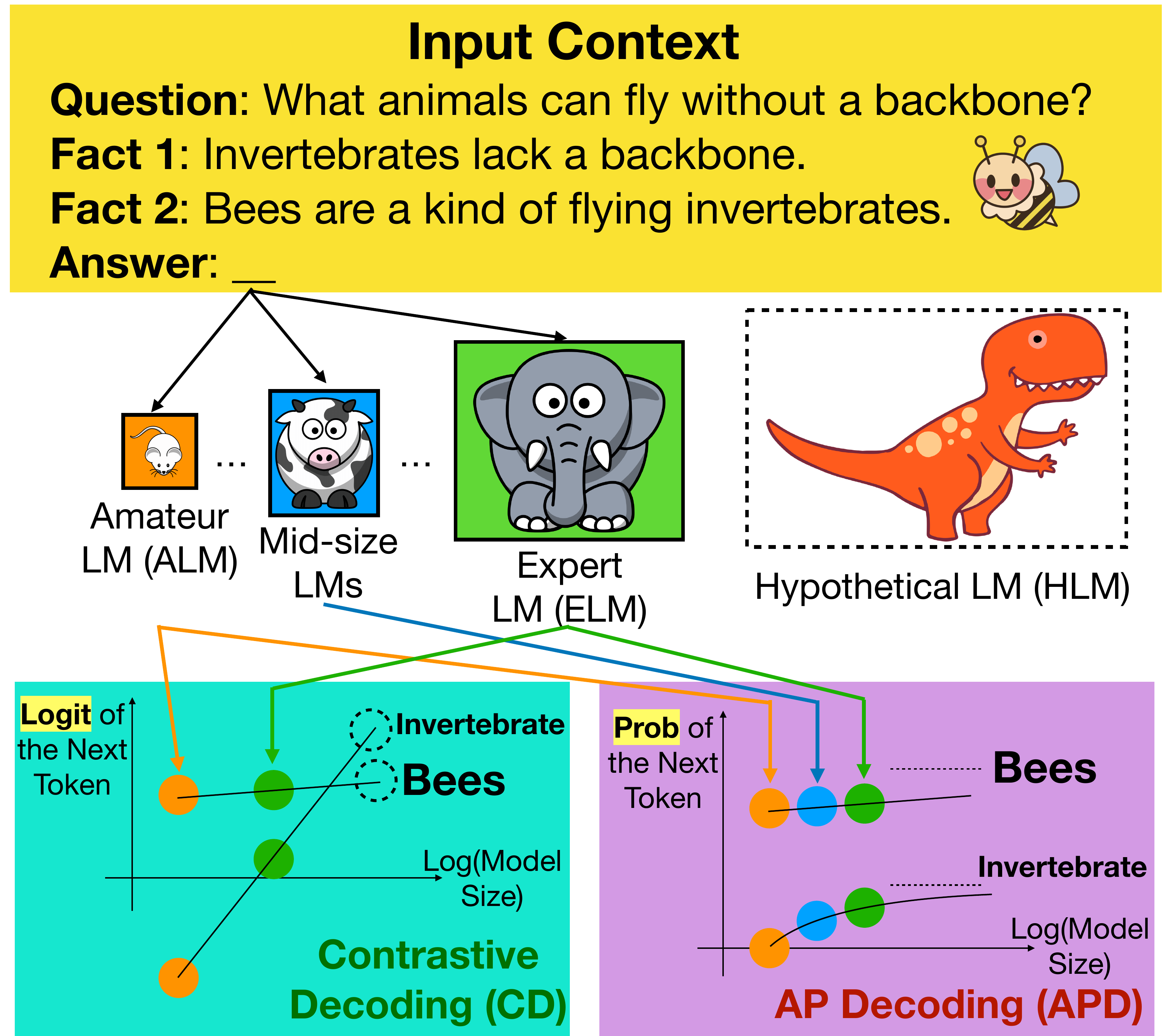}
\caption{Given a simple question with clues for which a tiny amateur LM could provide a correct answer, contrastive decoding (CD) could have a ``obvious blindness'' (i.e., assigning a higher logit to an uncommon answer \emph{Invertebrate} than the most obvious answer \emph{Bees}). In contrast, the proposed asymptotic probability decoding (APD) correctly assigns the highest probability to \emph{Bees} by leveraging the probabilities from multiple LMs of different sizes to extrapolate the probabilities from an infinitely large and hypothetical LM.}
\label{fig:first_fig}
\end{figure}



Contrastive Decoding~\citep{li2022contrastive} (CD) is a simple heuristic that uses the logit of a small LM (amateur LM) to improve the logit of a large LM (expert LM).\footnote{Since the amateur LM is easier to fail, CD adjusts the logit of an expert LM by subtracting the logit of an amateur LM, so a lower amateur's logit implies a higher output logit of CD. Using \Cref{fig:first_fig} as an example, CD produces a large logit for \textit{Invertebrates} because its amateur's logit is low. }
The potential of CD has been demonstrated in various open-ended text generation tasks~\citep{li2022contrastive} and reasoning tasks using the expert LMs up to 65B~\citep{o2023contrastive}. Several variants are also proposed to reduce toxicity~\citep{liu2021dexperts}, improve factuality in NLP tasks~\citep{chuang2023dola,zhang2023alleviating,shi2023trusting,sanchezstay}, text evaluation~\citep{lu2023open}, and vision tasks~\citep{wan2024contrastive}. However, due to the insufficient theoretical understanding of CD, it is difficult to identify and overcome the failure modes of CD, which hinders its wide applications. 

Scaling law demonstrates that language models (LMs) are able to generate more factual next tokens as their sizes increase~\citep{kaplan2020scaling,lee2022factuality}. However, their growing energy consumption and costs limit their further applications~\citep{strubell2019energy,lacoste2019quantifying,kaack2022aligning}, necessitating techniques that can reduce LM model sizes without compromising their superior performances.
In this work, we theoretically demonstrate how contrastive decoding (CD) addresses this LM size-reduction challenge using a simple linear extrapolation. The theory also helps us to identify the limitations of CD and propose APD, a more factual decoding method.

%




First, we discover that CD actually uses the tiny amateur LM to help the large expert LM infer the logit of a huge and hypothetical LM. Specifically, the logits from CD could often be viewed as a linear extrapolation of the logit curves from the expert LM and amateur LM.
The finding explains several prior empirical observations and also reveals weaknesses of CD. For example, CD tends to neglect the most obvious answer and overemphasize less likely answers in its output distribution instead. We call this tendency ``obvious blindness''. The rare answers could sometimes degrade the generation's factuality. 
For example, both amateur and expert LM in \Cref{fig:first_fig} can identify that \textit{Bees} is the most clear answer suggested by the clues but only the expert LM realizes that there are also some other possible answers such as \textit{Invertebrates}. Then, the aggressive linear extrapolation of CD makes \textit{Invertebrates} become the most probable next token. This is not a totally factual answer because many invertebrates cannot fly.

Motivated by this theoretical explanation, we propose a novel decoding method called Asymptotic Probability Decoding (APD). APD predicts the asymptotic probability from a hypothetical LM with an infinite size. By explicitly modeling the changes of the next-token probabilities as the size of LM increases, APD is able to output the correct probabilities for both easy/common and difficult/uncommon answers.  For the example in \Cref{fig:first_fig}, the probabilities of \textit{Bees} and \textit{Invertebrates} are both increasing as the LM becomes larger. By leveraging the probabilities of mid-size LMs, we can reasonably infer that \textit{Bees} should still receive a larger probability from a huge LM than \textit{Invertebrates}. Finally, modeling the probability curves for many next tokens on the fly is too time-consuming, so we fine-tune an amateur LM such that the output probability of APD is close to asymptotic probability, which makes APD as efficient as CD.







The main goal of our experiments is to check if APD can further improve the factuality compared to CD. We choose our expert LMs and amateur LMs from the LLM families that provide smaller LMs, including Pythia (6.9B, 70M)~\citep{biderman2023pythia}, OPT (6.7B, OPT 125M)~\citep{zhang2022opt}, and Qwen1.5 (4B, 0.5B)~\citep{qwen}. By comparing different sampling methods in \textsc{FactualityPrompts}~\citep{lee2022factuality}, we demonstrate that APD consistently and robustly outperforms CD with the best temperature and other state-of-the-art distribution modification methods such as DoLa~\citep{chuang2023dola} and temperature sampling~\citep{ficler2017controlling}. After being combined with dynamically adjusted top-p sampling~\citep{REAL}, our method can help Pythia 6.9B to simultaneously achieve the factuality of top-p sampling~\citep{holtzman2019curious} with $p=0.4$ and diversity of top-p with $p=0.7$.

We also compare the perplexity of APD and CD using seven datasets. We found that the improvement gap is especially large when CD makes more mistakes on easier tasks. For example, in LAMBADA~\citep{paperno2016lambada} and CommonsenseQA~\citep{talmor-etal-2019-commonsenseqa}, APD on top of Pythia 6.9B could achieve a similar or better perplexity than Pythia 12B, while outperforming CD by a large margin. We plan to release our code to reproduce the results after our work is accepted.

Our main contributions include
\begin{itemize}[leftmargin=.1in,topsep=1pt]
\setlength\itemsep{-0.1em}
    \item We provide theoretical support for contrastive decoding (CD) and demonstrate that our theory can explain many prior findings from \citet{li2022contrastive, o2023contrastive}.
    \item We propose a new distribution modification method, asymptotic probability decoding (APD), which addresses the ``obvious blindness'' of CD.
    \item We conduct extensive experiments, which indicate that APD could significantly improve the generation factuality of CD.
\end{itemize}

\section{Contrastive Decoding as Extrapolation}

First, we review contrastive decoding (CD) and its justification. In CD, the probability of the next token $w$ for context $c$ is determined by 
\begin{equation}
P^{CD}_c(w) = \frac{ \exp( L^{CD}_c(w) )}{ \sum_{x} \exp( L^{CD}_c(x) ) }, 
\label{eq:CD_prob}
\end{equation}
where $x$ is all the words in the vocabulary, and the logit for the token $w$ is
\begin{equation}
L^{CD}_c(w) = L^{ELM}_c(w) - \frac{1}{T} L^{ALM}_c(w), 
\label{eq:LCD}
\end{equation}
where $L^{ELM}_c$ and $L^{ALM}_c$ are the logit of the expert LM (ELM) and amateur LM (ALM), respectively. $T$ is the softmax temperature of ALM. In the original paper~\citep{li2022contrastive}, the effectiveness of CD is mostly explained by removing the bad distribution of ALM from the ELM. The explanation is correct, but not specific enough to explain many prior empirical findings in \citet{li2022contrastive, o2023contrastive}. 

Here, we provide another intuitive justification of \Cref{eq:LCD}: ELM and ALM usually come from the same LLM family and they are trained on the same training corpus using similar hyperparameters, so their main difference is the model sizes. $L^{CD}_c(w)$ would be smaller when $L^{ELM}_c(w)$ is small and $L^{ALM}_c(w)$ is large. In this case, it means that the LMs' logits are decreasing for the token $w$ as the LM's model size increases. We can reasonably infer that the larger LM will output an even smaller logit if the trend continues. Thus, CD actually infers the logit of a larger LM by subtracting $L^{ALM}_c(w)$ from $L^{ELM}_c(w)$. We describe this intuition in a formal way next.





\begin{figure}[t!]
\centering
\includegraphics[width=1\linewidth]{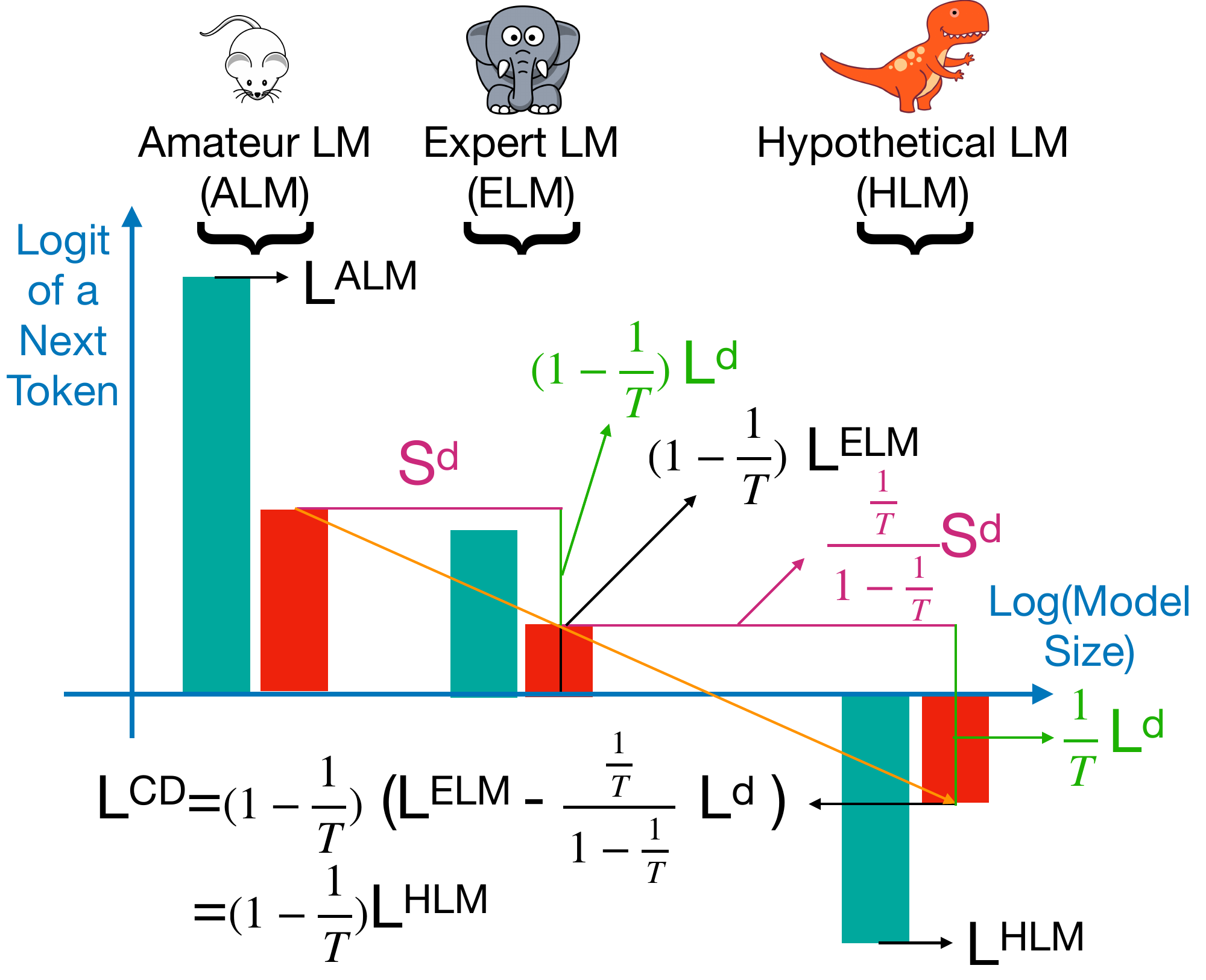}
\caption{Illustration of our proof for \Cref{thm:main_theorem}. Teal bars are original logits, and red bars are the logits scaled by $1-\frac{1}{T}$. $L^d = L^{ALM} - L^{ELM}$, $S^d$ is the size difference of ELM and ALM in a logarithm space. We drop the word $w$ and the context $c$ in the notations of this figure for simplicity.}
\label{fig:theory}
\end{figure}

\subsection{Theoretical Analysis}

\begin{theorem}
\label{thm:main_theorem}
If \\  
a) the ALM's temperature $T>1$, and \\ 
b) the logits of LMs and the logarithm of the LM sizes have a linear relationship, then \\ 
the logit of contrastive decoding (CD) for the token $w$ $L^{CD}_c(w)= (1-\frac{1}{T} )L^{HLM}_c(w)$, where $L^{HLM}_c(w)$ is the logit of a LM with size $s^{HLM} = \left(\frac{ (s^{ELM})^T}{s^{ALM}}\right)^{ \frac{1}{T-1}} $.
\end{theorem}

Setting $T>1$ for ALM is effective in various tasks~\citep{o2023contrastive}, so our first assumption often holds in practice. The linear relationship means we can always draw a line to connect all the logits from LMs with different sizes as in \Cref{fig:theory}. Since $T$ is a global hyperparameter, $L^{CD}_c(w)$ is the logit of HLM using the temperature $\frac{T}{T-1}$.




\begin{proof}

\vspace{-3pt}

\footnotesize
\begin{align}
& L^{CD}_c(w) = L^{ELM}_c(w) - \frac{1}{T} L^{ALM}_c(w) \nonumber \\[-1pt]
& = (1-\frac{1}{T} ) L^{ELM}_c(w) + \frac{1}{T} (L^{ELM}_c(w) - L^{ALM}_c(w)) \nonumber \\[-4pt]
& = (1-\frac{1}{T} ) ( L^{ELM}_c(w) - \frac{ \frac{1}{T} }{1-\frac{1}{T}} L^{d}_c(w) )  
\end{align}

\normalsize

\vspace{-6pt}

From \Cref{fig:theory}, we can see that the size difference between HLM and ELM should be $\frac{ \frac{1}{T} }{1-\frac{1}{T}} S^d$.

\vspace{-10pt}

\footnotesize
\begin{align}
& \log s^{HLM} = \log s^{ELM} + \frac{ \frac{1}{T} }{1-\frac{1}{T}} S^d \nonumber \\[-5pt]
& = \log s^{ELM} + \frac{1}{T-1} (\log s^{ELM} - \log s^{ALM}) \nonumber \\[-4pt]
& = \frac{T}{T-1} \log s^{ELM} - \frac{1}{T-1} \log s^{ALM} \nonumber \\[-1pt]
& = \frac{1}{T-1}\log ( \frac{ (s^{ELM})^T}{s^{ALM}} ) 
\end{align}

\vspace{-3pt}

\normalsize

\end{proof}





\subsection{Implications of the Theorem}

Our theory explains several prior findings and provides insights into CD's limitations.

\expsec{Why does CD generally work well?}
\citet{li2022contrastive, o2023contrastive} show the empirical success of CD across various domains and LMs. From the perspective of \Cref{thm:main_theorem}, the source of effectiveness relies on the validity of the linear extrapolation for the individual token logits.

\expsec{Why is the best temperature task-dependent?}
The different downstream applications often have different optimal temperatures $T$ of ALM in the CD~\citep{o2023contrastive}. We hypothesize that the linear relationship assumption in \Cref{thm:main_theorem} is often violated in some downstream tasks. For example, in the commonsense task of \Cref{fig:first_fig}, the LLM with the size of the HLM should not have a large logit for the word \textit{Invertebrates}. Thus, it could be more appropriate to use a larger $T$ for these tasks, which makes the size of HLM closer to the ELM and thus reduces the aggressiveness of the extrapolation.





\expsec{Why does CD prefer a large size difference?}
The experiments in ~\citet{li2022contrastive,o2023contrastive} show that CD usually works better when the size difference between ALM and ELM is larger. \Cref{fig:theory} provides one explanation: Given the same $T$, a larger size difference $S_d$ increases the size of HLM; the observations on a longer x-axis range could often support further extrapolation. Moreover, a smaller ALM is less likely to answer the simple questions correctly, so the problem in \Cref{fig:first_fig} is less likely to happen. 



\expsec{Why does CD use LMs from the same family?}
\citet{li2022contrastive,o2023contrastive} choose to use the smallest LM from the same LLM family as the ALM. Our theory supports the choice and suggests that when ALM and ELM are trained on different corpora, CD would reverse the tendency or bias of ALM. For example, if ALM is more much likely to mention \emph{Biden} than \emph{Trump} compared to ELM, CD would output more \emph{Trump} than \emph{Biden}, which might lead to the unfactual sentence like \emph{``In 2020-2024, the USA president is Trump''}.




\expsec{When could CD fail?}
The simplicity of linear extrapolation in the logit space is both CD's advantage and limitation. If $L^{ALM}_c(w)$ is very small in \Cref{eq:LCD}, $L^{CD}_c(w)$ might become very large even if $L^{ELM}_c(w)$ is still not large. For example, in \Cref{fig:first_fig}, the ALM's logit for \textit{Invertebrate} might be very small because it is a rare word in general. The truncation/thresholding methods such as top-p or $\alpha$-masking~\citep{li2022contrastive} filter out the rare tokens with small $L^{ELM}_c(w)$, which alleviates, but does not completely solve the problem. After all, if the thresholding method filters out too many next tokens, the output would have a very low diversity. This motivates us to propose a more sophisticated extrapolation method to further improve CD.



\begin{figure}[t!]
\centering
\includegraphics[width=1\linewidth]{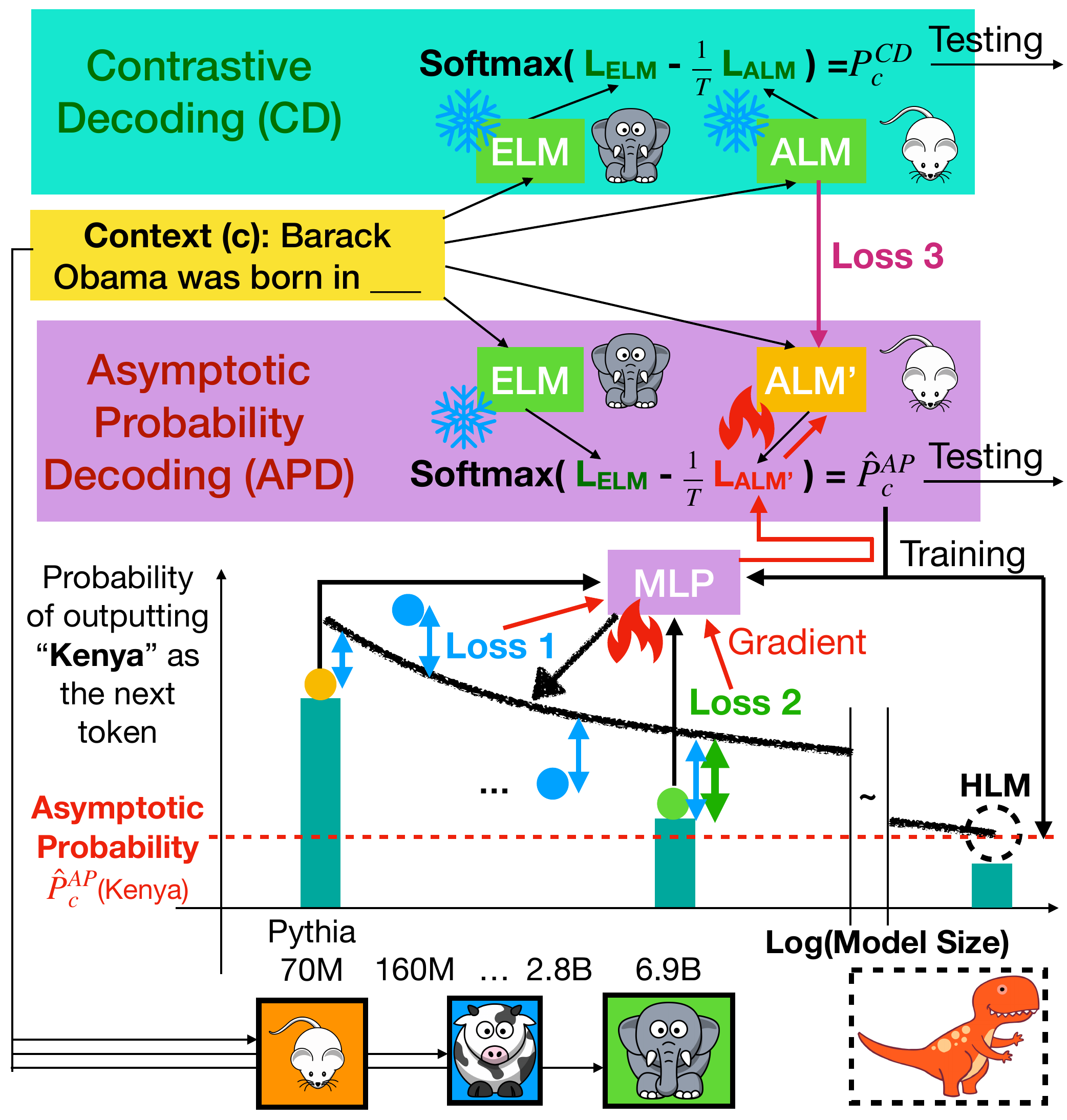}
\caption{Fine-tuning ALM to predict the asymptotic probability ($P_c^{AP}$). During the training time, the predicted $\hat{P}_c^{AP}(\text{Kenya})$ and the empirical probabilities from the LLM family $\{p(w|c,\theta_{s_i})\}_{i=1}^N$ are inputted into an MLP. If $\hat{P}_c^{AP}(\text{Kenya})$ is too high to model the empirical probabilities well, the probability curve outputted by MLP would be far away from the empirical probabilities and thus, incur a high \textcolor{colorloss1}{loss 1} and \textcolor{colorloss2}{loss 2}. Then, the resulting gradients would be backpropagated through MLP and ALM and reduce $\hat{P}_c^{AP}(\text{Kenya})$. Finally, we add a regularization \textcolor{colorloss3}{loss 3} to control the changes in the ALM's logit output. }
\label{fig:APD}
\end{figure}




\section{Asymptotic Probability Decoding}

We propose asymptotic probability decoding (APD) to overcome the limitations of CD. In an LLM family, there are often LMs with sizes between the sizes of the amateur LM (ALM) and expert LM (ELM). As shown in \Cref{fig:first_fig}, APD leverages them to improve the extrapolation and models the probability curves instead of the logit curves to avoid outputting a large probability based on a very small logit from ALM (e.g., \textit{Invertebrates}). Furthermore, for those easy answers (e.g., \textit{Bees}), $L^{ELM}_c(w)$ and $L^{ALM}_c(w)$ are both high. APD can extrapolate the high probability curve to output a desired high probability while based on \Cref{eq:LCD}, the CD might output a low logit.

To make APD as efficient as CD\footnote{It demands too much extra computational overhead during inference to run a series of LM with different sizes and conducting extrapolation for each possible next token.}, APD requires us to fine-tune the ALM in CD such that the resulting output probabilities would be close to the asymptotic probabilities (AP) from an HLM with an infinite size. However, we cannot get the ground truth AP to supervise our model directly, so during training on an unsupervised text corpus, we first collect the predicted AP of the top likely next tokens and the empirical probabilities from the LMs with different sizes. Next, as illustrated in \Cref{fig:APD}, we use an MLP energy network~\citep{lecun2006tutorial,belanger2017end} to output a curve that is close to the observed probabilities and would approach the predicted AP as LM's size goes to infinity. After the training, the APD uses the updated amateur LM (ALM') in the same way as the CD to produce the next-token distribution: 


\vspace{-10pt}
\footnotesize
\begin{equation}
\hat{P}^{AP}_c(w) = \frac{ \exp( L^{ELM}_c(w) - \frac{1}{T} L^{ALM'}_c(w) )}{ \sum_x \exp( L^{ELM}_c(x) - \frac{1}{T} L^{ALM'}_c(x) ) }. 
\label{eq:AP}
\end{equation}
\normalsize
We provide more details of our procedure of training ALM' in the follow subsections.







\subsection{Training Setup}
Given that we have a LLM family containing $M$ models with sizes $s_1, ..., s_M$ in a logarithmic scale, we use $\{\theta_{s_1}, \theta_{s_2}, ... \theta_{s_M} \}$ to represent the (parameters of) LLMs.  
The largest LLM that can fit into our GPUs is $\theta_{s_N}$, so we use $\theta_{s_N}$ as our ELM and $\theta_{s_1}$ as our ALM.



On our training text corpus, we run the LLMs in the family to collect their probabilities $\{p(w|c,\theta_{s_i})\}_{i=1}^N$. Storing the probabilities of all possible tokens is not feasible, so for each context $c$, we select a set of tokens $A_c$, including the $20$ tokens with the highest ELM probabilities and some randomly sampled tokens with smaller probabilities. We normalize the probabilities so that their summation within $A_c$ is $1$, and only keep their normalized probabilities for training. After all, most tokens with small ELM probabilities are often truncated during inference time by thresholding methods such as top-$p$ sampling.

To reduce the training cost and avoid overfitting the training data, we choose to only update ALM. One simple but suboptimal approach is to first train a curve prediction model that extrapolates the AP using $\{p(w|c,\theta_{s_i})\}_{i=1}^N$ and then train ALM to encourage the APD to output AP. Instead, we propose to merge the two training stages together and jointly optimize the ALM and curve prediction model in the next subsections.








\subsection{Curve Parameterization}

To model both increasing or decreasing curves using the same parameterization way, we propose a preprocessing step $R(\cdot)$ that flips the probabilities if the probabilities increase as the model size increases. Specifically, we compute 

\vspace{-7pt}
\footnotesize
\begin{align}
& \{\hat{P'}^{AP}_{c}(w)\} \cup \{p'(w|c,\theta_{s_i})\}_{i=1}^N = R(Q) \nonumber \\
& =\begin{cases}
 & Q \;\;\;\; \text{ if } p(w|c,\theta_{s_1}) \geq p(w|c,\theta_{s_N}) \\ 
 & \{ 1-p | p \in Q\} \;\;\;\; \text{ o.w } 
\end{cases}
,
\label{eq:flip}
\end{align}
\normalsize
where $Q=\{\hat{P}^{AP}_{c}(w)\} \cup \{p(w|c,\theta_{s_i})\}_{i=1}^N$ and $\hat{P}^{AP}_{c}(w)$ is the APD's output from ELM and ALM' using \Cref{eq:AP} and $T=1$.

After flipping, we use a simple exponential function to model the decay trends from $\{p'(w|c,\theta_{s_i})\}_{i=1}^N$: 


\vspace{-8pt}
\footnotesize
\begin{equation}
     \hat{p}_{w,c}(s) = \hat{P'}^{AP}_{c}(w) + a_{w,c} e^{-\max(0, b_{w,c} (s-d_{w,c}) ) }, 
\label{eq:exp}
\end{equation}
\normalsize
where $\hat{p}_{w,c}(s)$ is the predicted probability given the model size $s$ in a logarithm scale, $a_{w,c}$, $b_{w,c}$, $d_{w,c}$ are all positive parameters, and $\hat{P'}^{AP}_{c}(w)$ is the (flipped) output probability from \Cref{eq:flip}. Besides, $\hat{P'}^{AP}_{c}(w)$ is also an AP because $\lim_{s \to \infty } \hat{p}_{w,c}(s) = \hat{P'}^{AP}_{c}(w)$.

We choose a simple feedforward neural network, a 4-layer MLP (multilayer perceptron), for modeling the probability curve for each token $w$. The MLP takes the empirical probabilities and the predicted AP as the inputs, and outputs the parameters of the probability curves:

\vspace{-8pt}
\footnotesize
\begin{equation}
a_{w,c}, b_{w,c}, d_{w,c} = \text{MLP}\left(\hat{P'}^{AP}_{c}(w),\{p'(w|c,\theta_{s_i})\}_{i=1}^N \right) .
\label{eq:MLP}
\end{equation}
\normalsize



\begin{algorithm}[!t]

\small

\SetAlgoLined
\SetAlCapHSkip{0.2em} 
\setlength{\algomargin}{0em}

\SetKwInOut{Input}{Input}
\SetKwInOut{Output}{Output}
\SetKwProg{Fn}{Function}{:}{}
\SetKwFunction{FMain}{Fitting}

\SetInd{0.5em}{0.5em}

\Input{LLMs ($\{\theta_{s_i}\}_{i=1}^N$), including ALM (Amateur LM) and ELM (Expert LM), and Training Corpus $D$}
\Output{ALM’}

Compute $\{p(w|c,\theta_{s_i})\}_{i=1}^N$ (Probabilities of $N$ LLMs in $D$), $L^{ALM}_c(w)$, and $L^{ELM}_c(w)$ (Logits of ALM and ELM in $D$) \\
Initialize ALM’ $\leftarrow$ ALM \\
\ForEach{batch $B$ in $D$}{%
    \ForEach{context $c$ in $B$}{%
	\ForEach{$w$ in $A_c$ (top tokens of ELM) }{%
	   Predict asymptotic probability $\hat{P}^{AP}_c(w)$ using ALM’ and \Cref{eq:AP} \\
          $\{\hat{p}_{w,c}(s_i)\}_{i=1}^N$ $\leftarrow$ \FMain{\text{MLP}, $\hat{P}^{AP}_c(w)$, $\{p(w|c,\theta_{s_i})\}_{i=1}^N$ } \tcp{Function below}
        }
    }
    Compare $\{p(w|c,\theta_{s_i})\}_{i=1}^N$ with $\{\hat{p}_{w,c}(s_i)\}_{i=1}^N$ using \Cref{eq:loss1} and \eqref{eq:loss2} \\ 
    Regularize ALM’ using ALM and \Cref{eq:loss3} \\
    Update ALM’ and MLP to minimize the loss in \Cref{eq:loss_all} using backpropogation

}

\Fn{ \scriptsize \FMain{\text{MLP}, $\hat{P}^{AP}_c(w)$, $\{p(w|c,\theta_{s_i})\}_{i=1}^N$ }}{
          Reverse the increasing $\{p(w|c,\theta_{s_i})\}_{i=1}^N$ and $\hat{P}^{AP}_c(w)$ using \Cref{eq:flip} \\
          Predict decay curve parameters using the (reversed) probabilities, MLP, and \Cref{eq:MLP} \\
          Predict $\{\hat{p}_{w,c}(s_i)\}_{i=1}^N$ of $N$ LLMs using the curve parameters, $\hat{P'}^{AP}_{c}(w)$, and \Cref{eq:exp}\\
          \KwRet $\{\hat{p}_{w,c}(s_i)\}_{i=1}^N$
    
}

 \caption{Fine-tuning ALM'}
 \label{algo:training}
\end{algorithm}

\subsection{Loss Functions}
Our first loss computes the square root of the mean squared error (MSE) between the probability curves $\hat{p}_{w,c}(s)$ and the (flipped) empirical probability observations $\{p'(w|c,\theta_{s_i})\}_{i=1}^{N-1}$:

\vspace{-8pt}
\scriptsize
\begin{equation}
L_1 = \sqrt{ \frac{1}{Z \cdot (N-1)} \sum_{c \in B} \sum_{w \in A_c} \sum_{i=1}^{N-1} \left(p'(w|c,\theta_{s_i}) - \hat{p}_{w,c}(s_i) \right)^2 },
\label{eq:loss1}
\end{equation}
\normalsize
where $B$ is the training batch, $A_c$ is top token candidates of ELM, the normalization term $Z = |B| |A_c|$. Notice that, unlike a typical regression model, the goal of MLP is not predicting $\{p'(w|c,\theta_{s_i})\}_{i=1}^{N-1}$, which has been seen in its input. Instead, the MLP could be viewed as an energy network that checks the quality of APD's output $\hat{P}^{AP}_{c}(w)$. If $\hat{P}^{AP}_{c}(w)$ is not good, MLP cannot output a good curve $\hat{p}_{w,c}(s)$ that is close to all the empirical probabilities $\{p'(w|c,\theta_{s_i})\}_{i=1}^{N-1}$.

We found that only using the MSE loss often leads to an overestimation of $\hat{P}^{AP}_{c}(w)$ from a decay curve as illustrated in \Cref{fig:APD}. When the probabilities are decreasing, the AP should be smaller than the probability of ELM $p'(w|c,\theta_{s_N})$. Thus, we impose the second loss that pushes down the curve when its predicted probability of ELM is higher than the ground truth:

\vspace{-8pt}
\footnotesize
\begin{equation}
L_2 = \sqrt{ \frac{1}{Z} \sum_{c \in B} \sum_{w \in A_c}  \max\left(0, \hat{p}_{w,c}(s_N) - p'(w|c,\theta_{s_N}) \right) }.
\label{eq:loss2}
\end{equation}
\normalsize
We also use the square root here to emphasize the small probability differences.

Finally, we add a regularization term to control the logit changes of the ALM'. 

\vspace{-8pt}
\footnotesize
\begin{equation}
L_3 = \sqrt{ \frac{1}{Z} \sum_{c \in B} \sum_{w \in A_c}  \left(L^{ALM'}_c(w) - L^{ALM}_c(w) \right)^2 }.
\label{eq:loss3}
\end{equation}
\normalsize
Without the regularization, we found APD tends to output a trivial solution, where $L^{ALM'}_c(w) = 0$ and $\hat{P}^{AP}_c(w) = p'(w|c,\theta_{s_N})$. 

We combine all the terms as our final loss function for training ALM' and MLP 

\footnotesize
\begin{equation}
Loss = L_1 + \lambda_2 L_2 + \lambda_3 L_3
\label{eq:loss_all}
\end{equation}
\normalsize
We fix $\lambda_2$ to be a large number $10$. Different LM families have different logit value ranges, so their optimal $\lambda_3$ are different. We fix $\lambda_3$ to be $0.8$ for Pythia, $0.4$ for OPT, and $1.0$ for Qwen in all our experiments except for the ablation study.

In \Cref{fig:APD}, we show that the gradients would flow through MLP and ALM' to make APD output the better asymptotic probability. Our training algorithm for $1$ epoch is summarized at \Cref{algo:training}. During testing, we just replace ALM in CD with ALM' without running MLP. That is, APD can conduct a more sophisticated extrapolation than CD without increasing its inference cost. 




\begin{figure*}[t!]
\centering
\begin{subfigure}{.33\textwidth}
  \centering
  \includegraphics[width=1\linewidth]{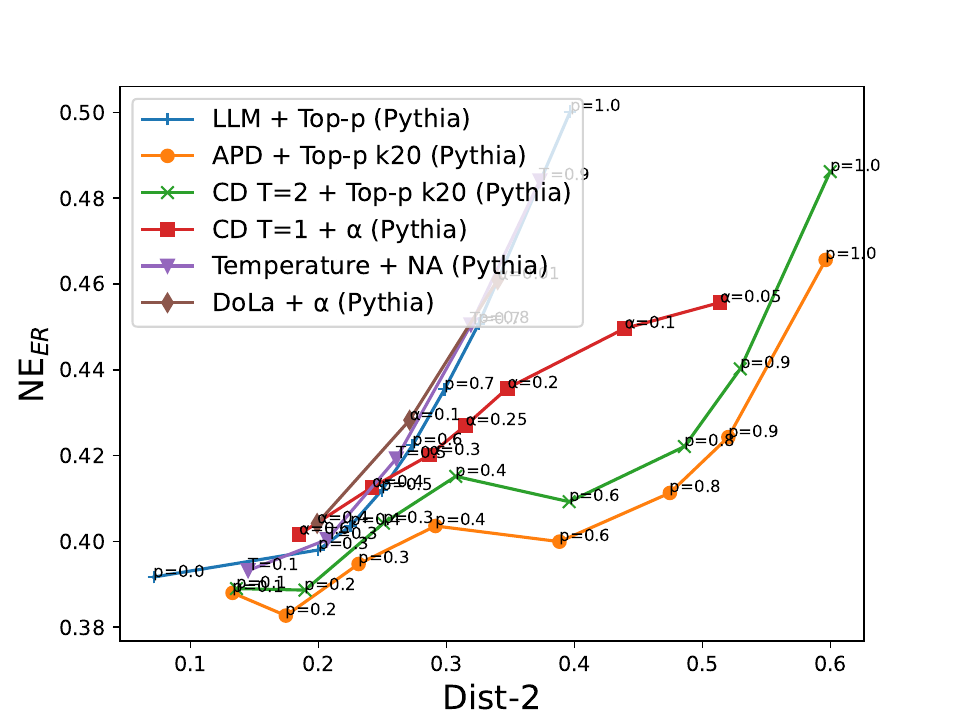}
  \caption{Ours v.s. SOTA (Pythia)}
  \label{fig:div_ne_topp}
\end{subfigure}%
\begin{subfigure}{.33\textwidth}
  \centering
  \includegraphics[width=1\linewidth]{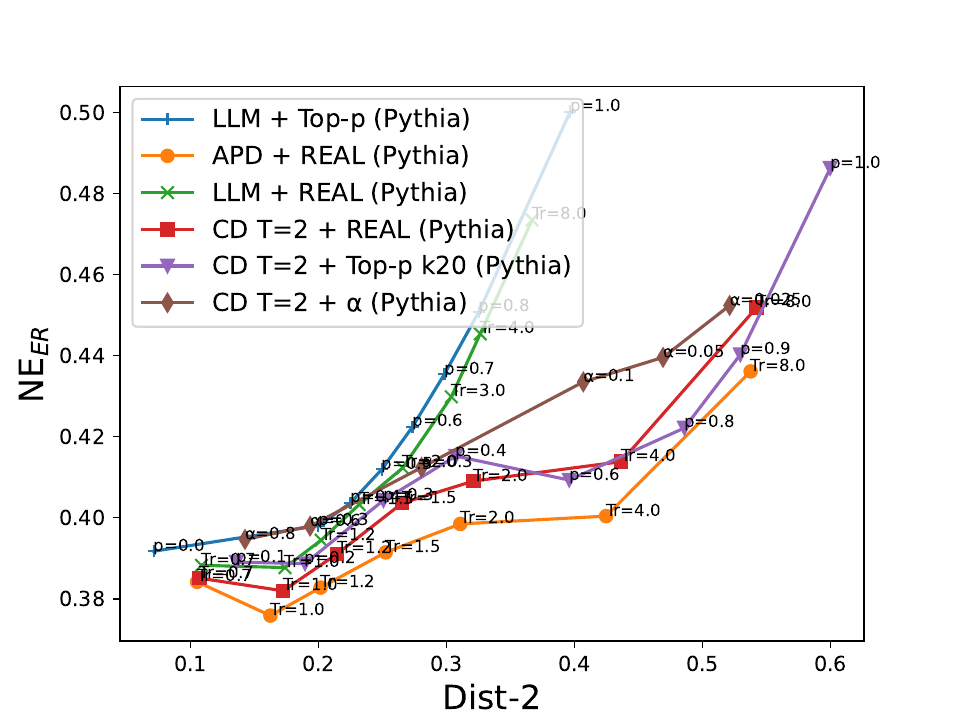}
  \caption{With REAL Sampling (Pythia)}
  \label{fig:div_ne_cd}
\end{subfigure}%
\begin{subfigure}{.33\textwidth}
  \centering
  \includegraphics[width=1\linewidth]{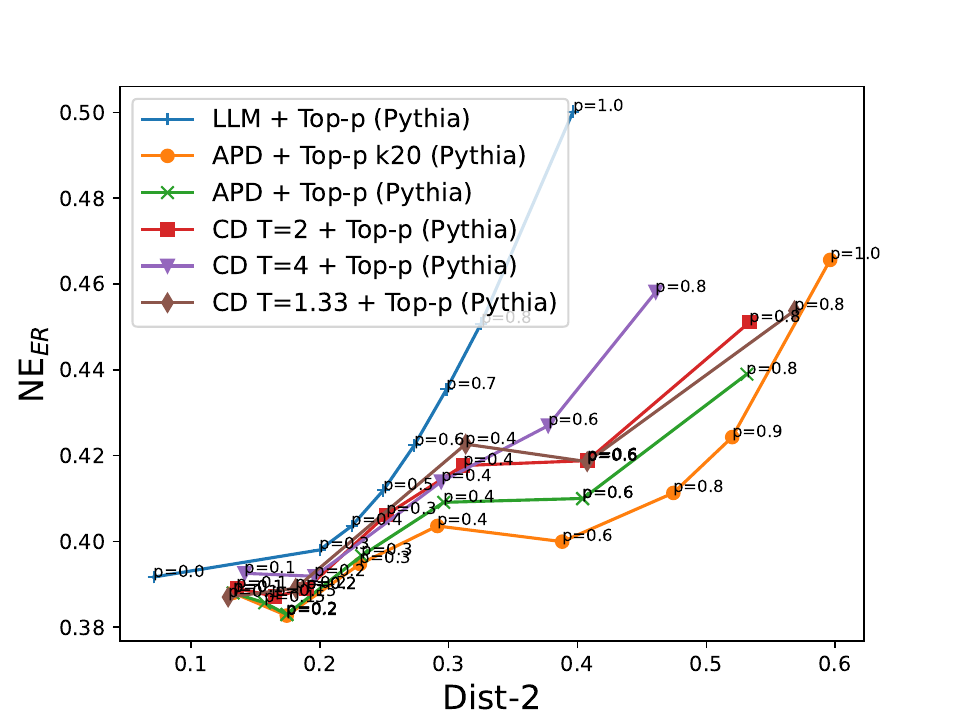}
  \caption{Temperature Tuning of CD (Pythia)}
  \label{fig:div_ne_temp}
\end{subfigure}
\begin{subfigure}{.33\textwidth}
  \centering
  \includegraphics[width=1\linewidth]{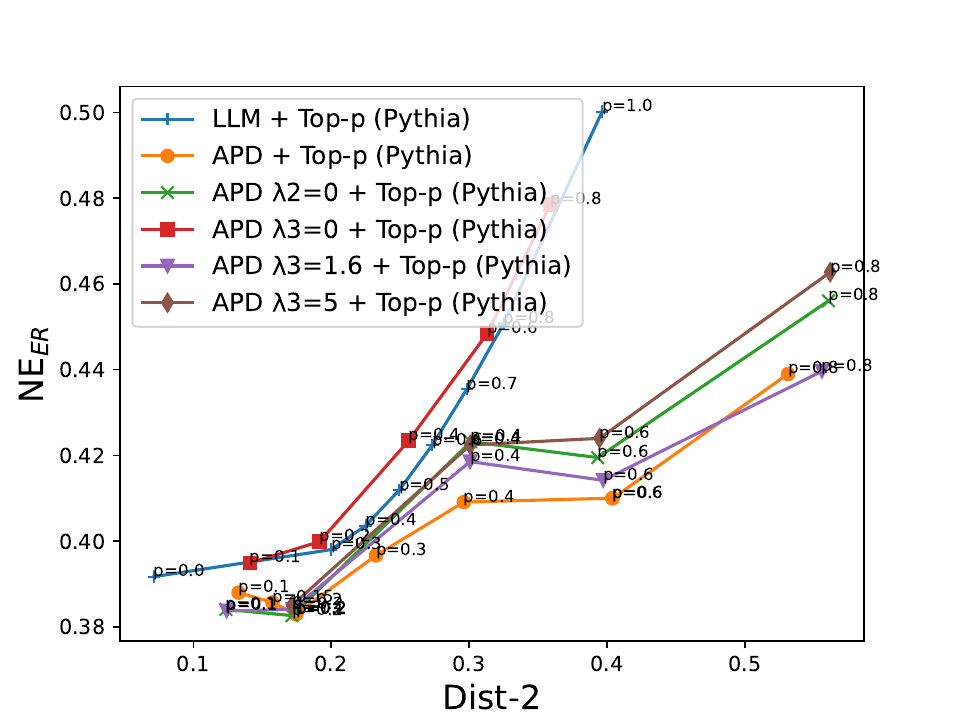}
  \caption{Loss Term Ablation (Pythia)}
  \label{fig:div_ne_term_ablation}
\end{subfigure}%
\begin{subfigure}{.33\textwidth}
  \centering
  \includegraphics[width=1\linewidth]{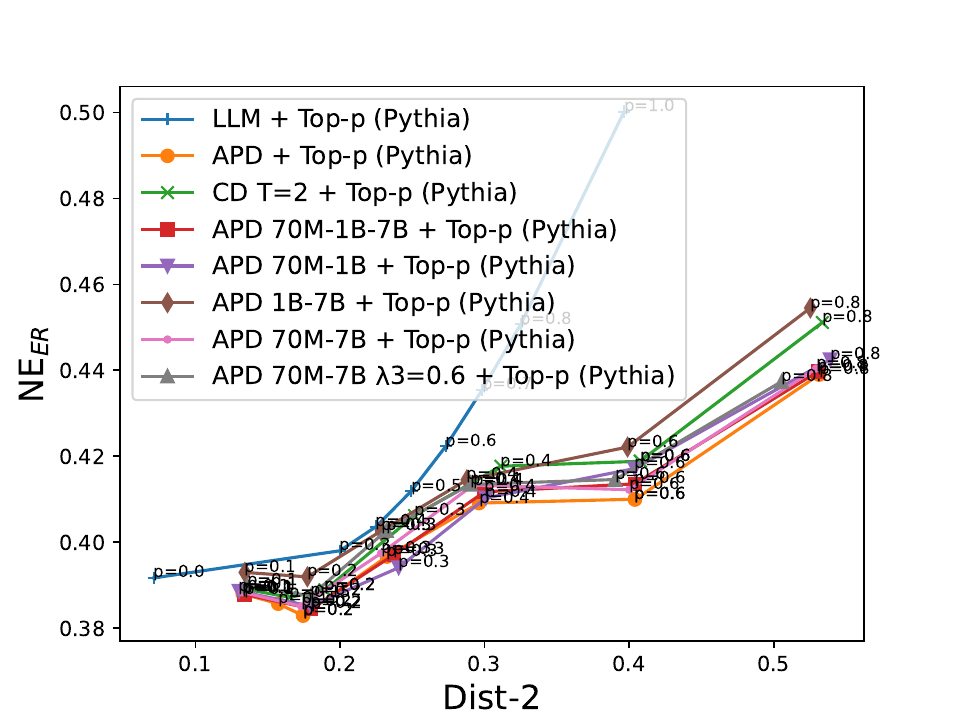}
  \caption{Mid-size LLM Ablation (Pythia)}
  \label{fig:div_ne_model_ablation}
\end{subfigure}%
\begin{subfigure}{.33\textwidth}
  \centering
  \includegraphics[width=1\linewidth]{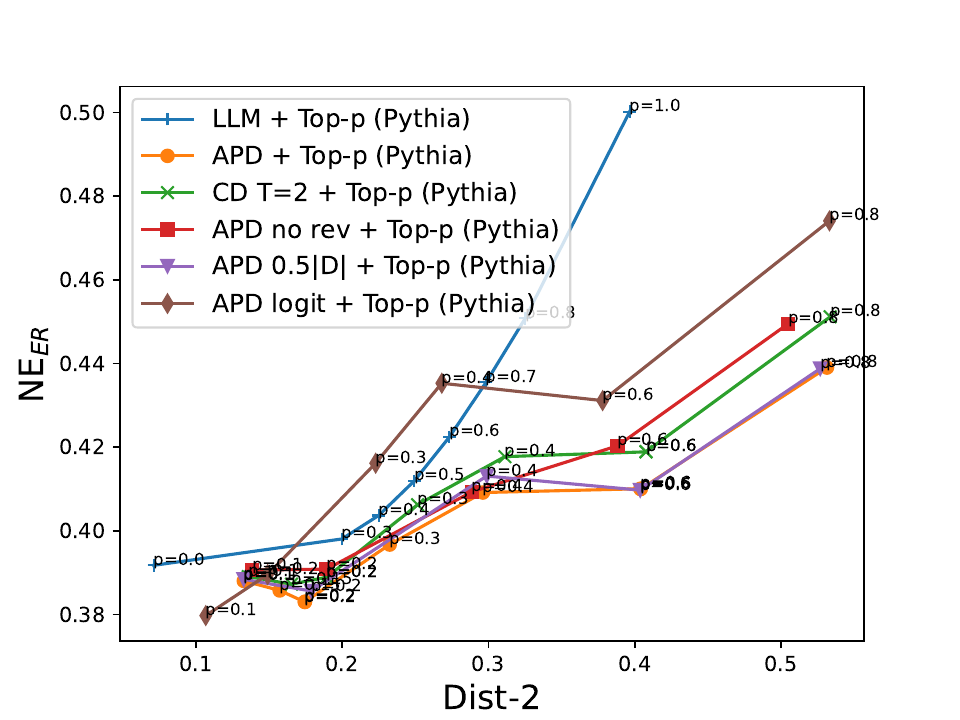}
  \caption{Other Ablation (Pythia)}
  \label{fig:div_ne_other_ablation}
\end{subfigure}
\begin{subfigure}{.33\textwidth}
  \centering
  \includegraphics[width=1\linewidth]{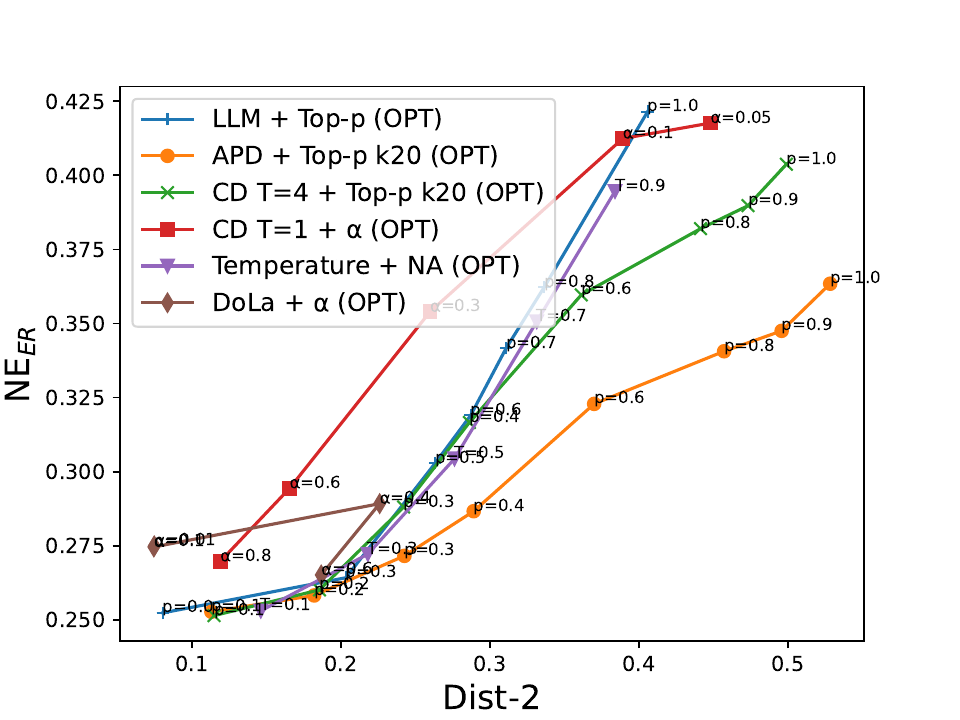}
  \caption{Ours v.s. SOTA (OPT)}
  \label{fig:div_ne_opt}
\end{subfigure}%
\begin{subfigure}{.33\textwidth}
  \centering
  \includegraphics[width=1\linewidth]{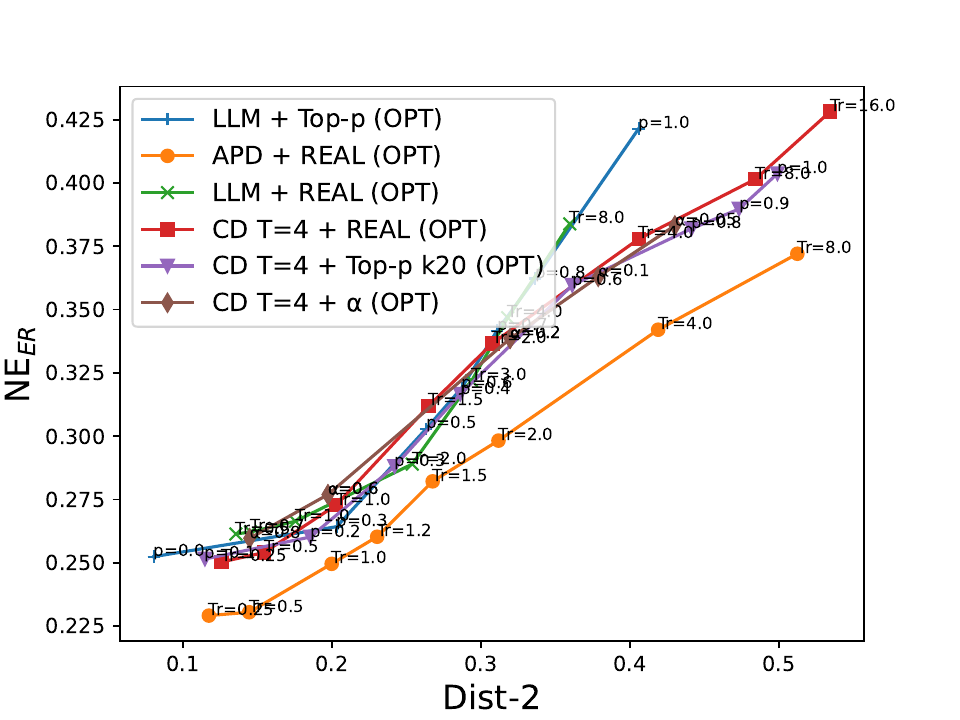}
  \caption{With REAL Sampling (OPT)}
  \label{fig:div_ne_opt_real}
\end{subfigure}%
\begin{subfigure}{.33\textwidth}
  \centering
  \includegraphics[width=1\linewidth]{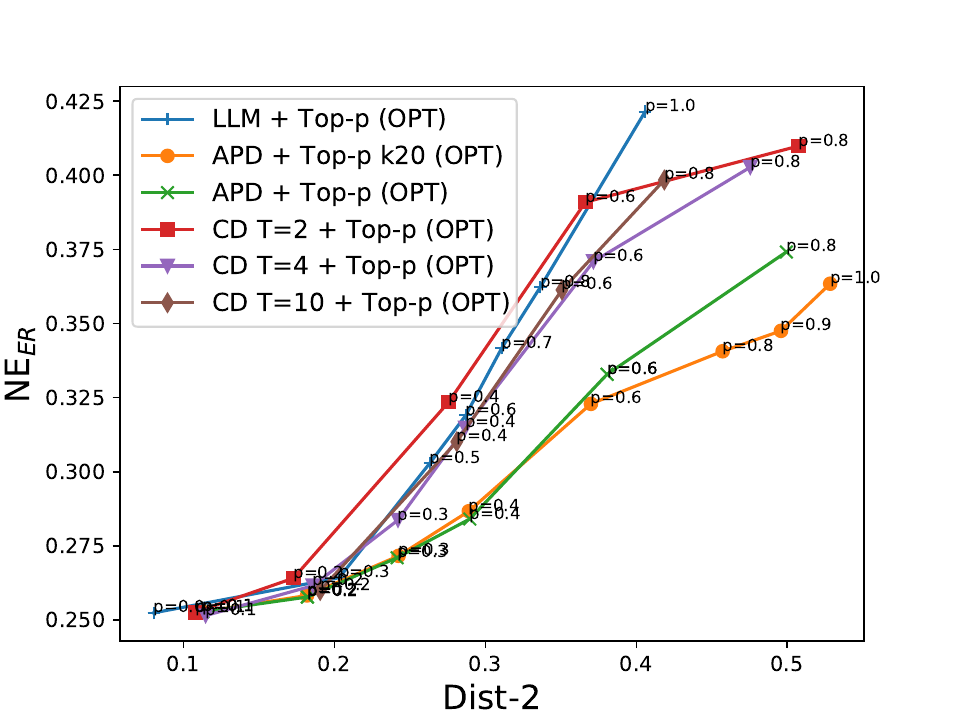}
  \caption{Temperature Tuning of CD (OPT)}
  \label{fig:div_ne_cd_temp}
\end{subfigure}
\caption{Factuality evaluation of the open-ended text generation using \textsc{FactualityPrompts}. The x-axis is a diversity metric (dist-2) and the y-axis is a hallucination metric (NE$_{ER}$), the ratio of containing potential hallucinated entities in generation, so the curves closer to the lower right corner are better.}
\label{fig:comp_div_NE_gen}
\end{figure*}

\section{Experiments}


In many applications, factuality is arguably the most important aspect~\citep{huang2023survey}. It is especially difficult to improve factuality in general domains without increasing LM's size, training LMs on more high-quality data, or adding more information into the context~\citep{tonmoy2024comprehensive}. 

\Cref{fig:first_fig} shows that CD might assign a lower probability to the most obvious answer. In open-ended generation tasks, this could cause suboptimal factuality and diversity. Furthermore, this could degrade the quality of answers in the question-answering tasks. Thus, we focus on these aspects in our evaluation.

In all experiments, APD is trained in a very small training corpus, just 1M lines (1.6\%) in Wikipedia 2021, to test the generalization capability of our method. Pythia, OPT, and Qwen LLM families are the ideal test beds because they provide several smaller LMs of different sizes. Hence, we select de-duplicated Pythia 6.9B and OPT-6.7B as the ELM ($\theta_{s_N}$) and Pythia 70M and OPT-125M as the ALM ($\theta_{s_1}$). The results of Qwen-1.5 are presented in the appendix.

\subsection{Open-ended Text Generation Evaluation using \textsc{FactualityPrompts}}

\textsc{FactualityPrompts}~\citep{lee2022factuality} is an open-ended text generation benchmark that provides 8k factual sentences and 8k non-factual sentences from Wikipedia as prompts. Given a prompt, a good decoding method could output factual and diverse continuations. 

\expsec{Metrics:}
Evaluating the factuality of the generated text on a large scale is a challenging task. \citet{lee2022factuality} propose to measure the ratio of the named entities that are not mentioned in the relevant Wikipedia pages and call it named entity error (NE$_{ER}$). Besides, it also measures Entail$_{R}$, which is the ratio of generated sentences that are supported/entailed by the sentences in the relevant pages. In addition to the automatic retrieval-based metrics, we also conduct expensive and small-scale human experiments in \Cref{sec:human_eval}.

Each decoding method would generate 8 different continuations given a prompt. Then, the generation diversity of the method is evaluated by the ratio of unique bi-gram (Dist-2)~\citep{li2016diversity} and the ratio of generating continuations with severe repetitions (Rep)~\citep{holtzman2019curious}. 


\expsec{Methods:}
To simplify our experiments, every method uses sampling as in \citet{o2023contrastive} rather than beam search. Every sampling method is denoted as ``\textbf{distribution modification + thresholding}''. The distribution modification methods we tested include \textbf{LLM} (i.e., the probabilities from the ELM), \textbf{CD}, \textbf{APD}, \textbf{DoLa}~\citep{chuang2023dola}, and softmax \textbf{temperature} adjustment~\citep{ficler2017controlling}. We fix the $T$ of ALM in APD to be $1$ while select the best $T$ for CD from \{1, 1.33, 2, 4, 10\}.

We conduct a series of ablation studies to verify our design choices. 
First, APD uses $\lambda_2=10$ and $\lambda_3=0.8$ by default in \Cref{eq:loss_all} for Pythia. To verify the effectiveness of each loss term, we individually set \textbf{$\lambda_2=0$}, \textbf{$\lambda_3=0$}, \textbf{$\lambda_3=1.6$}, and \textbf{$\lambda_3=5$}. Second, to check if APD could still be applied to the LLM families without many different model sizes, we test four combinations of training Pythia (\textbf{70M}, \textbf{1B}, and \textbf{6.9B}). Finally, \textbf{APD no rev} removes the reverse function in \Cref{eq:flip} and allows negative $a_{w,c}$ in \Cref{eq:exp}, \textbf{APD 0.5|D|} uses only half of our training Wikipedia, and \textbf{APD logit} applies the exponential decay function in the logit space rather than the probability space in \Cref{eq:exp}. 


The thresholding methods could increase the factuality by filtering out the unlikely tokens from the ELM at the cost of lower diversity. We test the following thresholding methods. 
\begin{itemize}[leftmargin=.1in,topsep=0pt]
\setlength\itemsep{-0.3em}
\item \textbf{NA}: No filtering for temperature sampling.
\item \textbf{$\alpha$}: The default filtering method used by CD~\citep{li2022contrastive,o2023contrastive}.
\item \textbf{Top-$p$}: A widely-used method for controlling the generation diversity~\citep{holtzman2019curious}.
\item \textbf{Top-$p$ $k$20}: Combination of top-p and top-k sampling~\citep{fan2018hierarchical}. The $k$ is fixed to be 20.
\item \textbf{REAL}: A method that dynamically adjusts the threshold in top-p sampling \citep{REAL}.
\end{itemize}

\expsec{Results:}
Across the thresholding methods, LLM families, and the whole diversity spectrum, \textbf{APD} achieves consistent factuality improvement in \Cref{fig:div_ne_topp,fig:div_ne_cd,fig:div_ne_opt,fig:div_ne_opt_real}, and \Cref{tb:gen_both}. \Cref{fig:div_ne_temp,fig:div_ne_cd_temp} indicate that the improvement cannot be achieved by tuning \textbf{CD}'s temperature. \textbf{CD}'s improvement in OPT over \textbf{Top-$p$} sampling is much smaller compared to Pythia, which indicates the linear extrapolation assumption is less valid for OPT, while \textbf{APD} achieves larger improvements for OPT than Pythia by fixing the problems of \textbf{CD}. 

We report our loss ablation study results at \Cref{fig:div_ne_term_ablation}, which verify the importance of each term in \Cref{eq:loss_all}. The degradation of $\lambda_3=5$ suggests that our improvements cannot be achieved by only using the regularization term and our improvements indeed come from modeling the probability decay. \Cref{fig:div_ne_model_ablation} shows that \textbf{APD} performs better as we have more LMs with different model sizes in the LLM family. Nevertheless, \textbf{APD 70M-7B} is still better than \textbf{CD} using only ALM and ELM without any mid-size LMs, which makes APD even more practical. The worse performances of \textbf{APD 1B-7B} highlight the importance of having smaller LMs in modeling the exponential decay curves. In \Cref{fig:div_ne_other_ablation}, \textbf{APD 0.5|D|} performs slightly worse than \textbf{APD}, which suggests a small training corpus is sufficient but larger ones could further expand APD's improvement. Finally, the worse factualities of \textbf{APD logit} and \textbf{APD no rev} verify our choice of modeling the decay curve in the probability space. 


\begin{table}[t!]
\scalebox{0.72}{
\begin{tabular}{cll|cccc}
 & & & \multicolumn{2}{c}{Factuality} & \multicolumn{2}{c}{Diversity} \\
    &       &     & NE$_{ER}$ ($\downarrow$)           & Entail$_R$       & Dist-2         & Rep ($\downarrow$)     \\ \hline      
\multirow{5}{*}{Pythia} & $p=0.8$       &  & 43.28 &	5.40 &	29.73 &	1.26  \\
                        & $p=0.8$ & CD  & 42.21          & 3.54          & \textbf{48.57} & \textbf{0.44}          \\
                        & $p=0.6$ & CD  & 40.93 &	6.02 &	39.58 &	1.48 \\ 
                        & $p=0.8$ & APD  & 41.13 & 4.47 & 47.44          & 1.46          \\
                        & $p=0.6$ & APD  & \textbf{40.00} &	\textbf{6.58} &	38.81 &	2.72 \\ \hline
\multirow{5}{*}{OPT}    & $p=1$       &  & 34.33 &	10.70 &	31.29 &	2.93   \\
                        & $p=0.8$ & CD  & 38.20          & 7.56          & 44.13          & \textbf{0.78}          \\
                        & $p=0.6$ & CD  & 35.96 &	11.66 & 36.12 &	2.49  \\ 
                        & $p=0.8$ & APD  & 34.06 & 8.88 & \textbf{45.72} & 3.39         \\
                         & $p=0.6$ & APD  & \textbf{32.29} &	\textbf{13.38} &	36.99 &	5.15  \\ 
\end{tabular}
}
\caption{Comparing different distributions for top-$p$ k20 sampling in \textsc{FactualityPrompts}. All numbers are percentages. APD is significantly more factual than CD while having similar diversity.} 
\label{tb:gen_both}
\end{table}

\begin{table*}[t!]
\scalebox{0.65}{
\begin{tabular}{cl|c|cc|cccc|cc|cc}
\multicolumn{1}{l}{}    &                 & LAMBADA        & \multicolumn{2}{c|}{CQA}         & \multicolumn{4}{c|}{QASC}                                          & \multicolumn{2}{c|}{ARC}         & \multicolumn{2}{c}{SocialIQA}   \\
\multicolumn{1}{l}{}    &                 &                &                &                & \multicolumn{2}{c}{Q+Fact}      & \multicolumn{2}{c|}{Q Only}           &                &                &                &                \\
\multicolumn{1}{l}{}    &                 & ppl ($\downarrow$)            & ppl ($\downarrow$)            & acc            & ppl ($\downarrow$)            & acc            & ppl ($\downarrow$)            & acc            & ppl ($\downarrow$)            & acc            & ppl ($\downarrow$)            & acc            \\ \hline
 \multirow{7}{*}{Pythia} & LLM 6.9B & 2.264 & 8.380 & 0.658          & 5.702          & 0.856          & 8.127          & 0.621          & 4.433          & 0.692          & 8.441          & 0.662          \\
 & CD & 2.237 & 6.176 & 0.671 & 5.693 & 0.862          & \textbf{7.741} & \textbf{0.633} & 4.375          & \textbf{0.699} & 7.595          & 0.688          \\
 & APD &  \textbf{2.132}$\dagger$ & \textbf{5.882}$\dagger$ & \textbf{0.685} & \textbf{5.020}$\dagger$ & \textbf{0.874} & 7.766          & 0.632          & \textbf{4.310} & 0.698          & \textbf{7.378} & \textbf{0.691} \\
 & APD on the fly & 2.281 & 8.245 & 0.660 & 5.725 & 0.866 & 8.106 & 0.620 & 4.464 & 0.694 & 8.299 & 0.665          \\ \cline{2-13}
 & LLM 12B & 2.188 & 8.140 & 0.660      & 4.783          & 0.845          & 7.612          & 0.630          & 4.058          & 0.719          & 7.898          & 0.691          \\
 & APD vs CD &  138.52\%       & 122.34\%       & 650.00\%       & 73.30\%        & NA             & -4.86\%        & -12.50\%       & 17.34\%        & -4.17\%        & 39.92\%        & 11.54\%        \\
 & APD vs LLM 6.9B &  173.68\%       & 1039.11\%      & 1250.00\%       & 74.26\%        & NA             & 70.06\%        & 125.00\%       & 32.87\%        & 20.83\%        & 195.88\%       & 100.00\% \\ \hline

\end{tabular}
}
\caption{Perplexity (ppl) and accuracy (acc) comparison of one-shot QA using different decoding methods. The LLM 6B and 12B use the original distribution from ELM, which is the most popular SOTA method. APD vs CD is (APD - CD) / (LLM 12B - LLM 6.9B) (i.e., the ratio of improvement against CD and against doubling the model size), and APD vs LLM 6.9B is (APD - LLM 6.9B) / (LLM 12B - LLM 6.9B). NA means LLM 12B is worse than LLM 6.9B. CQA refers to CommonsenseQA. A lower perplexity is better. We highlight the best score among the distributions from ELM, CD, APD, and APD on the fly. $\dagger$APD is significantly better than CD with $p<0.05$. }
\label{tb:QA_ppl_pythia}
\end{table*}



\subsection{Distribution Evaluation using Question Answering Datasets}

Another way to evaluate the factuality of the next-token distribution is through question answering and checking which distribution assigns a higher probability to the correct answer(s). The evaluation method allows us to analyze APD's improvement gap in various domains and settings.

We compare APD and CD using the following 5 popular commonsense QA datasets, including LAMBADA~\citep{paperno2016lambada}, CommonsenseQA~\citep{talmor-etal-2019-commonsenseqa}, QASC~\citep{khot2020qasc}, ARC~\citep{clark2018think}, and SocialIQA~\citep{sap2019social}. In QASC, we report two results. One setting inputs only the question (Q only) and another one also inputs facts (Q+Fact).

\expsec{Evaluation Setup:}
The LLMs are evaluated using a one-shot in-context learning prompt. We use perplexity and accuracy of the correct answer(s) as our main metrics. The mean reciprocal rank (MRR)~\citep{radev2002evaluating} scores will also be reported at \Cref{tb:QA_mrr_both} in \Cref{sec:qa_more}. In each task, the optimal $T$ is different. We report the performances of CD and APD using their best $T$ from \Cref{eq:CD_prob} and \Cref{eq:AP}, respectively.


\textbf{CD} and \textbf{APD} are applied to Pythia 6.9B and the results of OPT and Qwen are reported at \Cref{sec:qa_more}. We also report the performance of Pythia 6.9B (\textbf{LLM 6.9B}) and Pythia 12B (\textbf{LLM 12B}) to compare the APD's improvement with the improvement of roughly doubling the LM size. \textbf{APD on the fly} is an ablation study that removes the fine-tuning step of ALM'. The decoding method first get $\{p(w|c,\theta_{s_i})\}_{i=1}^N$ from the $N$ LLMs. Next, it extrapolates the probabilities to estimate the asymptotic probability and curve parameters in \Cref{eq:exp} using gradient descent. More details can be found at \Cref{sec:APD_no_FT}.

To focus on the questions for which LLM is highly likely to provide the correct answer, we filter out the questions whose correct answers are not ranked in the top $20$ list of ELM (LLM 6.9B). We choose $20$ because ALM' is mostly trained to predict the probabilities of the top $20$ tokens. To make the comparison fair, we repeat this filtering preprocessing using LLM 12B. 

\expsec{Results:}
In \Cref{tb:QA_ppl_pythia}, \textbf{APD} usually improves \textbf{CD} significantly even though ALM' is not trained on commonsense question-answering datasets, which emphasizes the out-of-domain robustness of \textbf{APD}. \textbf{APD} performs similarly with \textbf{CD} if the prompt only contains a question from QASC. However, \textbf{APD} is drastically better than \textbf{CD} after we reduce the difficulty of the question by inserting the relevant facts into the prompt. Furthermore, our improvements on easier datasets such as LAMBDA and CommonsenseQA (CQA) are also larger. This also supports the recent findings that \textbf{CD} might not improve the commonsense questions answering~\citep{o2023contrastive}. Finally, the significantly worse performance of \textbf{ADP on the fly} suggests that fine-tuning ALM' makes APD not only more efficient but also more effective.

\section{Related Work}



Besides simulating a huge LM as CD, linear extrapolation could also be used to infer the output of an LM trained on less toxic data~\citep{liu2021dexperts}, an LM trained on more context~\citep{shi2023trusting,sanchezstay}, an LM trained on less hallucinated data~\citep{zhang2023alleviating}, an LM trained on more preference data~\citep{zheng2024weak}, and an LM with more hidden layers~\citep{chuang2023dola,das2024entropy}. APD shows that an exponential function can improve CD, so similarly, we might be able to use a sophisticated extrapolation function to improve these methods.

Scaling law shows that the global perplexity of a larger LM could be accurately extrapolated by the perplexities of smaller LMs~\citep{kaplan2020scaling}. For individual tasks, the performances of a larger LM are also predictable based on the performance of smaller LMs~\citep{srivastava2023beyond,ruan2024observational,owen2024predictable}. In our work, we show that it is possible to efficiently extrapolate the next-token probability distribution of a larger LM.

\citet{REAL} propose REAL sampling, which improves the factuality by extrapolating the entropy of an infinitely large LM. Similar to our work, 
\citet{REAL} conduct nonlinear extrapolation across LMs model sizes. 
Nevertheless, our motivations and methods are very different. Furthermore, REAL sampling is a thresholding method while our method is a distribution modification method, and our experiment shows that they are complementary. 








\section{Conclusion}
Can a tiny amateur LM help a large expert LM infer the probabilities of a huge hypothetical LM? Yes, we first theoretically show that CD did that and then propose APD to do even better. Our theory for CD explains several prior empirical observations and motivates the proposed APD. On the other hand, the APD addresses the limitations of CD and further supports our theoretical explanation. In our experiments, we show that fine-tuning the amateur LM on only 1.6\% text of Wikipedia is enough to boost the performances of seven QA datasets and the factuality in \textsc{FactualityPrompts}, judged by both retrieval-based metrics and Mturk workers, given the similar generation diversities.

\section{Acknowledgement}

This work was supported in part by the Center for Intelligent Information Retrieval. Any opinions, findings and conclusions or recommendations expressed in this material are those of the authors and do not necessarily reflect those of the sponsor.




\section{Limitations}


First, APD requires a series of LMs of different sizes that are trained in the same data for a similar amount of time. Due to the trade-off between efficiency and quality, many state-of-the-art LLMs have smaller counterparts, including GPT-4, Gemini 1.5, Claude 3, and LLaMA 3. However, the LLM families often lack the LMs smaller than 7B, which is close to the largest model we can fit into our GPUs. Thus, we are not able to test our methods on these powerful LLMs. Recently, \citet{yang2023frustratingly} show that simple n-gram LMs could also become good amateur models. This might alleviate the existence requirement of small pretrained LMs.

Second, ideally, ALM' should be fine-tuned for a long time on the same corpus for pretraining ALM and ELM. Our ablation studies in \Cref{fig:div_ne_other_ablation} also show that a larger fine-tuning corpus indeed improves the performance. However, due to our limited GPU and CPU memory, we only fine-tune ALM' on 1.6\% text of Wikipedia. Although the limited amount of training is sufficient to make APD outperform CD significantly, this prevents us from comparing APD with CD on more tasks such as reasoning or summarization. 

Third, our ablation studies in \Cref{fig:div_ne_term_ablation} show that the performances of APD depend on the values of $\lambda_3$ and the optimal value is different for different LLMs. Fortunately, we find that APD still performs well in various QA datasets even though its $\lambda_3$ is tuned for the validation set of \textsc{FactualityPrompts}. This suggests that tuning the $\lambda_3$ is needed when applying the APD on a new LLM family but might not be necessary for new tasks. 

Finally, our theory does not cover the case when the $T$ is larger than $1$. \citet{li2022contrastive} show that the smaller $T$ is beneficial to some creative writing tasks and this observation cannot be explained by our current theory.





\section{Impact Statement}
First, our work shows that it is possible to predict the probability distribution of a larger LLM well without extra training data or substantial computational cost. This finding suggests that the current cross-entropy loss during pretraining may not be optimal. We should be able to discover more efficient ways to train our LLMs and save more energy in the future (e.g., we might be able to train the ELM to directly predict the asymptotic probabilities).

Second, our theory and empirical results (e.g., \Cref{fig:div_ne_model_ablation}) indicate that the effectiveness of CD and APD depends on whether we have the LMs that are sufficiently small in the LLM family and trained using a similar setup. The result might encourage the LLM developers to train smaller LLMs and better control the training setup (e.g., using the same training text order as in Pythia).



Third, the x-axis is the model size in CD. In other variants of CD, the x-axis could be the amount of hallucinated data~\citep{zhang2023alleviating}, the amount of toxic data~\citep{liu2021dexperts}, or the amount of preference data~\citep{zheng2024weak}. We should be able to extend our theory and APD to these applications as well. 

Finally, we are not certain about whether the exponential function could fit the probability curves of other LLMs, especially after SFT or RLHF~\citep{ouyang2022training}. Besides, we haven't scaled up the training of ALM' and the evaluation of APD, so the out-of-domain generalization ability of APD remains unknown. If the LLM developers deployed our method without comprehensive testing, it might output strange or even unsafe results. 






\bibliography{custom}

\newpage

\appendix

\section{Appendix Overview}
\label{sec:appendix}


We first visualize the probability predicted by APD and compare more methods using more metrics in \Cref{sec:more_results}. Then, we conduct a human experiment and evaluate the creative writing ability of APD in \Cref{sec:more_exp}. Finally, we provide the details of our methods in \Cref{sec:method_details} and the details of our experiments in \Cref{sec:exp_details}.
\section{More Metrics and Analyses}
\label{sec:more_results}

Because of the page limit, we move visualization and some results of \textsc{FactualityPrompts} and question-answering datasets to this section. 


\begin{figure*}[t!]
\centering
\includegraphics[width=1\linewidth]{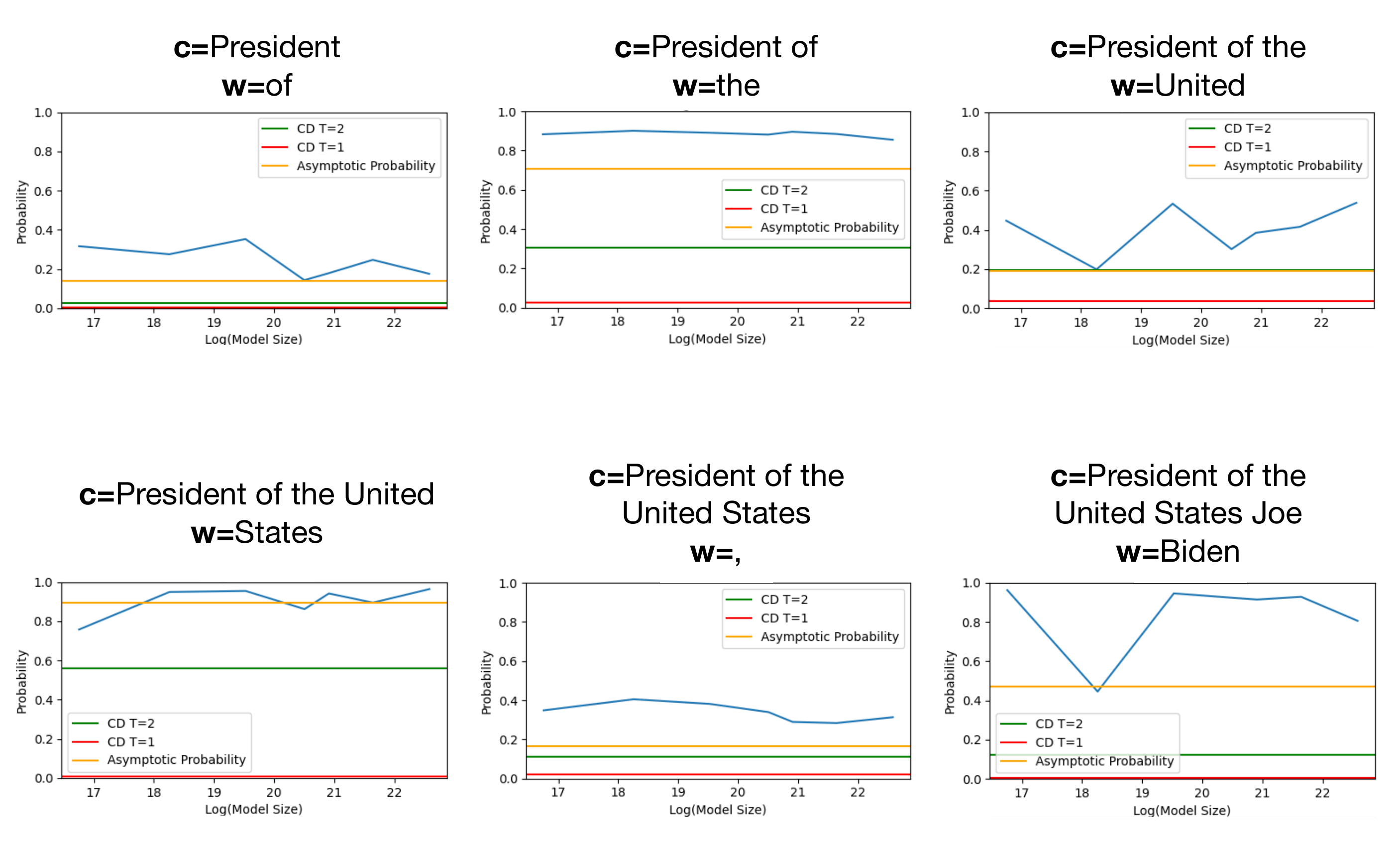}
\caption{Comparison of empirical probability curves (blue) and probabilities predicted by APD (orange) and CD (green and red). The next token $w$ has the highest probability in ELM. The LLM family is Pythia. }
\label{fig:vis}
\end{figure*}

\begin{figure*}[t!]
\centering
\begin{subfigure}{.33\textwidth}
  \centering
  \includegraphics[width=1\linewidth]{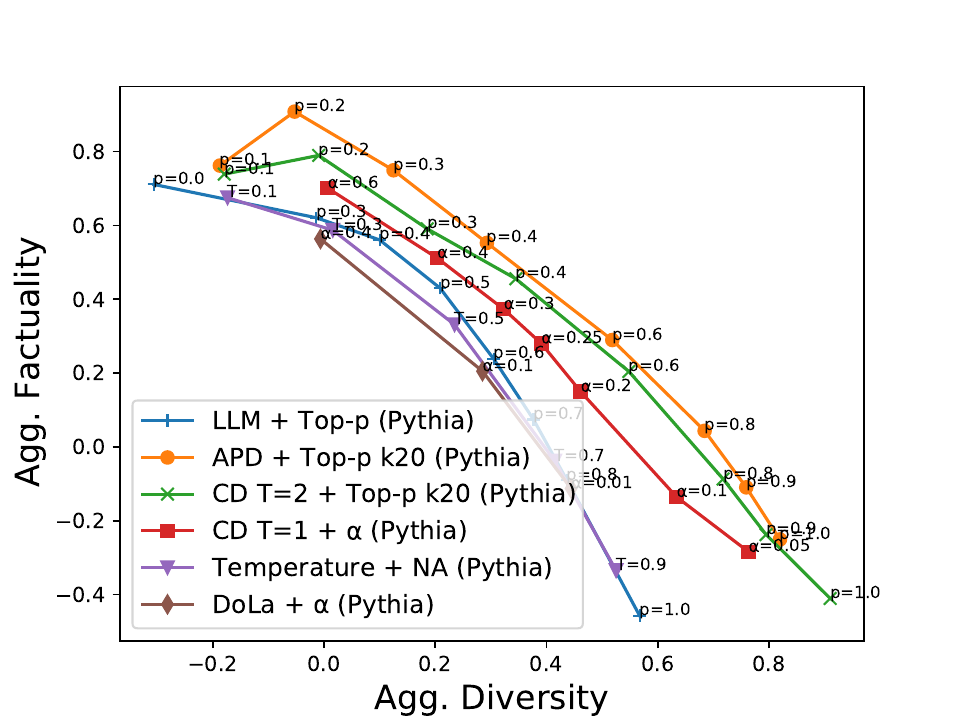}
  \caption{Ours v.s. SOTA (Pythia)}
  \label{fig:topp}
\end{subfigure}%
\begin{subfigure}{.33\textwidth}
  \centering
  \includegraphics[width=1\linewidth]{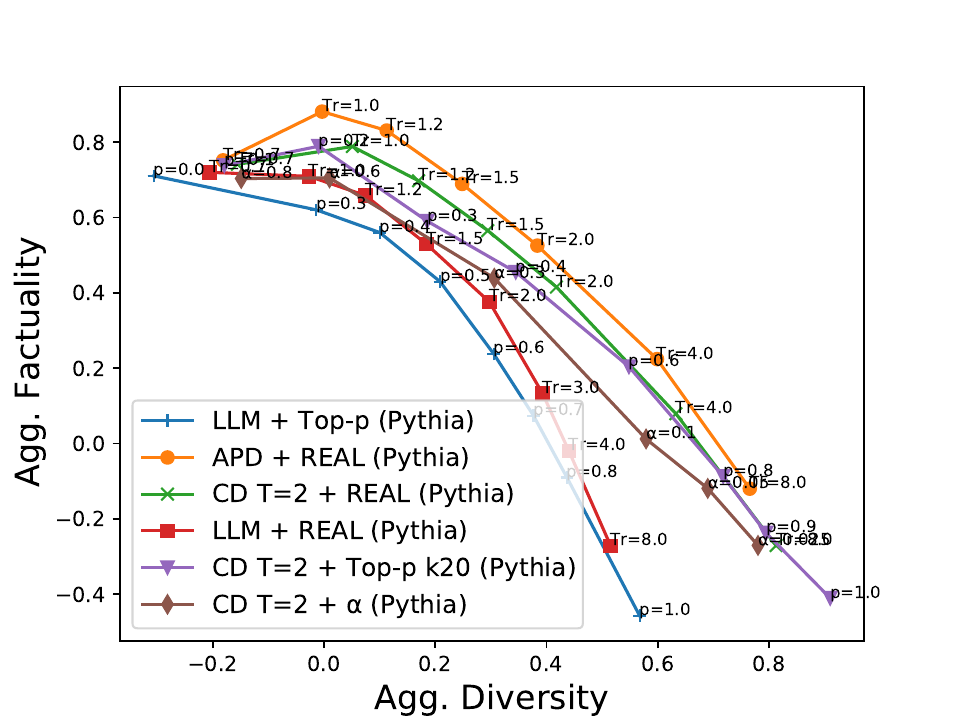}
  \caption{With REAL Sampling (Pythia)}
  \label{fig:cd}
\end{subfigure}
\begin{subfigure}{.33\textwidth}
  \centering
  \includegraphics[width=1\linewidth]{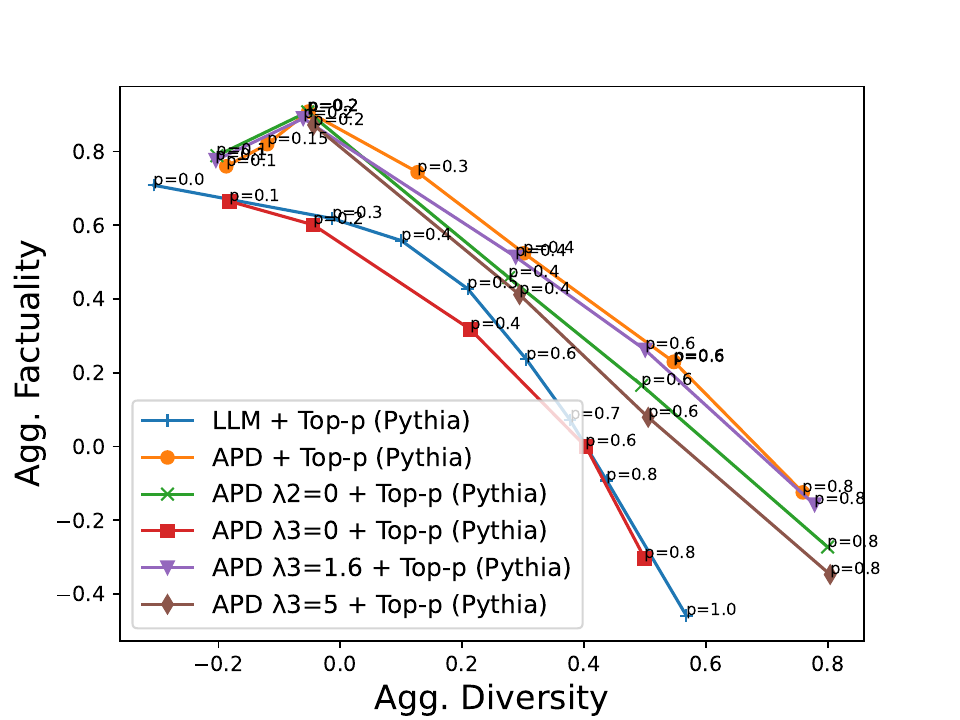}
  \caption{Loss Term Ablation (Pythia)}
  \label{fig:term_ablation}
\end{subfigure}%
\begin{subfigure}{.33\textwidth}
  \centering
  \includegraphics[width=1\linewidth]{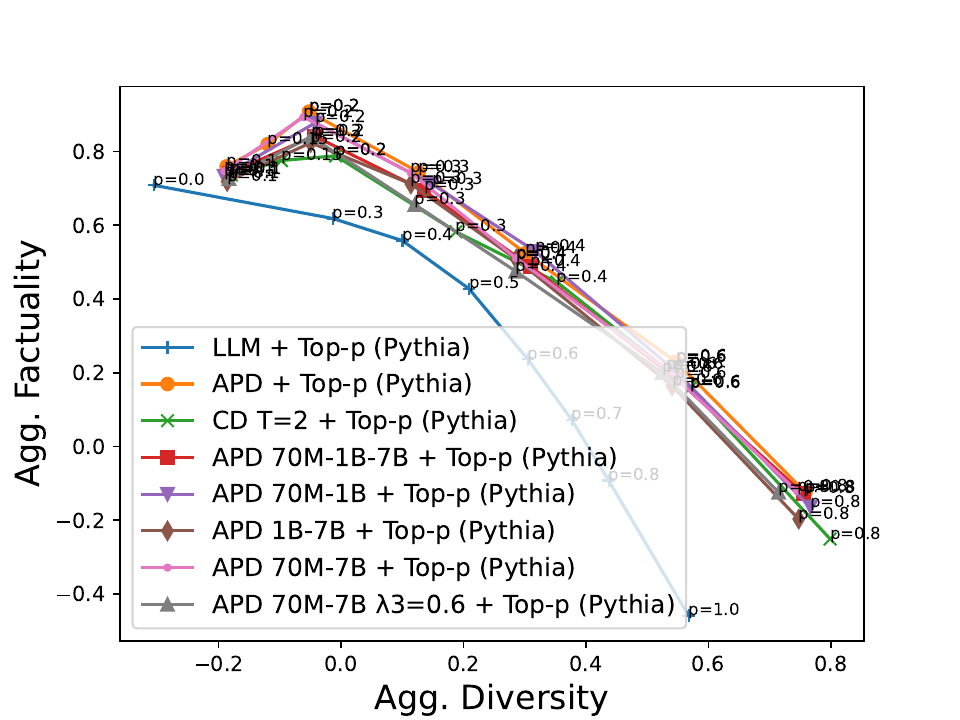}
  \caption{Mid-size LLM Ablation (Pythia)}
  \label{fig:model_ablation}
\end{subfigure}%
\begin{subfigure}{.33\textwidth}
  \centering
  \includegraphics[width=1\linewidth]{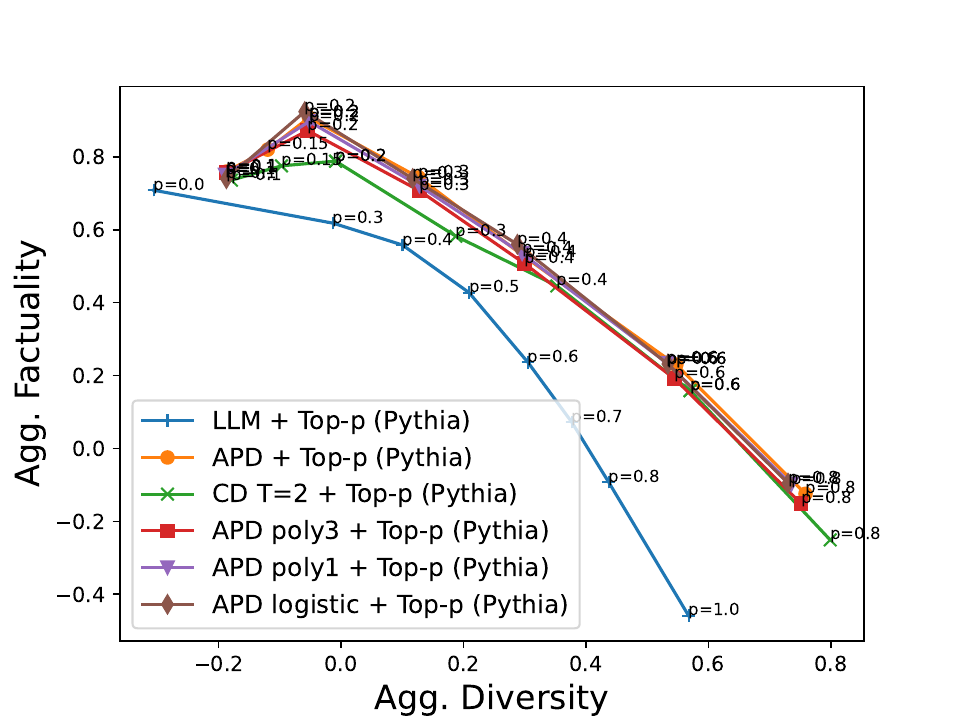}
  \caption{Function Ablation (Pythia)}
  \label{fig:func_ablation}
\end{subfigure}
\begin{subfigure}{.33\textwidth}
  \centering
  \includegraphics[width=1\linewidth]{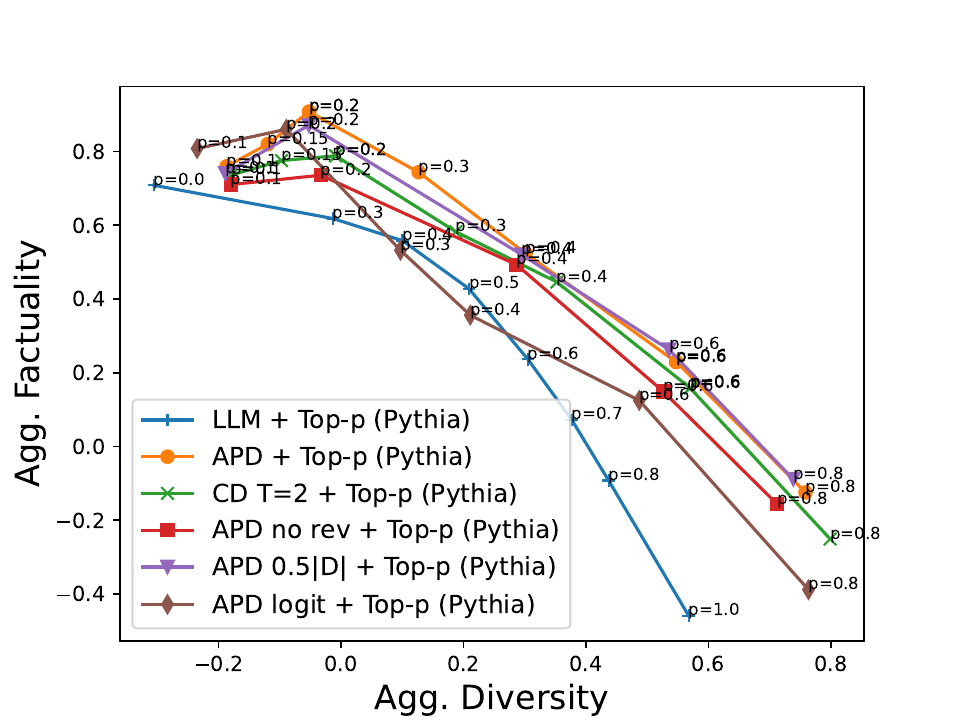}
  \caption{Other Ablation (Pythia)}
  \label{fig:other_ablation}
\end{subfigure}%
\begin{subfigure}{.33\textwidth}
  \centering
  \includegraphics[width=1\linewidth]{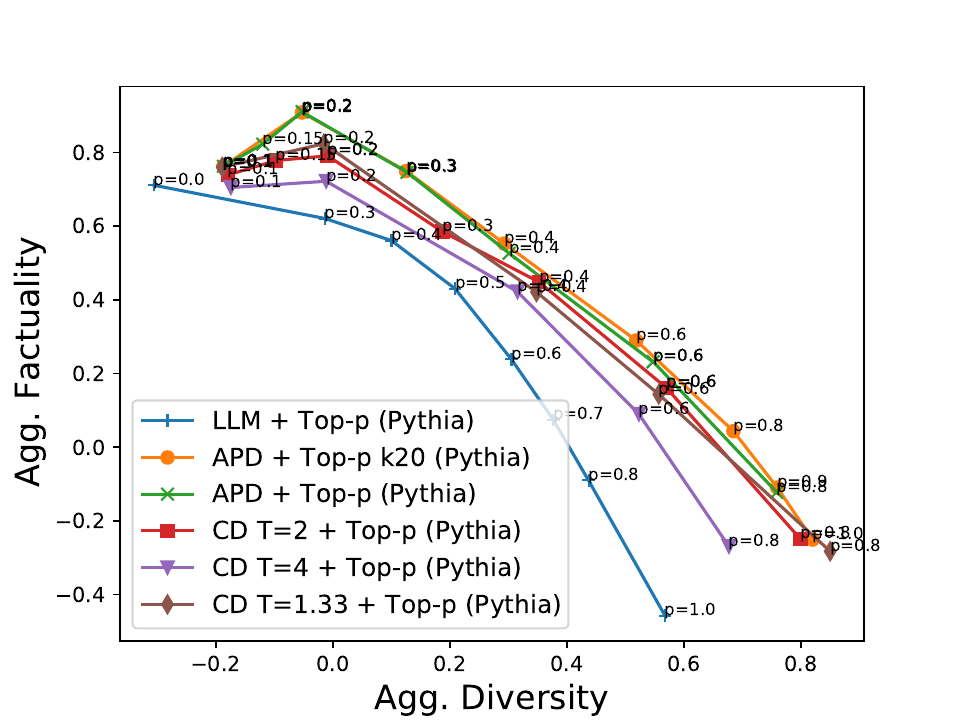}
  \caption{Temperature Tuning of CD (Pythia)}
  \label{fig:cd_temp}
\end{subfigure}%
\begin{subfigure}{.33\textwidth}
  \centering
  \includegraphics[width=1\linewidth]{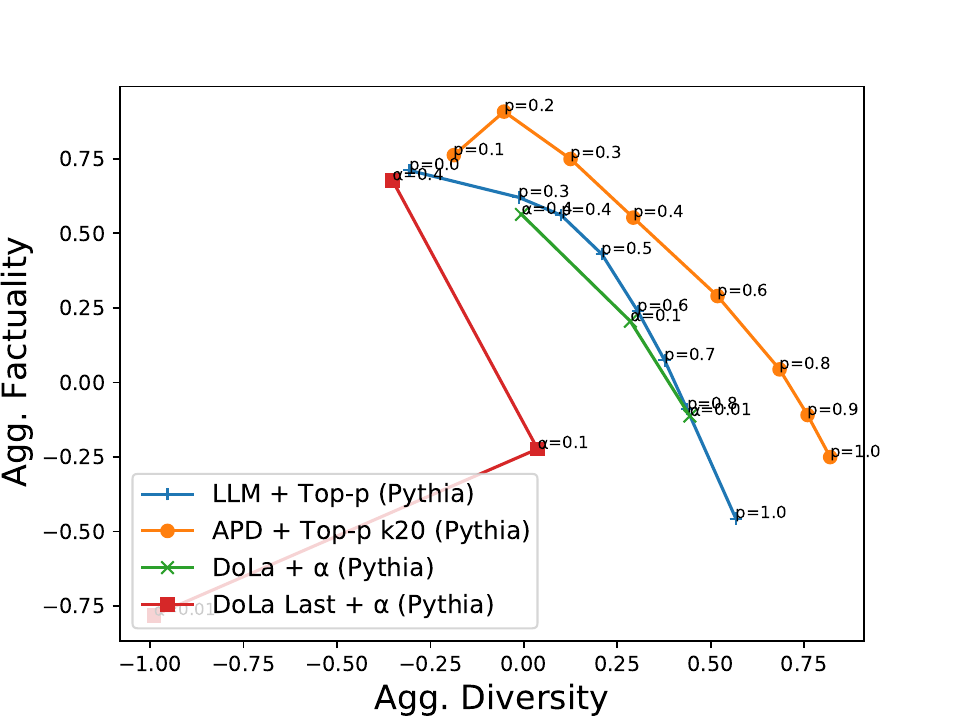}
  \caption{Layer Tuning of DoLa (Pythia)}
  \label{fig:dola_temp}
\end{subfigure}
\begin{subfigure}{.33\textwidth}
  \centering
  \includegraphics[width=1\linewidth]{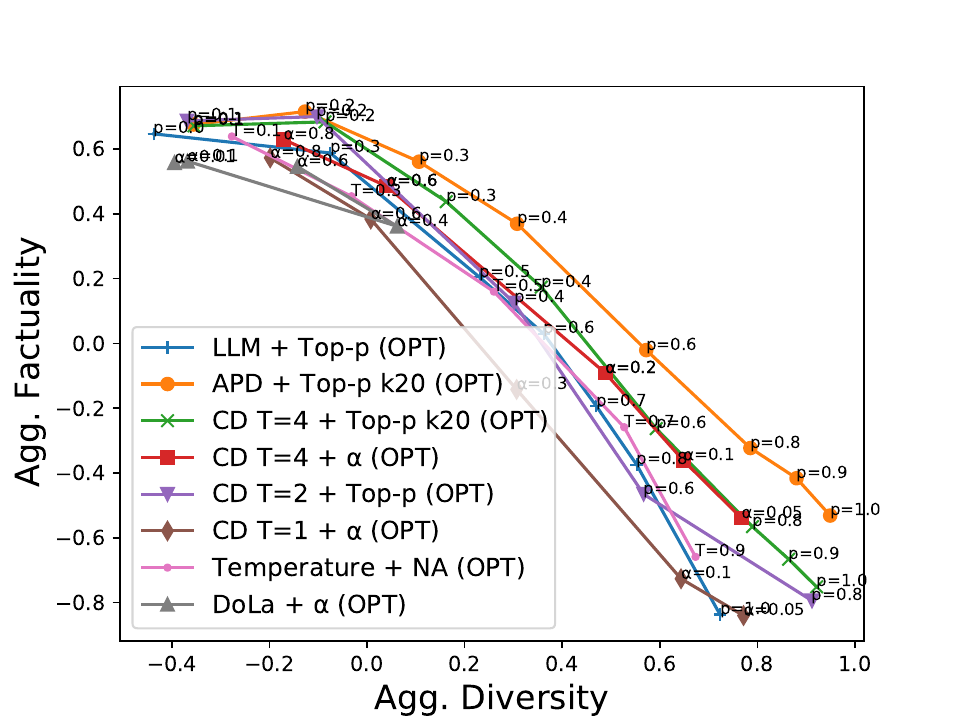}
  \caption{Ours v.s. SOTA (OPT)}
  \label{fig:opt}
\end{subfigure}%
\begin{subfigure}{.33\textwidth}
  \centering
  \includegraphics[width=1\linewidth]{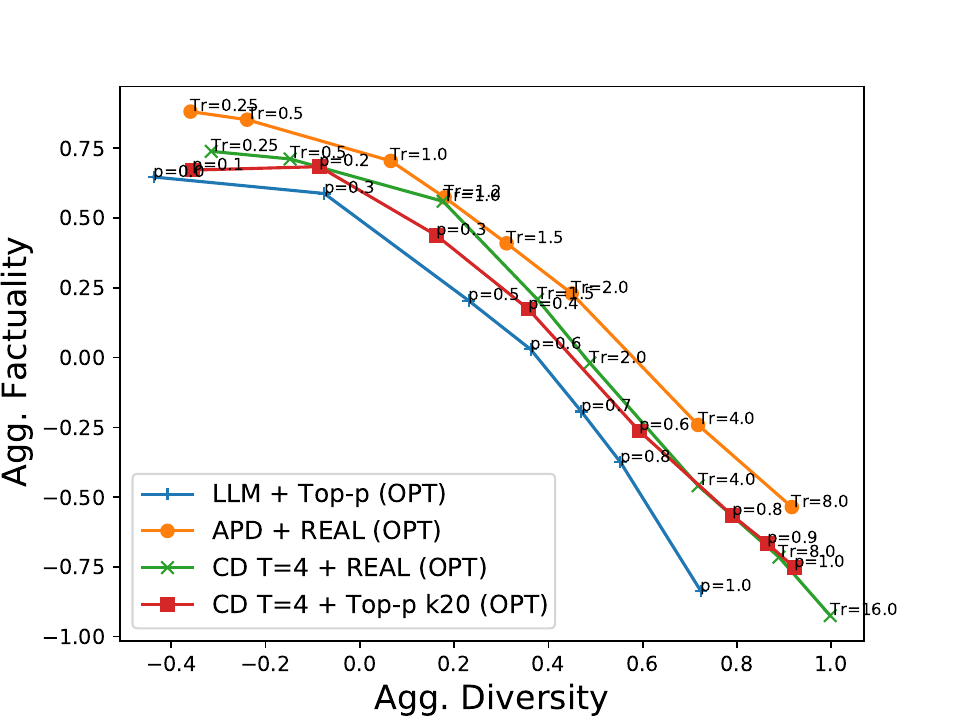}
  \caption{With REAL Sampling (OPT)}
  \label{fig:opt_real}
\end{subfigure}%
\begin{subfigure}{.33\textwidth}
  \centering
  \includegraphics[width=1\linewidth]{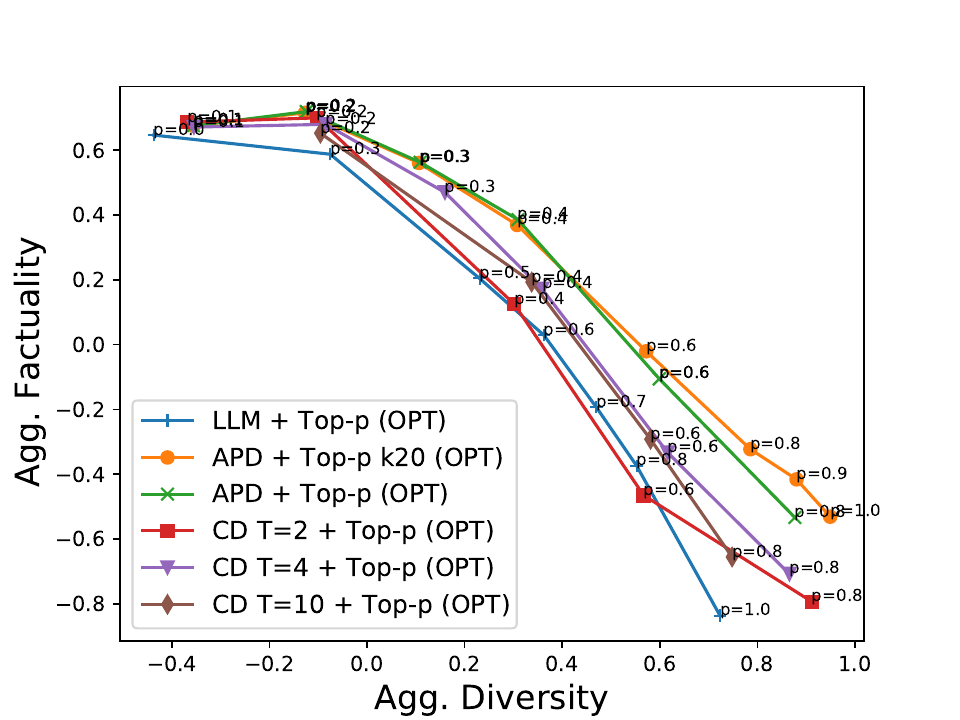}
  \caption{Temperature Tuning of CD (OPT)}
  \label{fig:opt_cd_temp}
\end{subfigure}
\begin{subfigure}{.33\textwidth}
  \centering
  \includegraphics[width=1\linewidth]{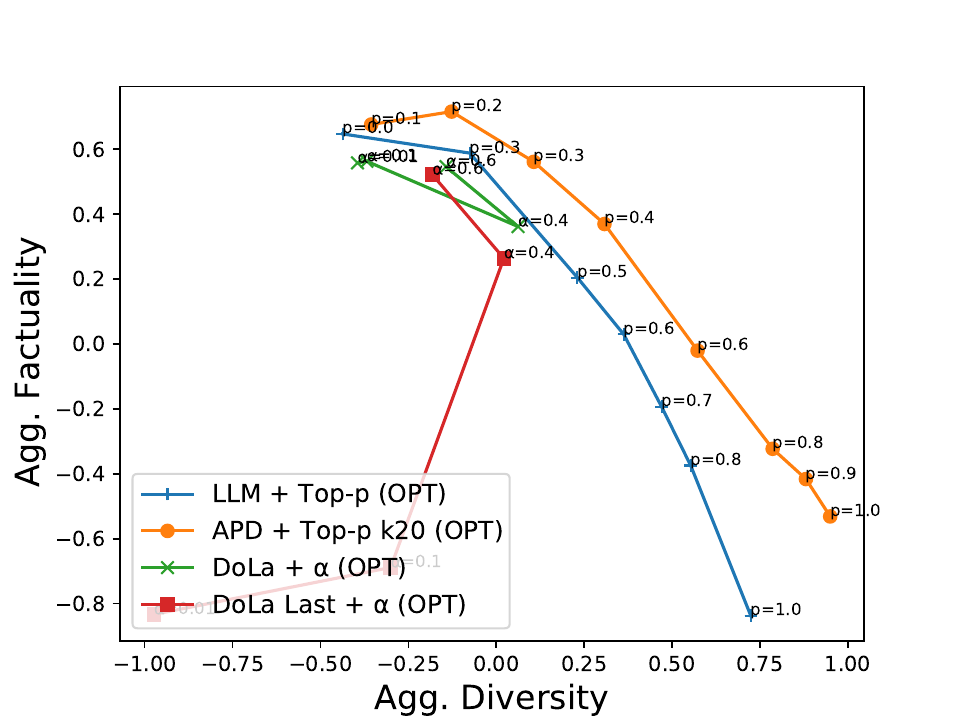}
  \caption{Layer Tuning of DoLa (OPT)}
  \label{fig:opt_dola}
\end{subfigure}%
\begin{subfigure}{.33\textwidth}
  \centering
  \includegraphics[width=1\linewidth]{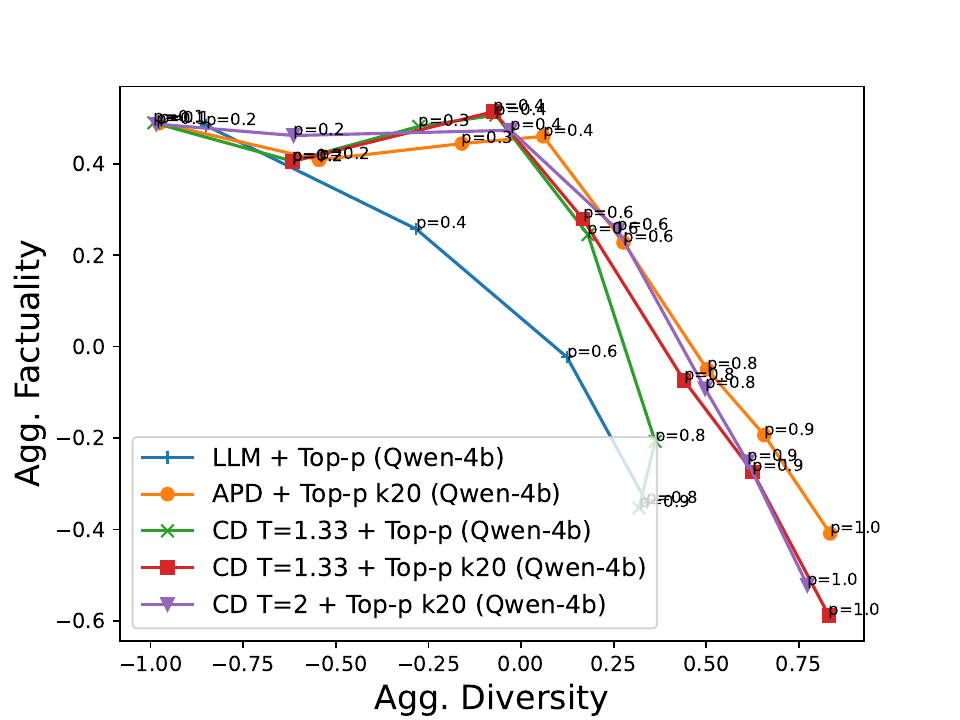}
  \caption{Ours v.s. CD (Qwen)}
  \label{fig:qwen_cd}
\end{subfigure}%
\begin{subfigure}{.33\textwidth}
  \centering
  \includegraphics[width=1\linewidth]{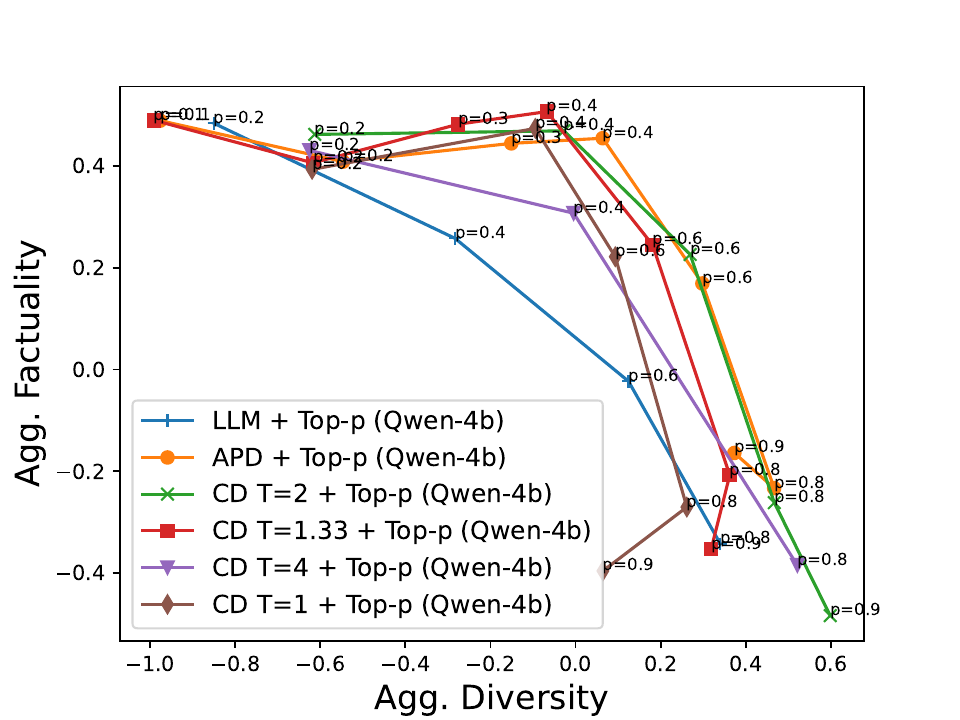}
  \caption{Temperature Tuning of CD (Qwen)}
  \label{fig:qwen_cd_temp}
\end{subfigure}
\caption{\textsc{FactualityPrompts} results. The x-axis is the diversity and y-axis is the factuality metrics from \citet{REAL}. The curves closer to the upper right corner are better. }

\label{fig:comp_agg_gen}
\end{figure*}

\begin{figure*}[t!]
\centering
\begin{subfigure}{.33\textwidth}
  \centering
  \includegraphics[width=1\linewidth]{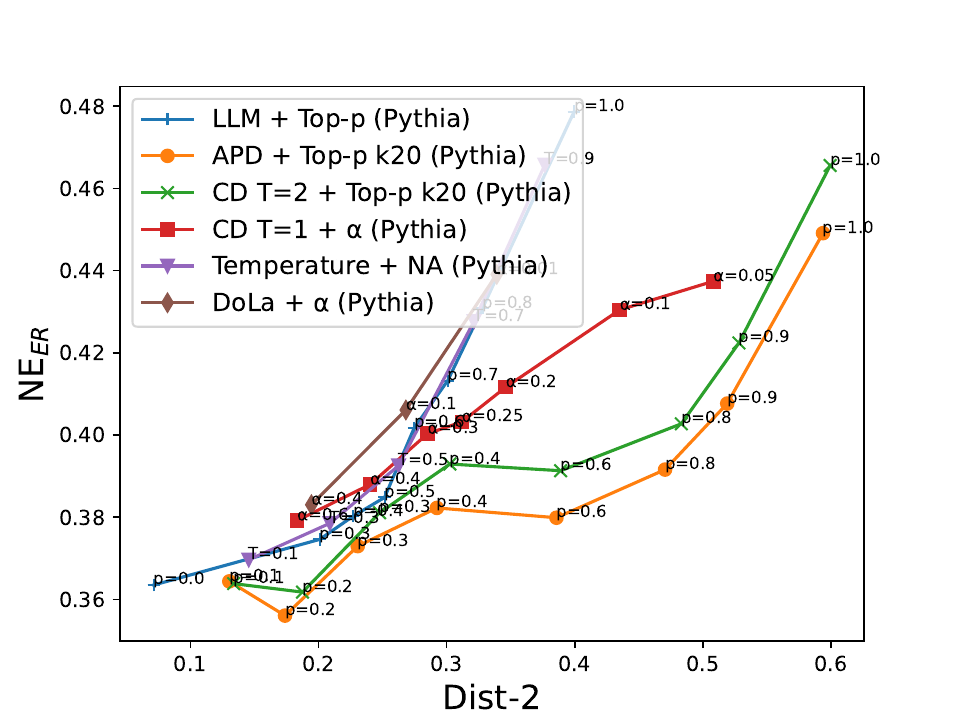}
  \caption{Ours v.s. SOTA (Pythia)}
\end{subfigure}%
\begin{subfigure}{.33\textwidth}
  \centering
  \includegraphics[width=1\linewidth]{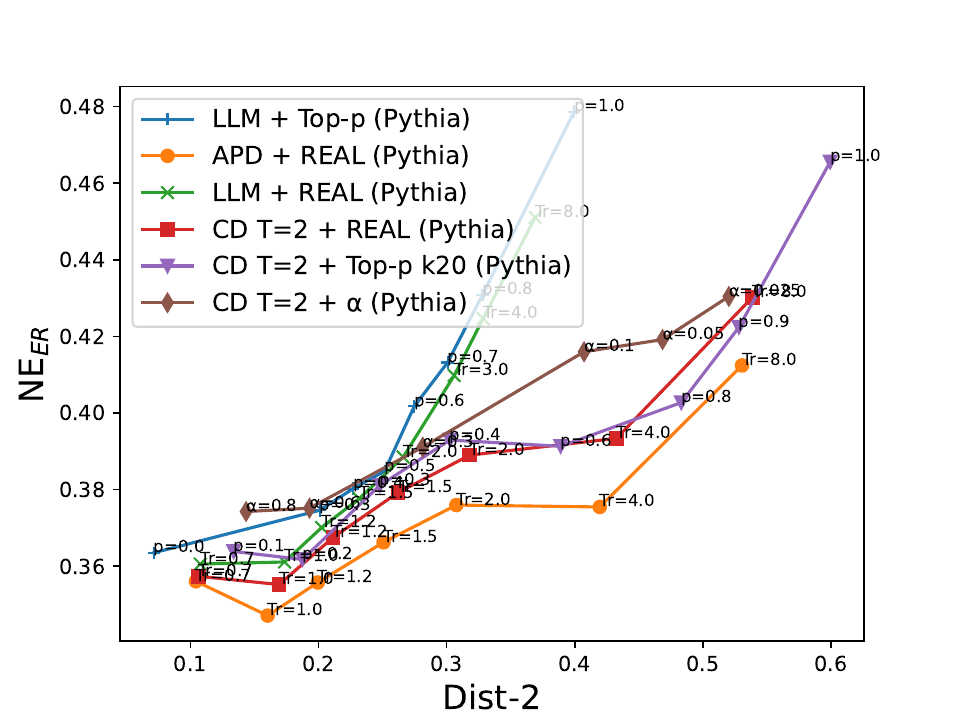}
  \caption{With REAL Sampling (Pythia)}
\end{subfigure}
\begin{subfigure}{.33\textwidth}
  \centering
  \includegraphics[width=1\linewidth]{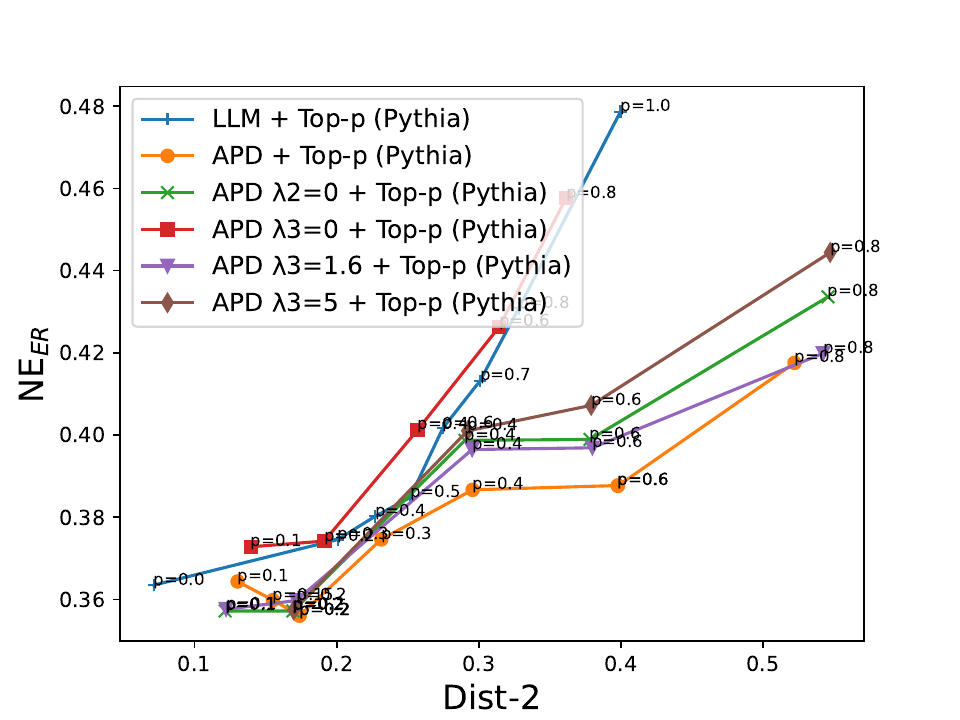}
  \caption{Loss Term Ablation (Pythia)}
\end{subfigure}%
\begin{subfigure}{.33\textwidth}
  \centering
  \includegraphics[width=1\linewidth]{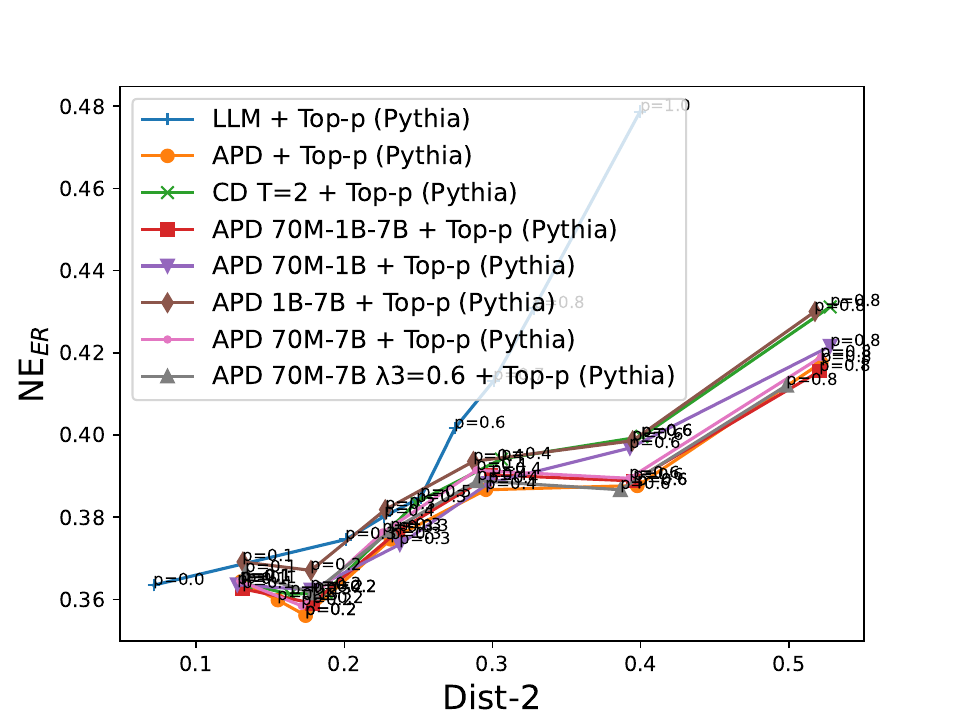}
  \caption{Mid-size LLM Ablation (Pythia)}
\end{subfigure}%
\begin{subfigure}{.33\textwidth}
  \centering
  \includegraphics[width=1\linewidth]{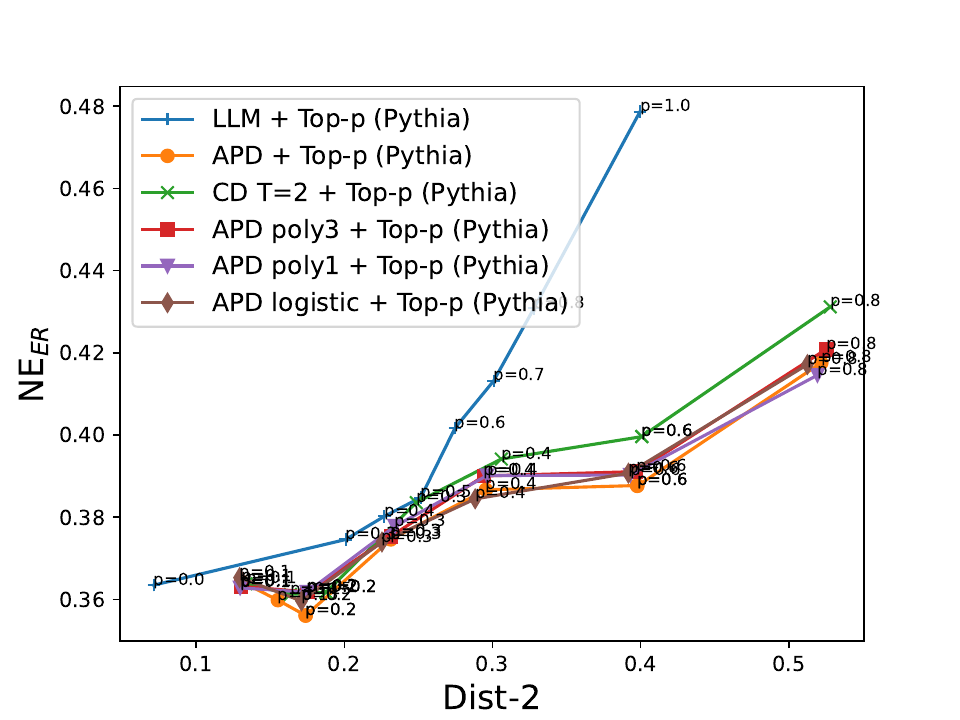}
  \caption{Function Ablation (Pythia)}
\end{subfigure}
\begin{subfigure}{.33\textwidth}
  \centering
  \includegraphics[width=1\linewidth]{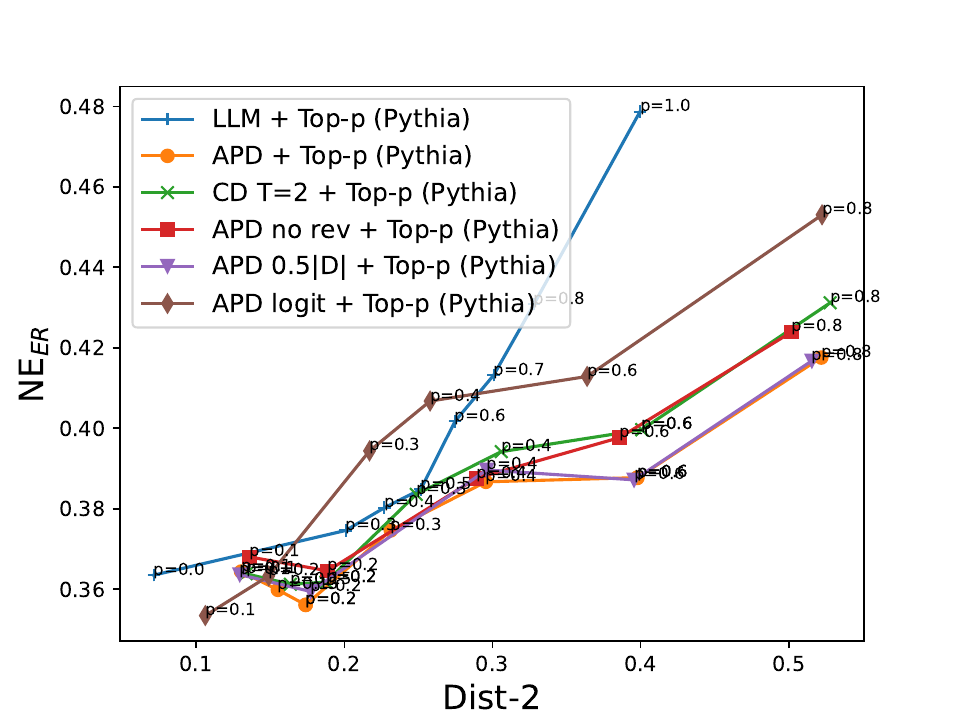}
  \caption{Other Ablation (Pythia)}
\end{subfigure}%
\begin{subfigure}{.33\textwidth}
  \centering
  \includegraphics[width=1\linewidth]{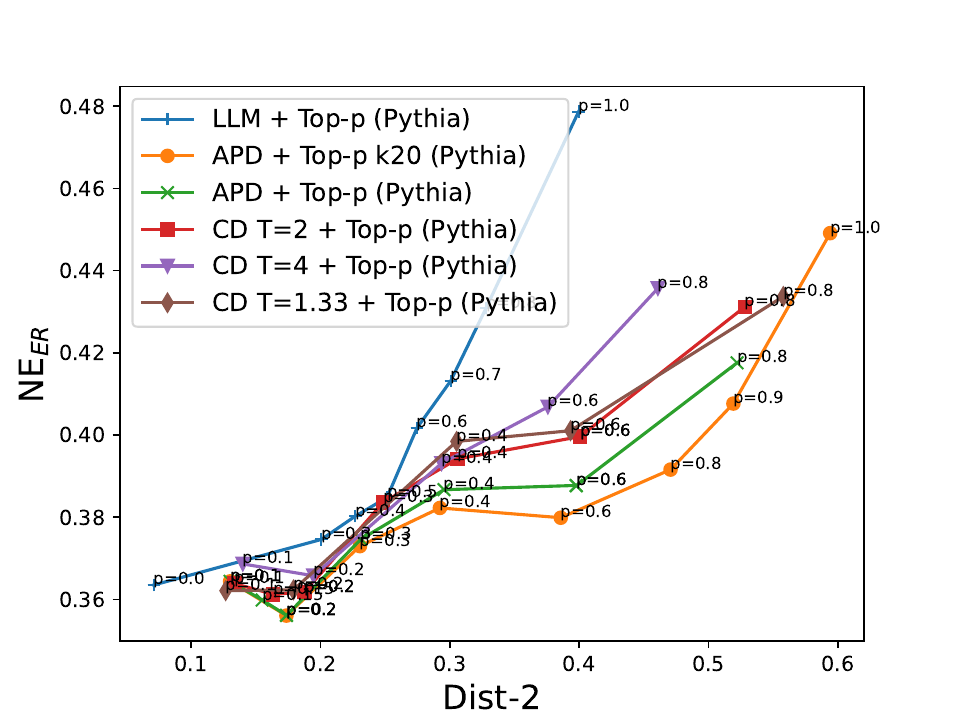}
  \caption{Temperature Tuning of CD (Pythia)}
\end{subfigure}%
\begin{subfigure}{.33\textwidth}
  \centering
  \includegraphics[width=1\linewidth]{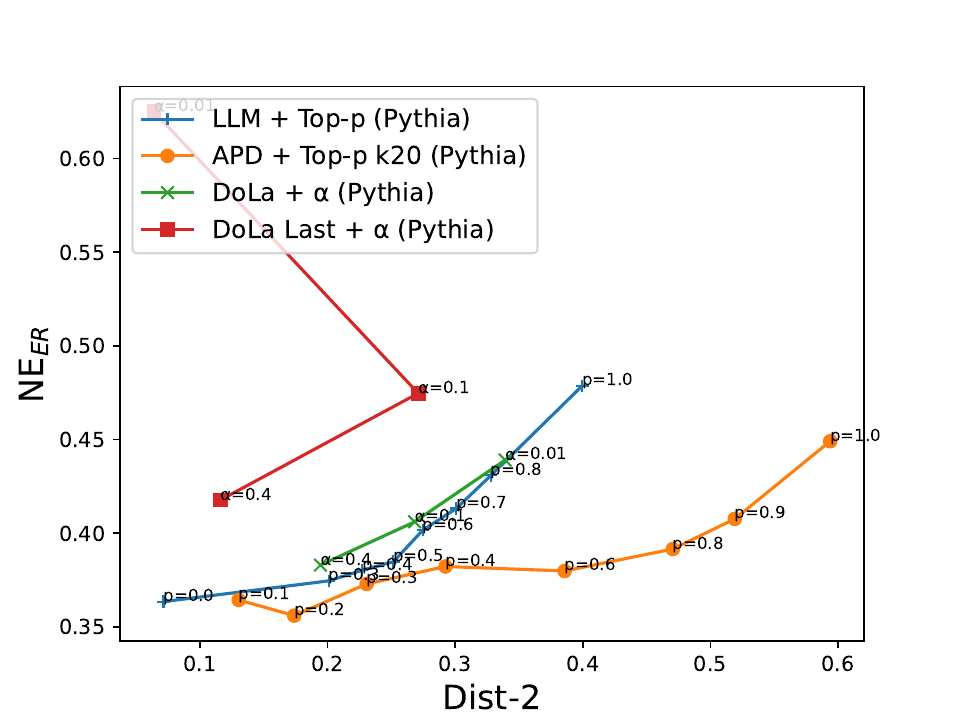}
  \caption{Layer Tuning of DoLa (Pythia)}
\end{subfigure}
\begin{subfigure}{.33\textwidth}
  \centering
  \includegraphics[width=1\linewidth]{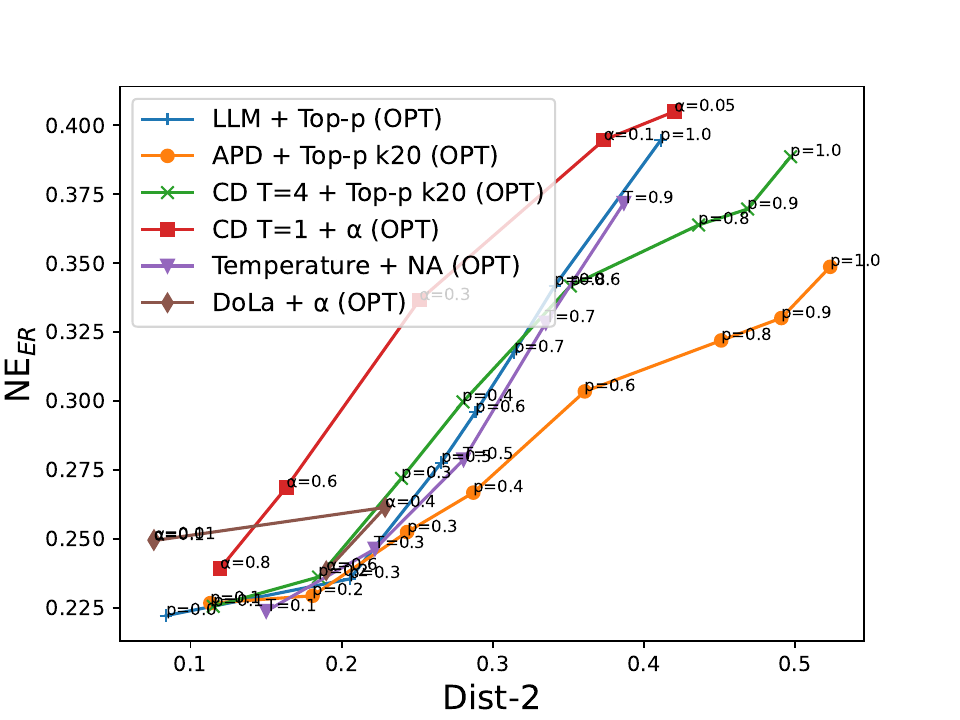}
  \caption{Ours v.s. SOTA (OPT)}
\end{subfigure}%
\begin{subfigure}{.33\textwidth}
  \centering
  \includegraphics[width=1\linewidth]{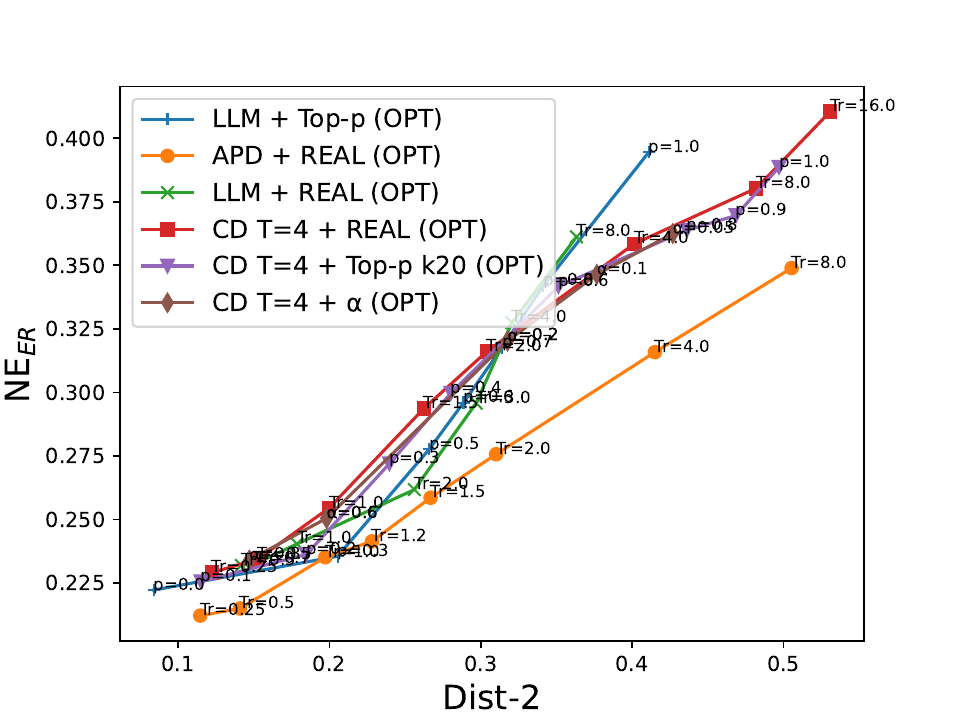}
  \caption{With REAL Sampling (OPT)}
\end{subfigure}%
\begin{subfigure}{.33\textwidth}
  \centering
  \includegraphics[width=1\linewidth]{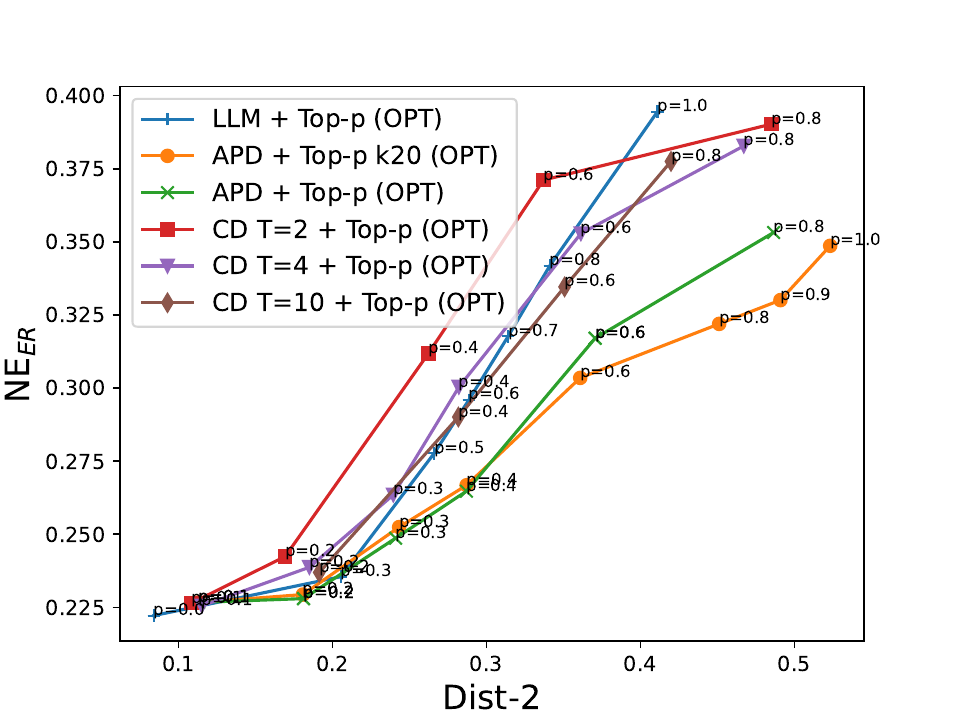}
  \caption{Temperature Tuning of CD (OPT)}
\end{subfigure}
\begin{subfigure}{.33\textwidth}
  \centering
  \includegraphics[width=1\linewidth]{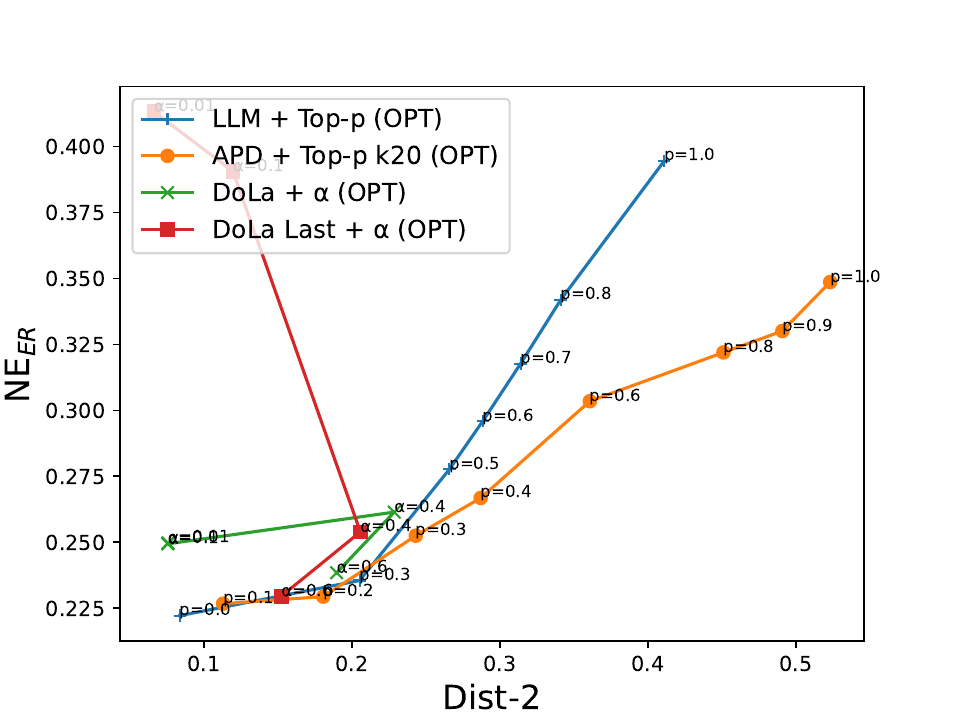}
  \caption{Layer Tuning of DoLa (OPT)}
\end{subfigure}%
\begin{subfigure}{.33\textwidth}
  \centering
  \includegraphics[width=1\linewidth]{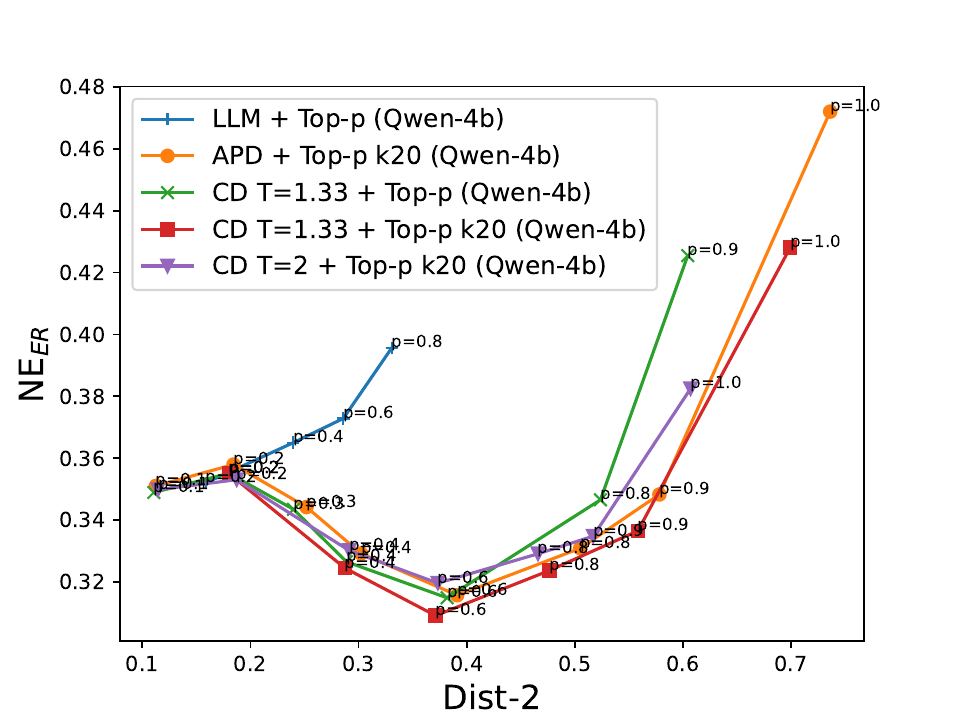}
  \caption{Ours v.s. CD (Qwen)}
\end{subfigure}%
\begin{subfigure}{.33\textwidth}
  \centering
  \includegraphics[width=1\linewidth]{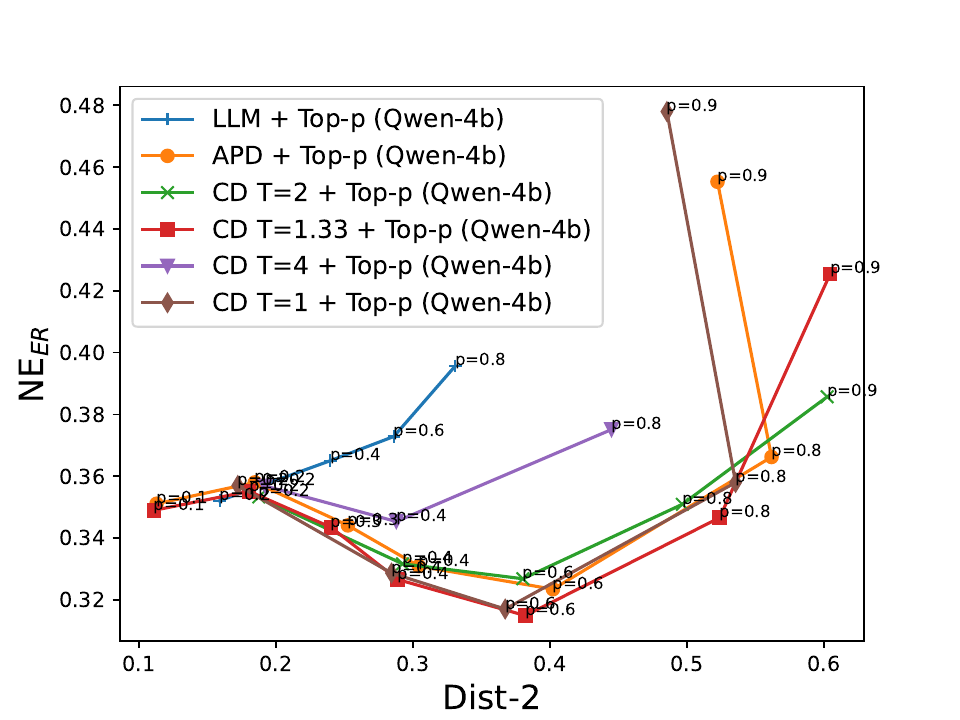}
  \caption{Temperature Tuning of CD (Qwen)}
\end{subfigure}
\caption{\textsc{FactualityPrompts} results using factual prompts. The curves closer to the lower right corner are better. The average standard error of NE$_{ER}$ is $0.0038$ and the maximal one is $0.0095$. }

\label{fig:comp_ne_dist_factual}
\end{figure*}

\begin{figure*}[t!]
\centering
\begin{subfigure}{.33\textwidth}
  \centering
  \includegraphics[width=1\linewidth]{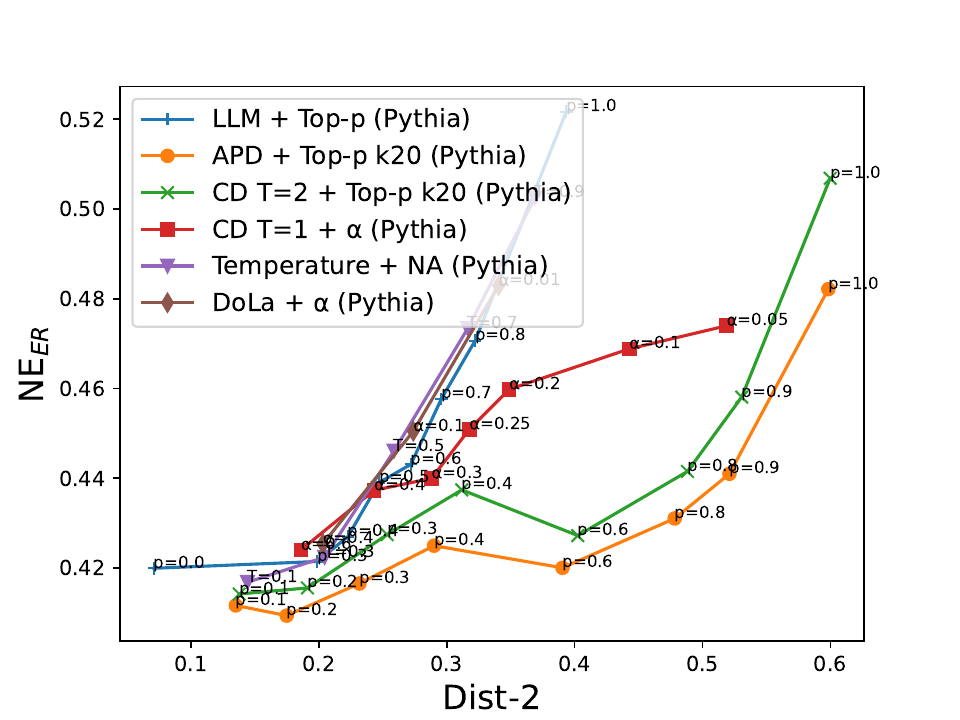}
  \caption{Ours v.s. SOTA (Pythia)}
\end{subfigure}%
\begin{subfigure}{.33\textwidth}
  \centering
  \includegraphics[width=1\linewidth]{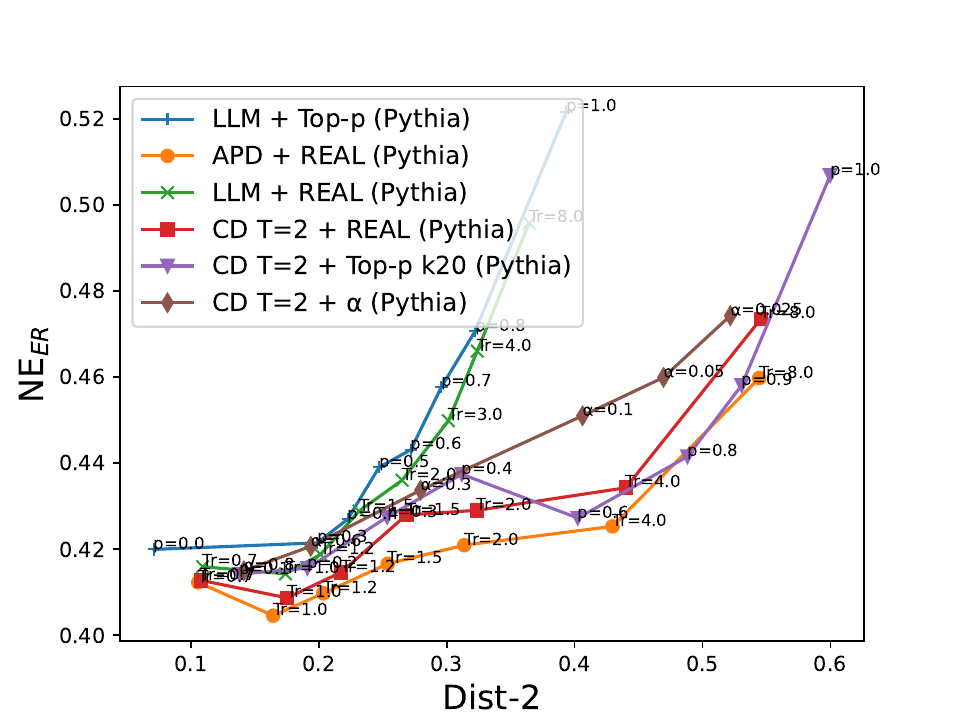}
  \caption{With REAL Sampling (Pythia)}
\end{subfigure}
\begin{subfigure}{.33\textwidth}
  \centering
  \includegraphics[width=1\linewidth]{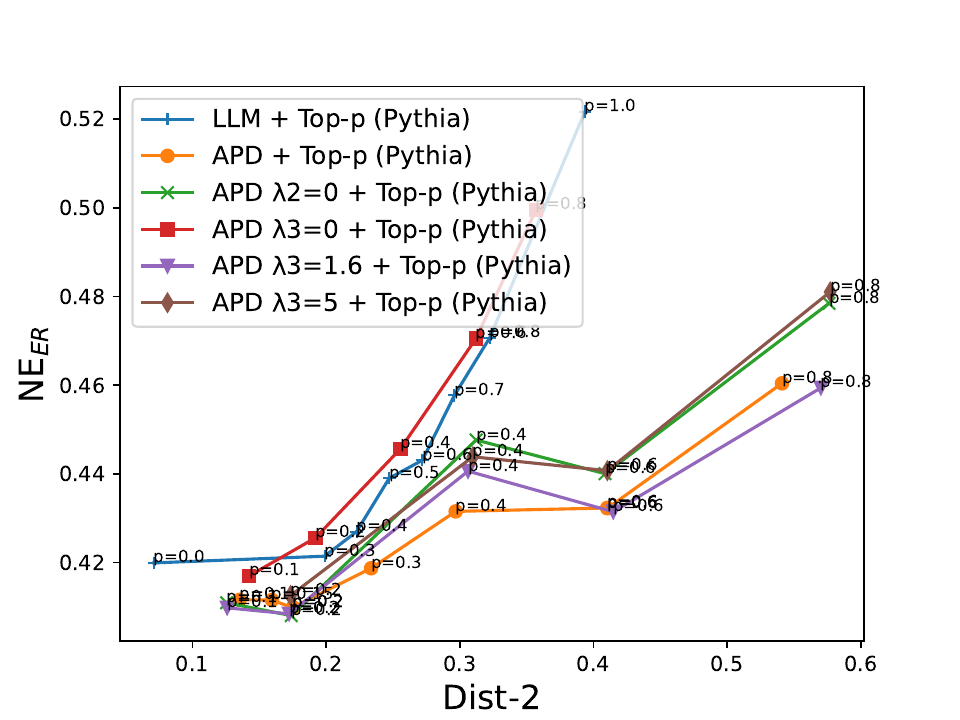}
  \caption{Loss Term Ablation (Pythia)}
\end{subfigure}%
\begin{subfigure}{.33\textwidth}
  \centering
  \includegraphics[width=1\linewidth]{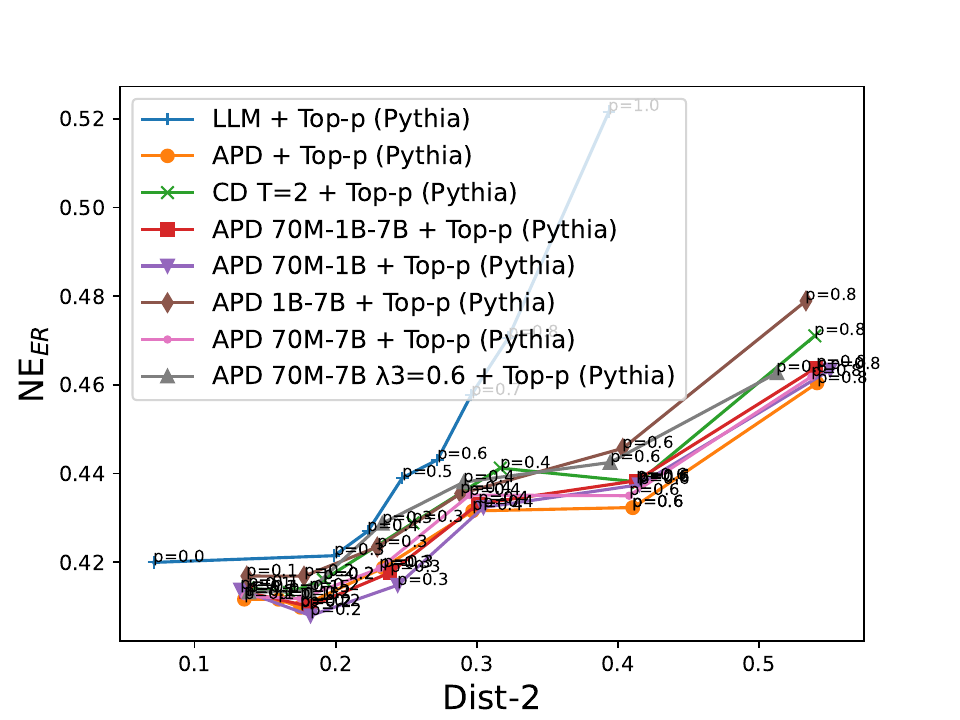}
  \caption{Mid-size LLM Ablation (Pythia)}
\end{subfigure}%
\begin{subfigure}{.33\textwidth}
  \centering
  \includegraphics[width=1\linewidth]{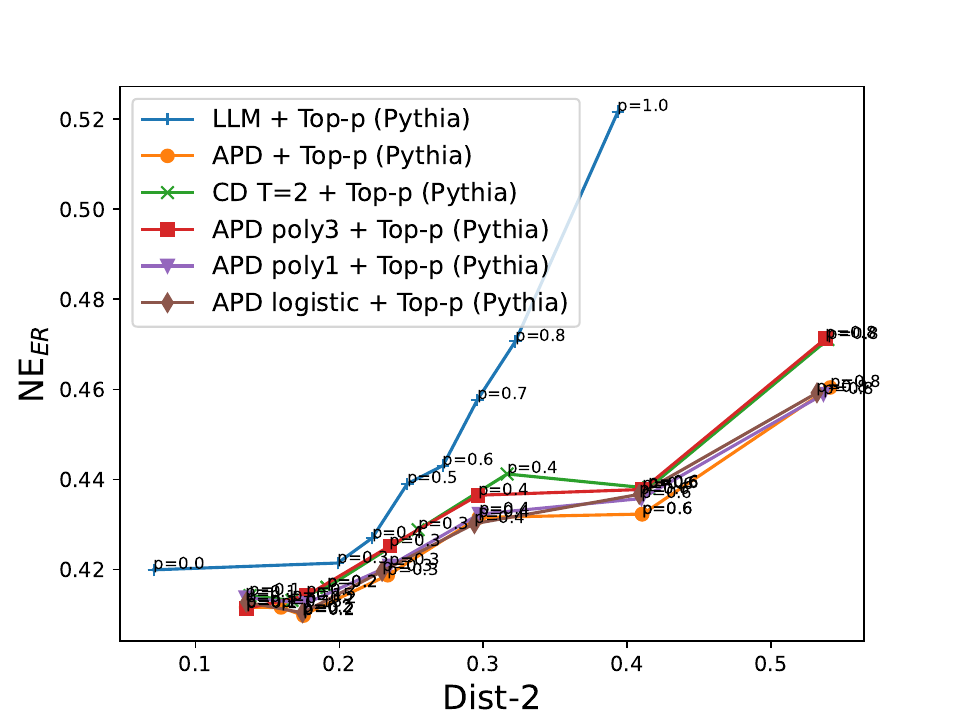}
  \caption{Function Ablation (Pythia)}
\end{subfigure}
\begin{subfigure}{.33\textwidth}
  \centering
  \includegraphics[width=1\linewidth]{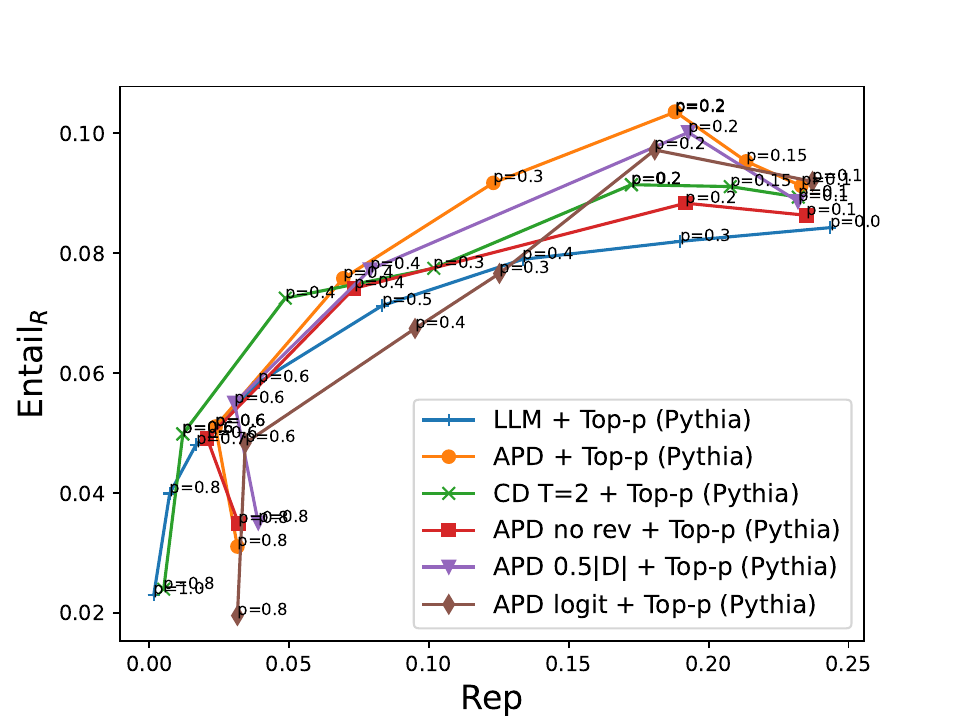}
  \caption{Other Ablation (Pythia)}
\end{subfigure}%
\begin{subfigure}{.33\textwidth}
  \centering
  \includegraphics[width=1\linewidth]{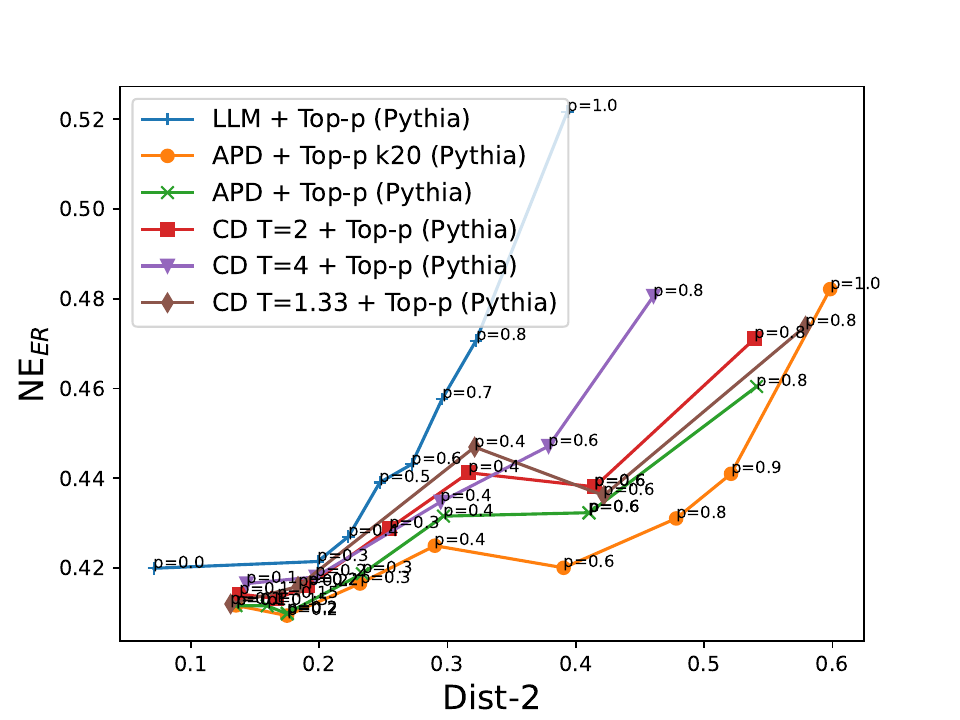}
  \caption{Temperature Tuning of CD (Pythia)}
\end{subfigure}%
\begin{subfigure}{.33\textwidth}
  \centering
  \includegraphics[width=1\linewidth]{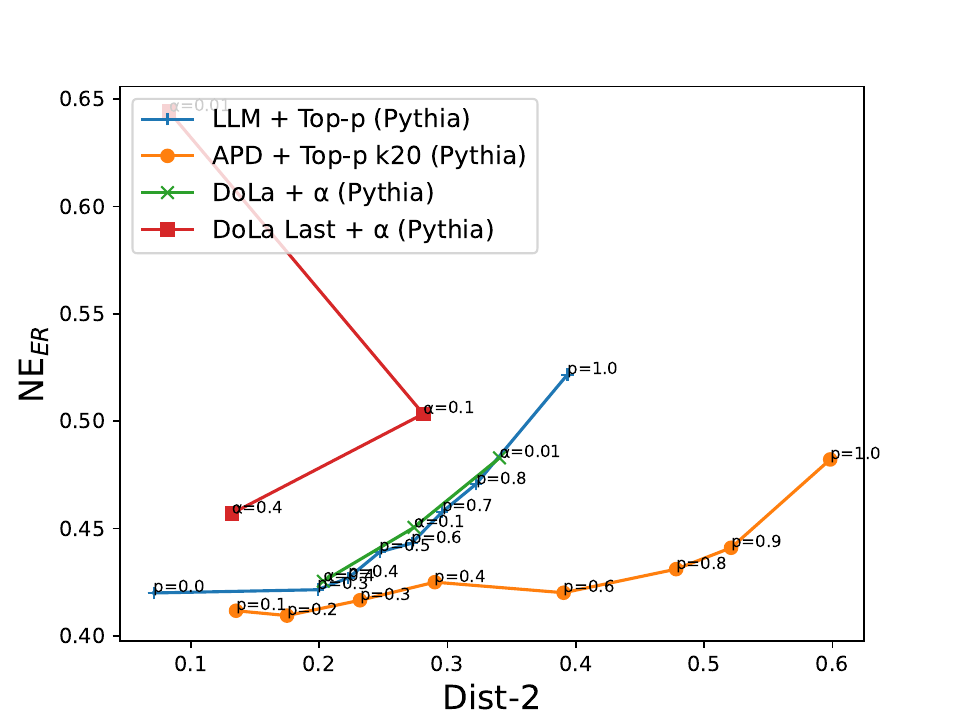}
  \caption{Layer Tuning of DoLa (Pythia)}
\end{subfigure}
\begin{subfigure}{.33\textwidth}
  \centering
  \includegraphics[width=1\linewidth]{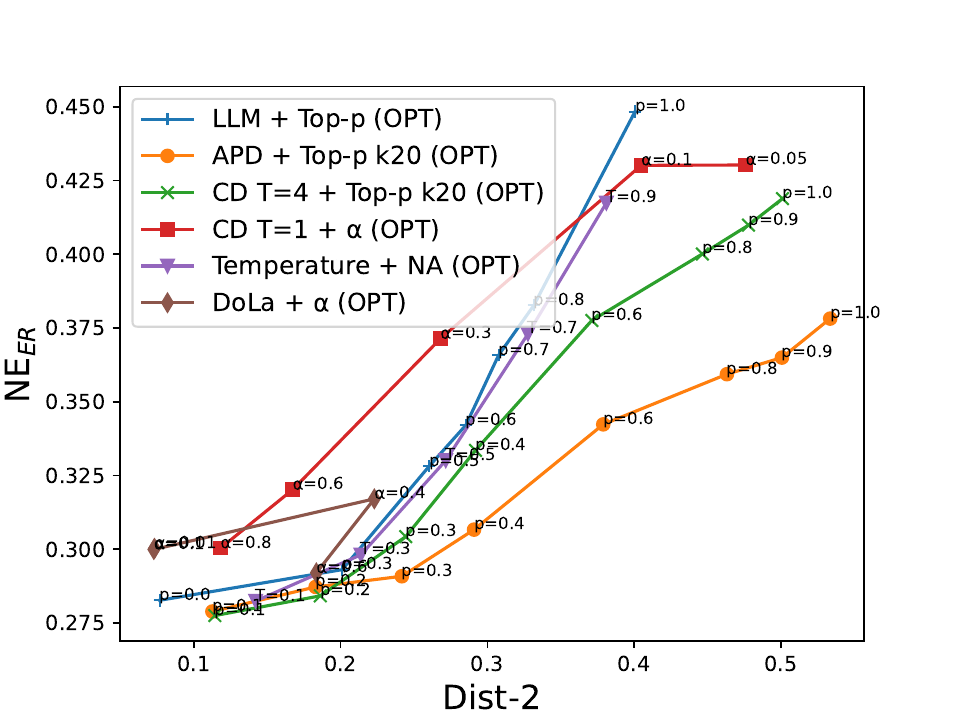}
  \caption{Ours v.s. SOTA (OPT)}
\end{subfigure}%
\begin{subfigure}{.33\textwidth}
  \centering
  \includegraphics[width=1\linewidth]{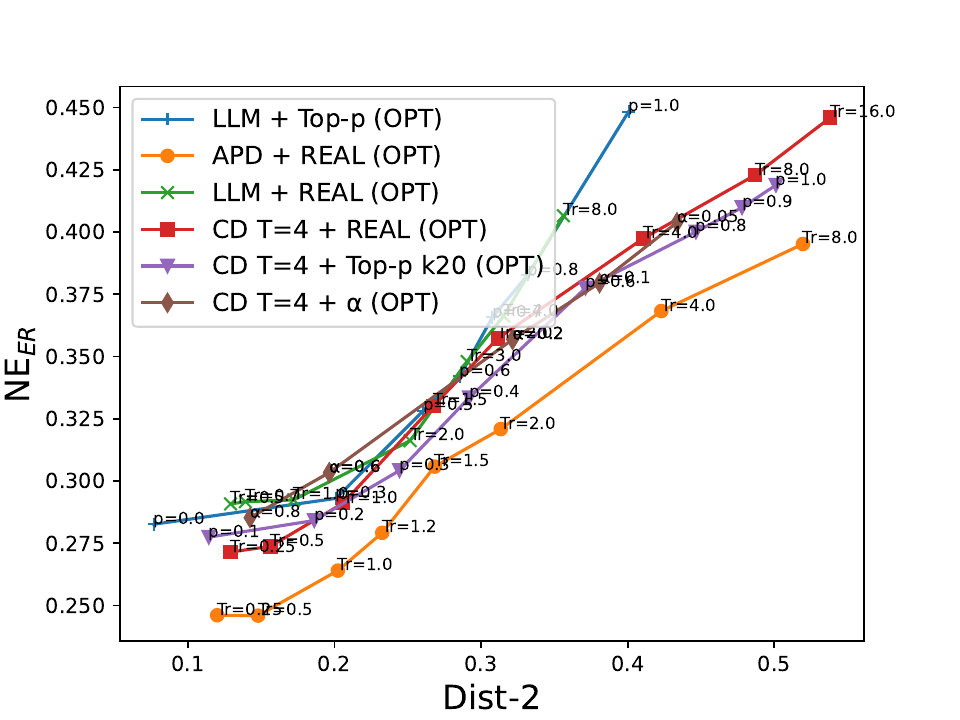}
  \caption{With REAL Sampling (OPT)}
\end{subfigure}
\begin{subfigure}{.33\textwidth}
  \centering
  \includegraphics[width=1\linewidth]{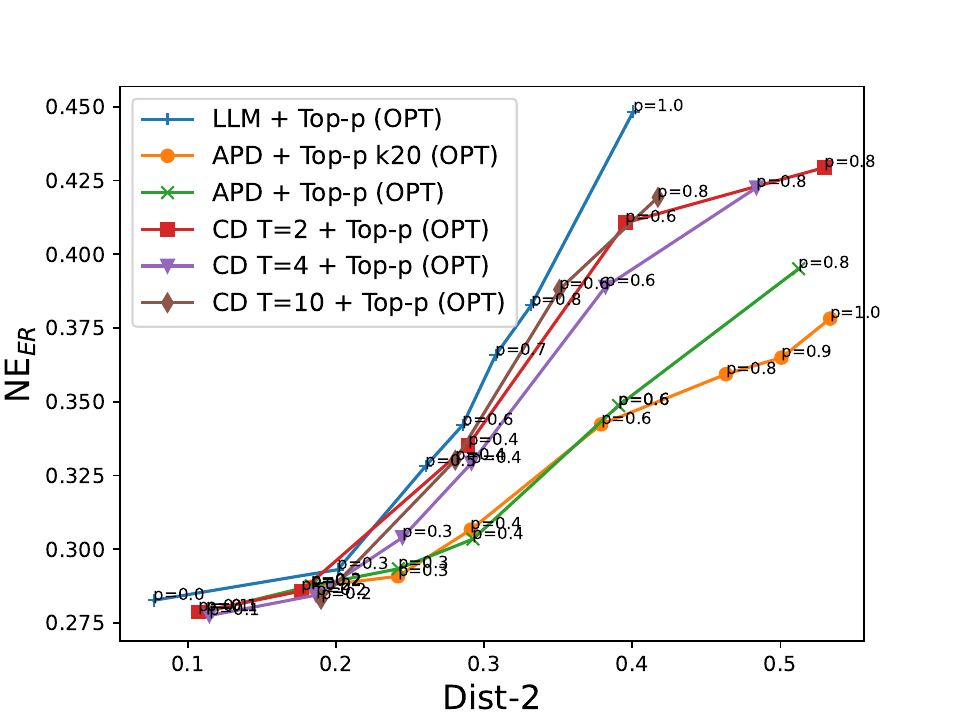}
  \caption{Temperature Tuning of CD (OPT)}
\end{subfigure}
\begin{subfigure}{.33\textwidth}
  \centering
  \includegraphics[width=1\linewidth]{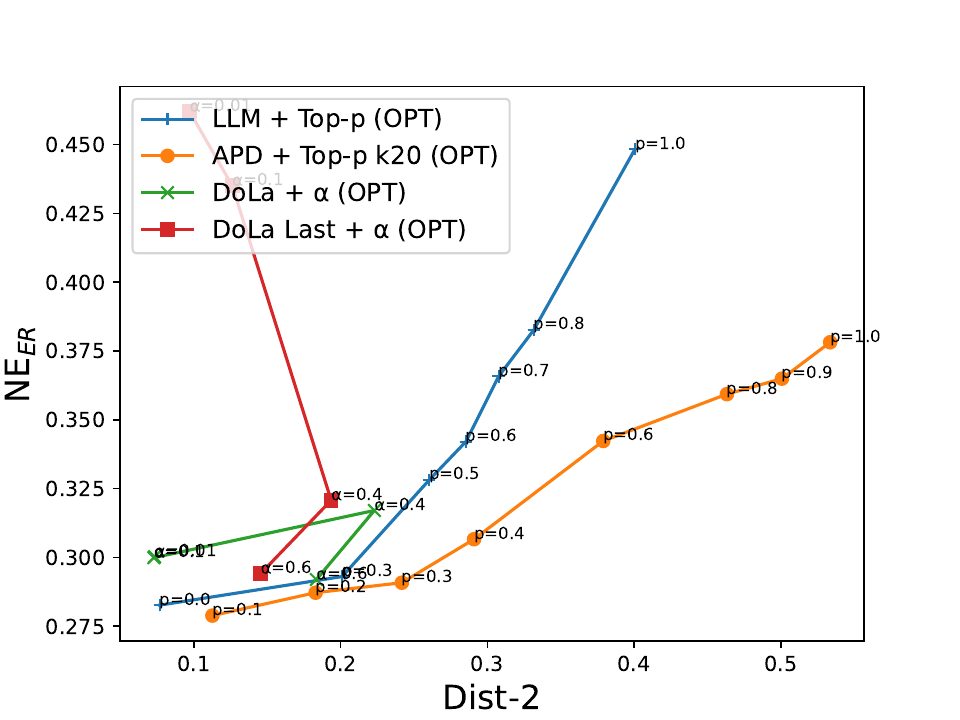}
  \caption{Layer Tuning of DoLa (OPT)}
\end{subfigure}
\begin{subfigure}{.33\textwidth}
  \centering
  \includegraphics[width=1\linewidth]{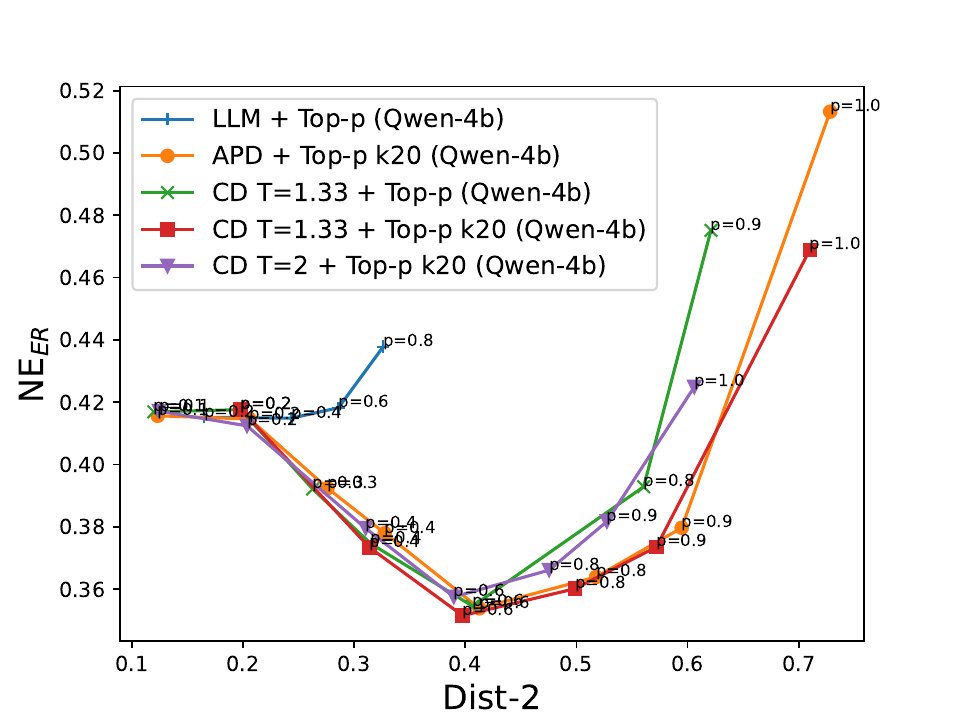}
  \caption{Ours v.s. CD (Qwen)}
\end{subfigure}%
\begin{subfigure}{.33\textwidth}
  \centering
  \includegraphics[width=1\linewidth]{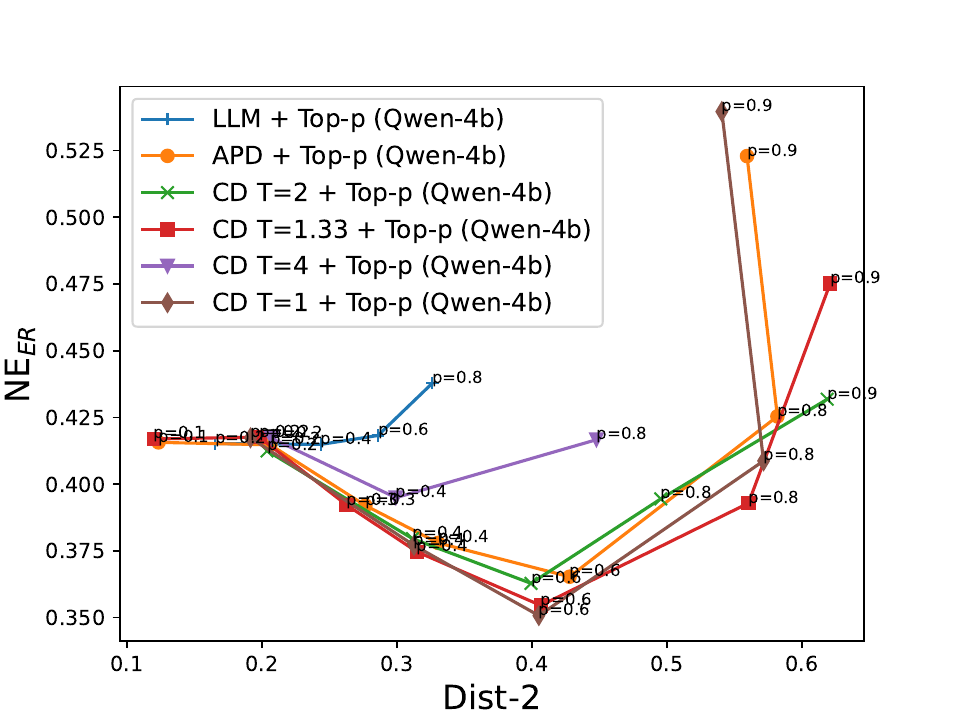}
  \caption{Temperature Tuning of CD (Qwen)}
\end{subfigure}
\caption{\textsc{FactualityPrompts} results using nonfactual prompts. The curves closer to the lower right corner are better. The average standard error of NE$_{ER}$ is $0.0040$ and the maximal one is $0.0111$. }

\label{fig:comp_ne_dist_nonfactual}
\end{figure*}

\begin{figure*}[t!]
\centering
\begin{subfigure}{.33\textwidth}
  \centering
  \includegraphics[width=1\linewidth]{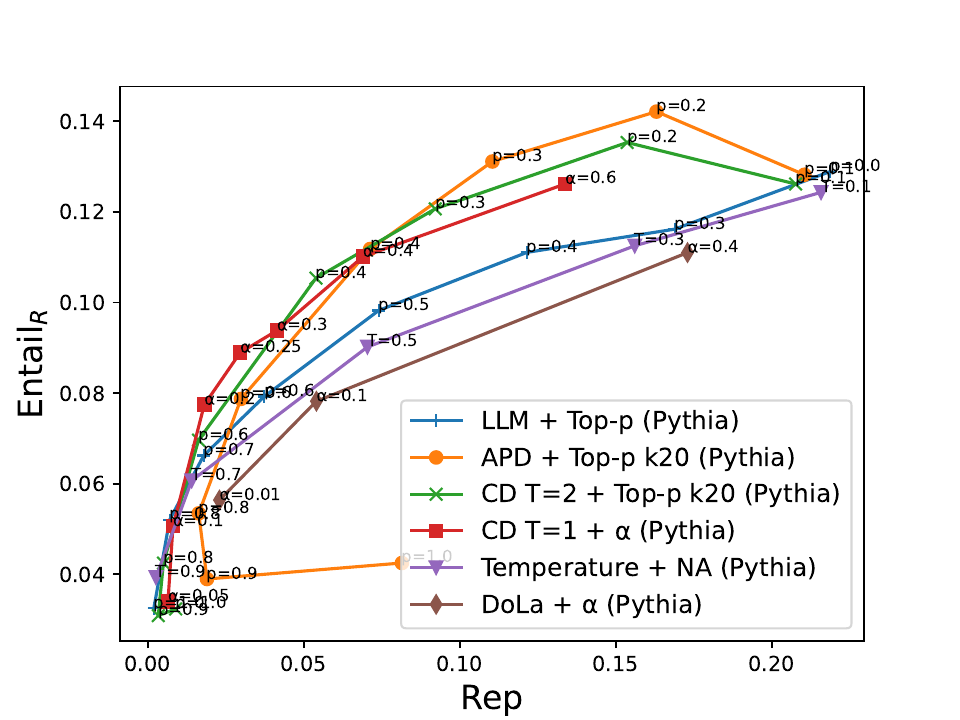}
  \caption{Ours v.s. SOTA (Pythia)}
\end{subfigure}%
\begin{subfigure}{.33\textwidth}
  \centering
  \includegraphics[width=1\linewidth]{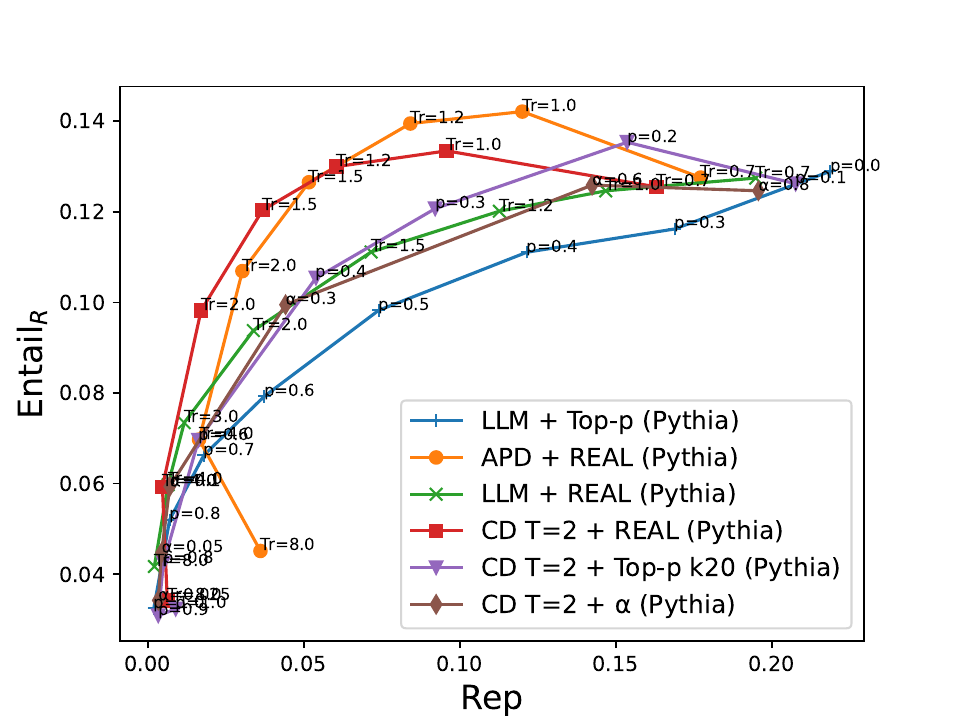}
  \caption{With REAL Sampling (Pythia)}
\end{subfigure}
\begin{subfigure}{.33\textwidth}
  \centering
  \includegraphics[width=1\linewidth]{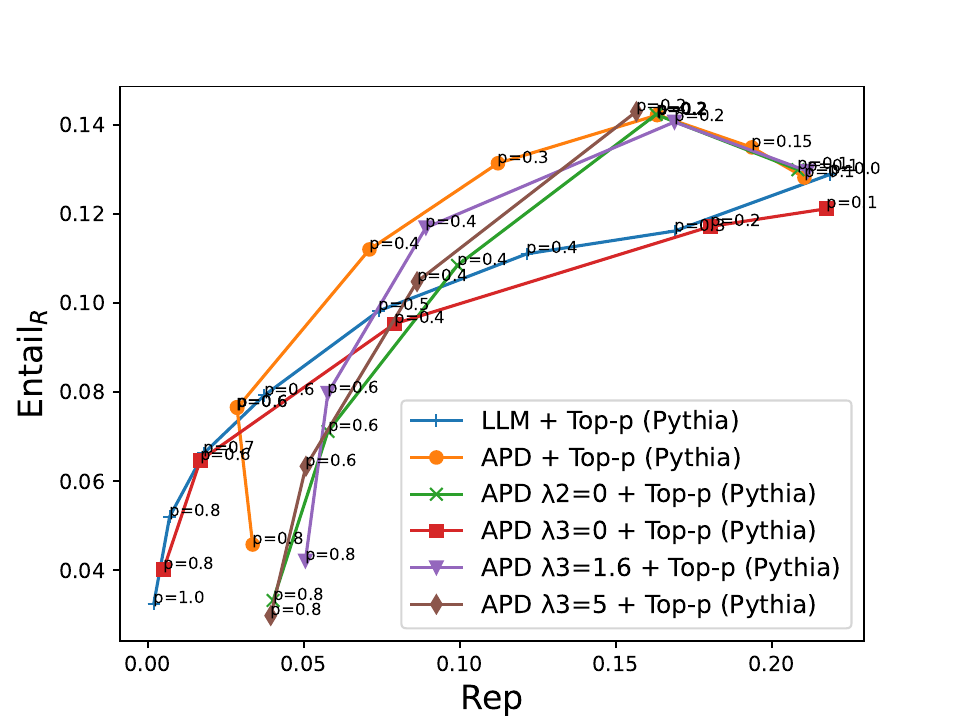}
  \caption{Loss Term Ablation (Pythia)}
\end{subfigure}%
\begin{subfigure}{.33\textwidth}
  \centering
  \includegraphics[width=1\linewidth]{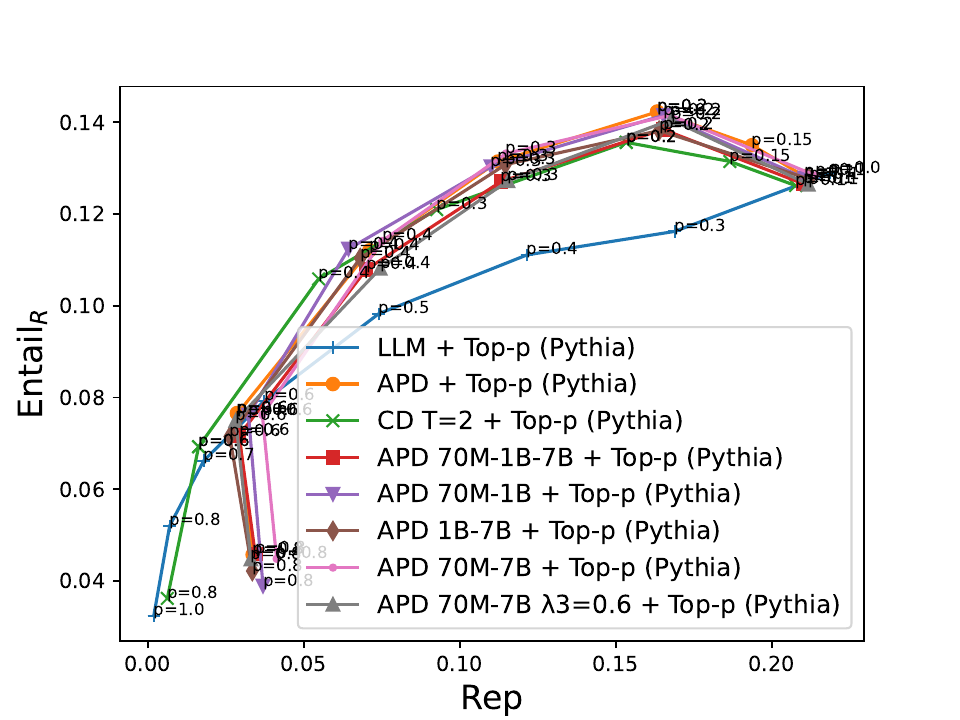}
  \caption{Mid-size LLM Ablation (Pythia)}
\end{subfigure}%
\begin{subfigure}{.33\textwidth}
  \centering
  \includegraphics[width=1\linewidth]{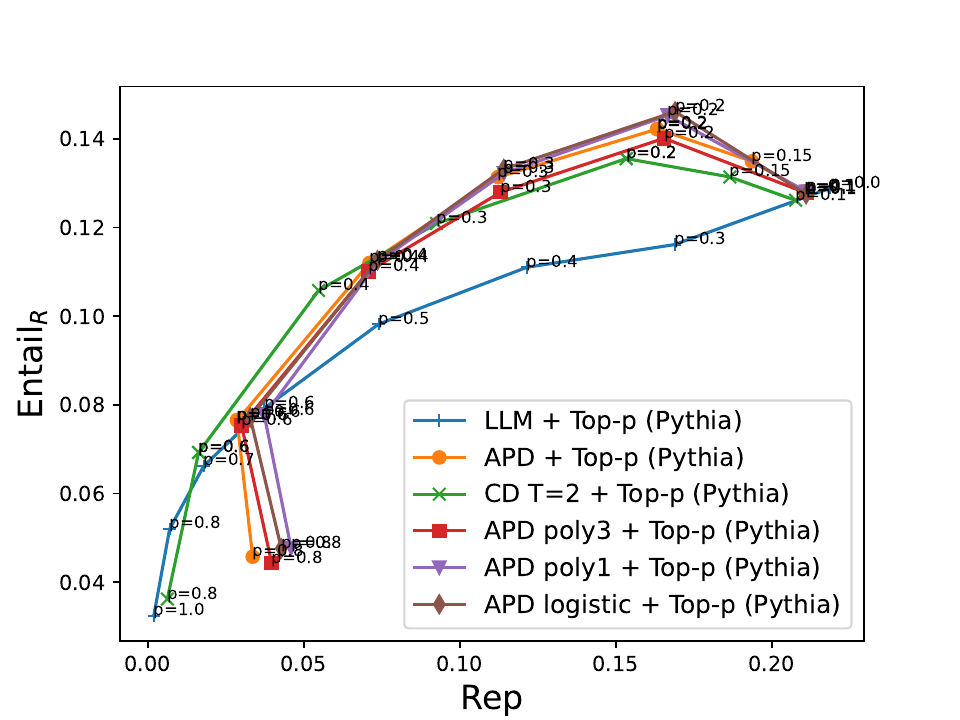}
  \caption{Function Ablation (Pythia)}
\end{subfigure}
\begin{subfigure}{.33\textwidth}
  \centering
  \includegraphics[width=1\linewidth]{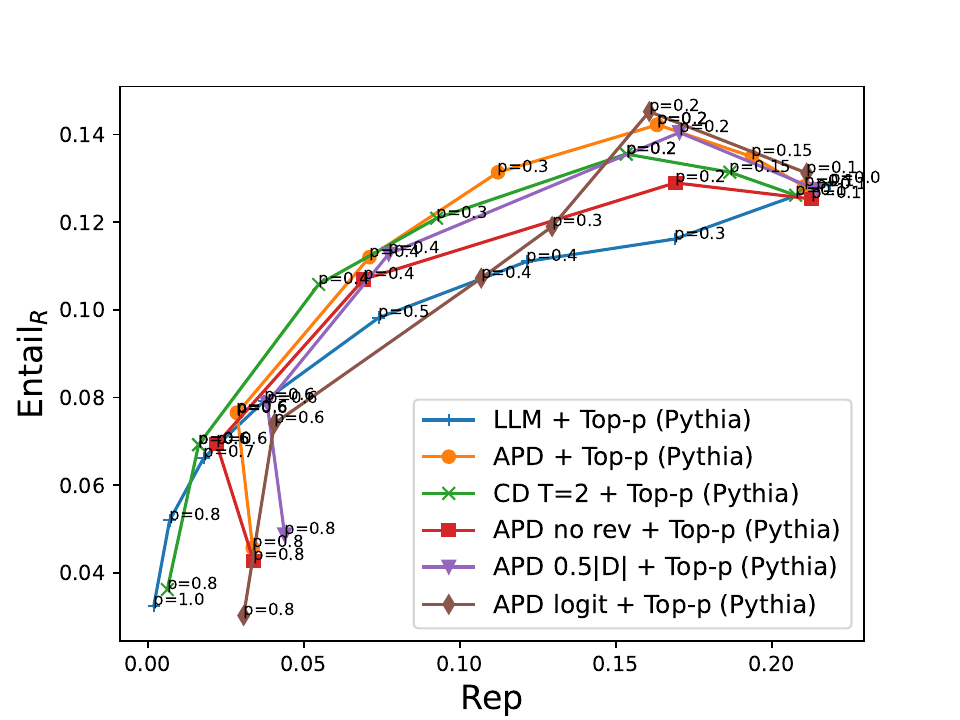}
  \caption{Other Ablation (Pythia)}
\end{subfigure}%
\begin{subfigure}{.33\textwidth}
  \centering
  \includegraphics[width=1\linewidth]{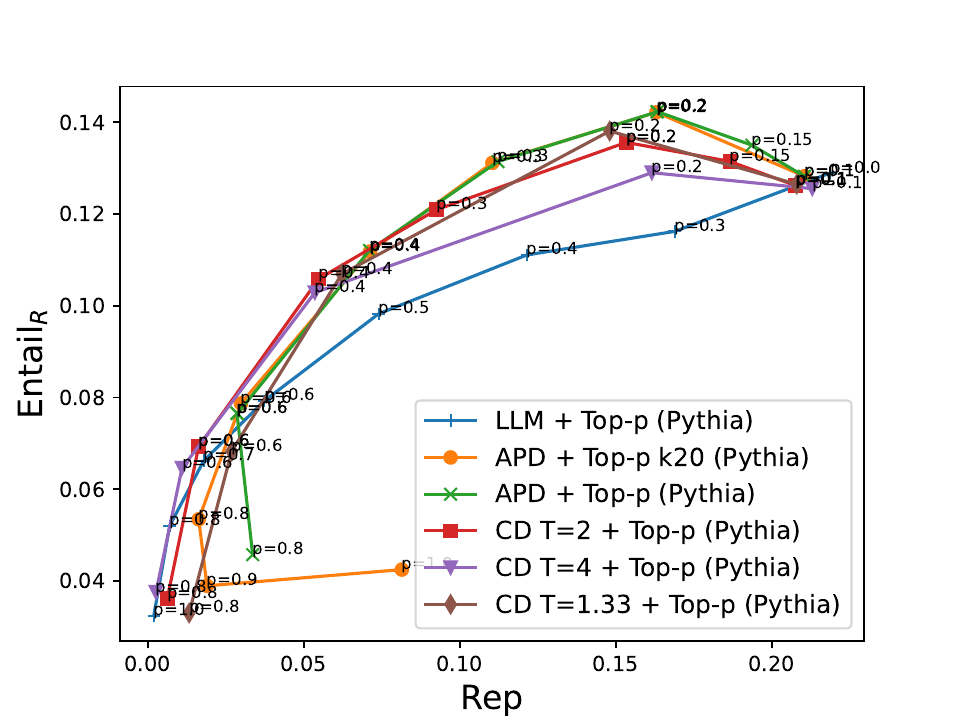}
  \caption{Temperature Tuning of CD (Pythia)}
\end{subfigure}%
\begin{subfigure}{.33\textwidth}
  \centering
  \includegraphics[width=1\linewidth]{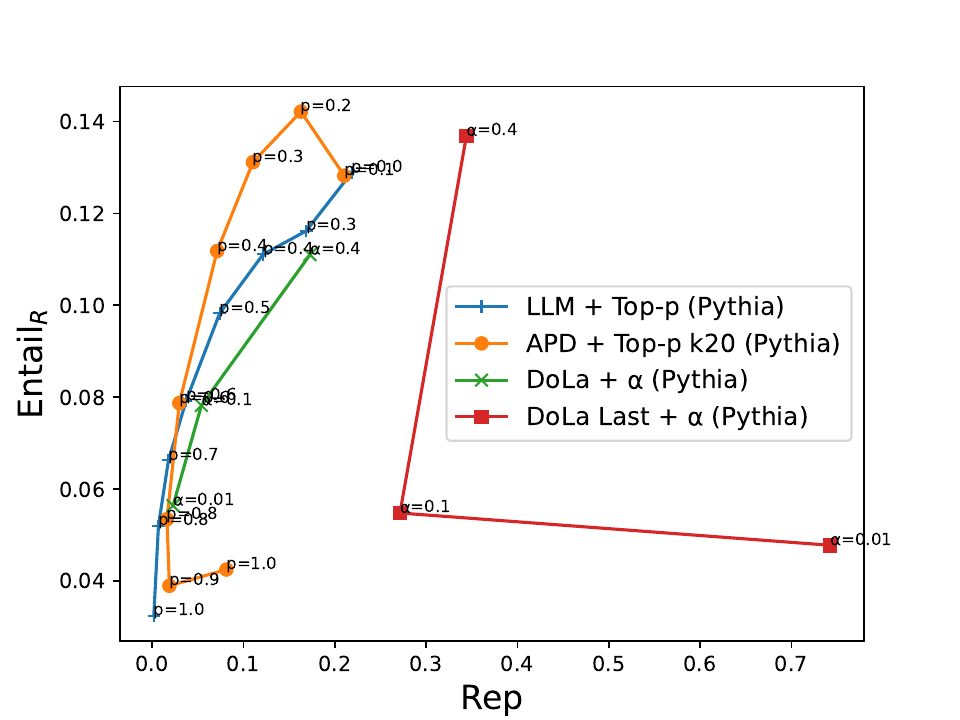}
  \caption{Layer Tuning of DoLa (Pythia)}
\end{subfigure}
\begin{subfigure}{.33\textwidth}
  \centering
  \includegraphics[width=1\linewidth]{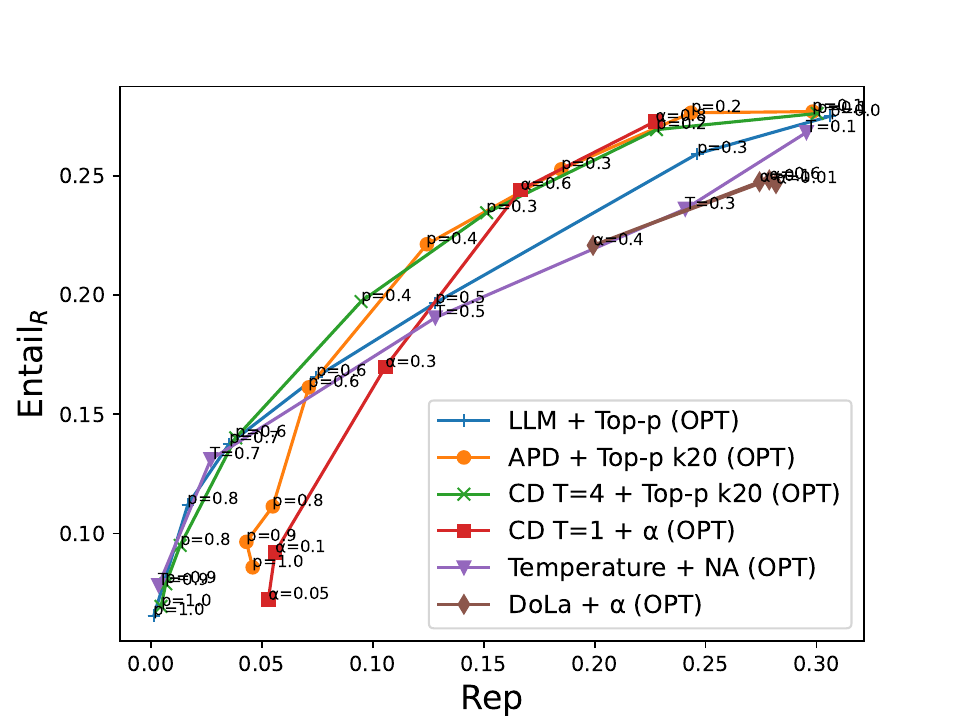}
  \caption{Ours v.s. SOTA (OPT)}
\end{subfigure}%
\begin{subfigure}{.33\textwidth}
  \centering
  \includegraphics[width=1\linewidth]{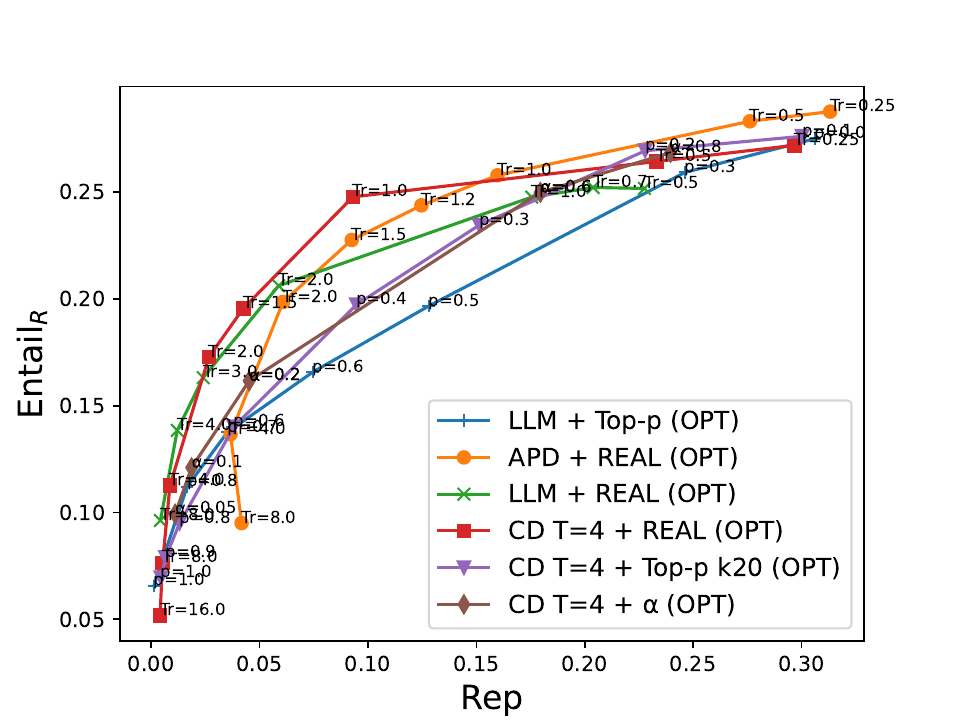}
  \caption{With REAL Sampling (OPT)}
\end{subfigure}
\begin{subfigure}{.33\textwidth}
  \centering
  \includegraphics[width=1\linewidth]{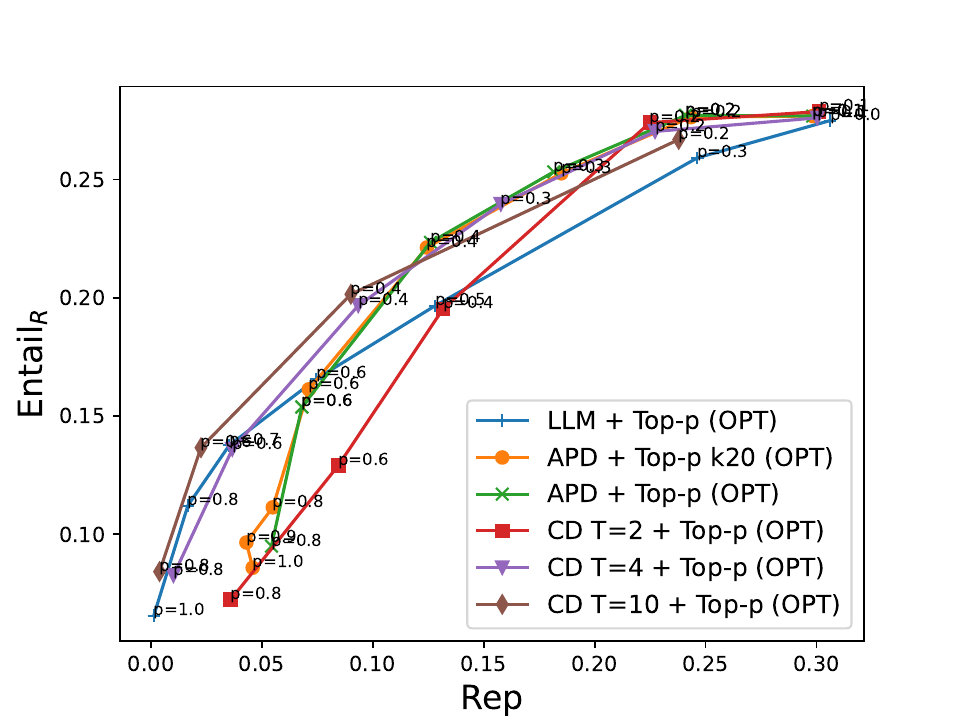}
  \caption{Temperature Tuning of CD (OPT)}
\end{subfigure}
\begin{subfigure}{.33\textwidth}
  \centering
  \includegraphics[width=1\linewidth]{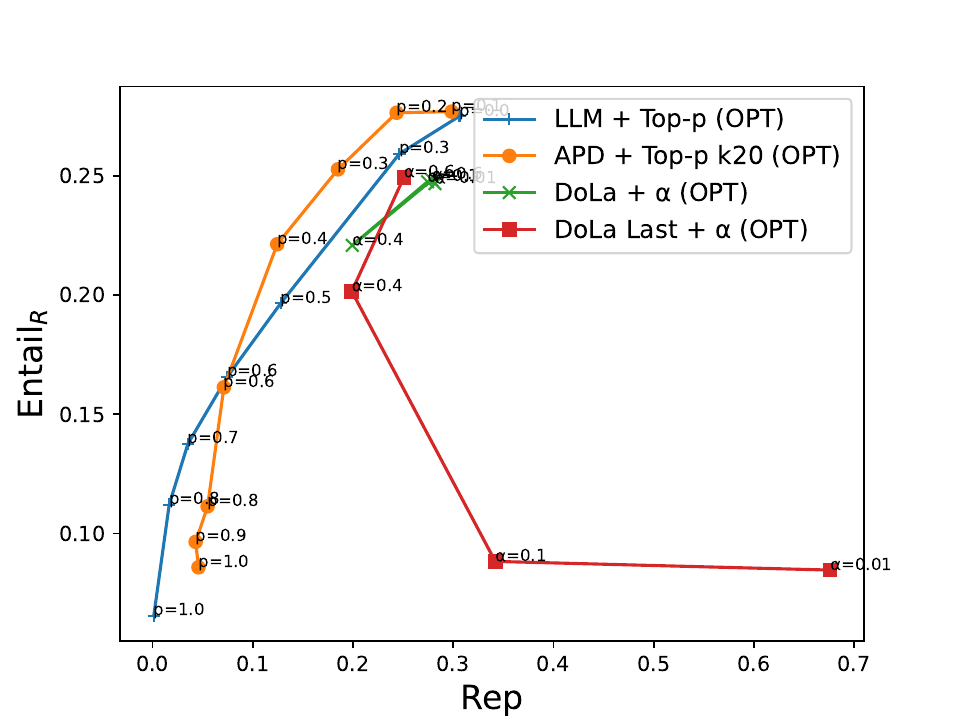}
  \caption{Layer Tuning of DoLa (OPT)}
\end{subfigure}
\begin{subfigure}{.33\textwidth}
  \centering
  \includegraphics[width=1\linewidth]{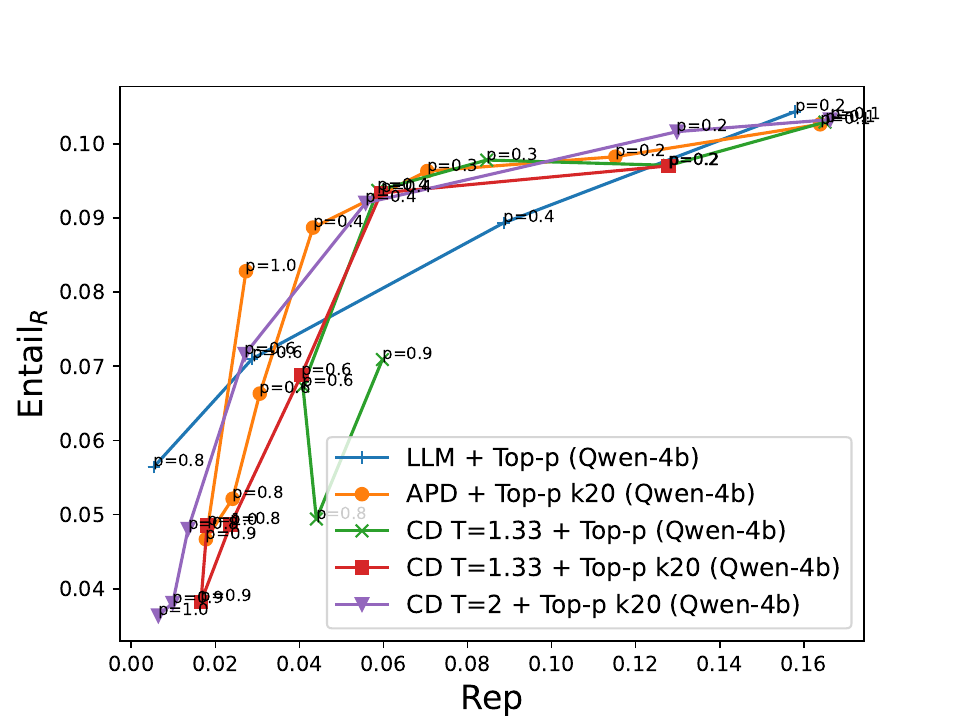}
  \caption{Ours v.s. CD (Qwen)}
\end{subfigure}%
\begin{subfigure}{.33\textwidth}
  \centering
  \includegraphics[width=1\linewidth]{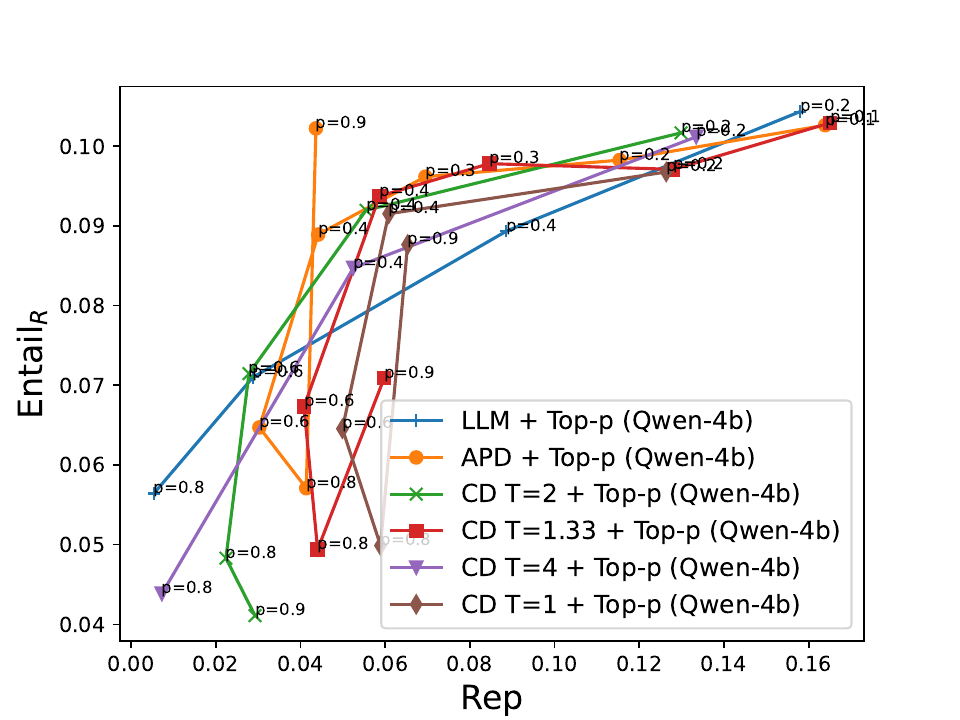}
  \caption{Temperature Tuning of CD (Qwen)}
\end{subfigure}
\caption{\textsc{FactualityPrompts} results using factual prompts. The curves closer to the upper left corner are better. The average standard error of Entail$_{R}$ is $0.0031$ and the maximal one is $0.0095$. }

\label{fig:comp_ent_rep_factual}
\end{figure*}

\begin{figure*}[t!]
\centering
\begin{subfigure}{.33\textwidth}
  \centering
  \includegraphics[width=1\linewidth]{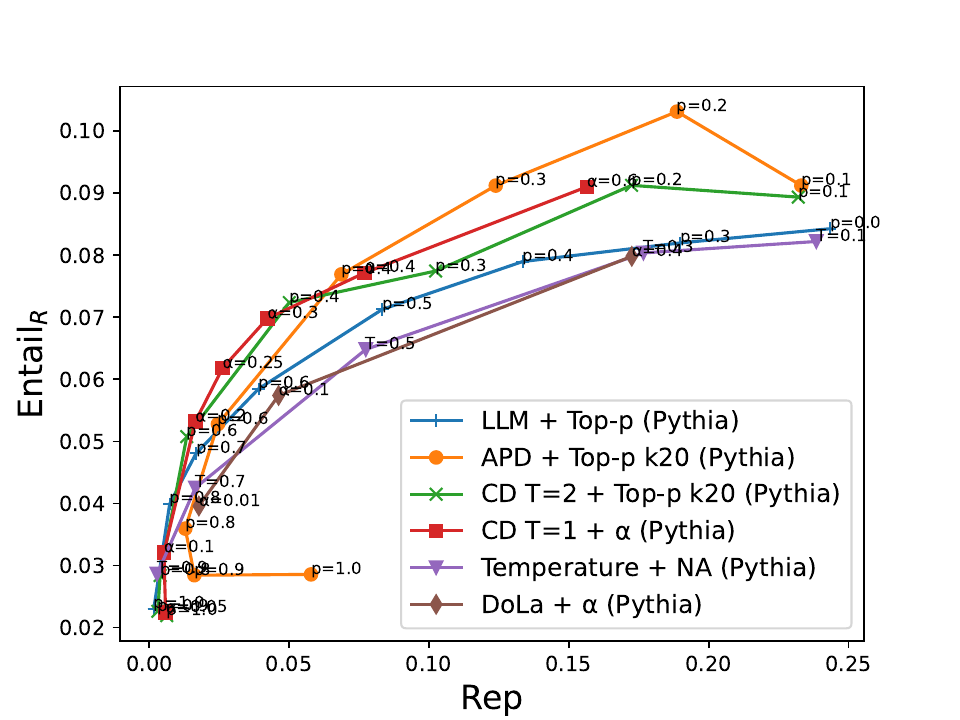}
  \caption{Ours v.s. SOTA (Pythia)}
\end{subfigure}%
\begin{subfigure}{.33\textwidth}
  \centering
  \includegraphics[width=1\linewidth]{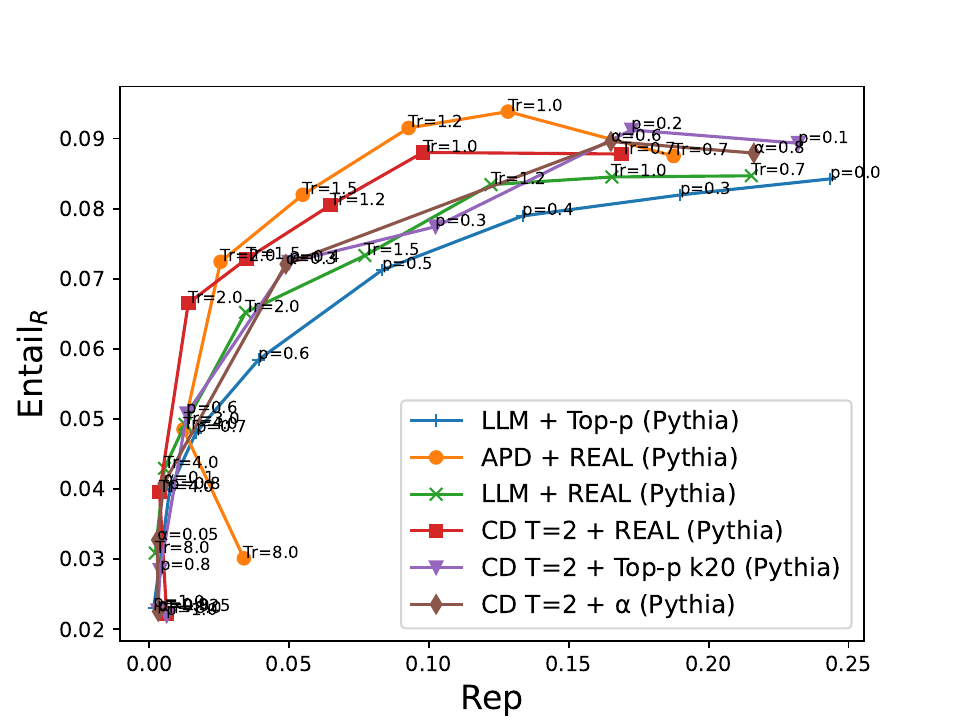}
  \caption{With REAL Sampling (Pythia)}
\end{subfigure}
\begin{subfigure}{.33\textwidth}
  \centering
  \includegraphics[width=1\linewidth]{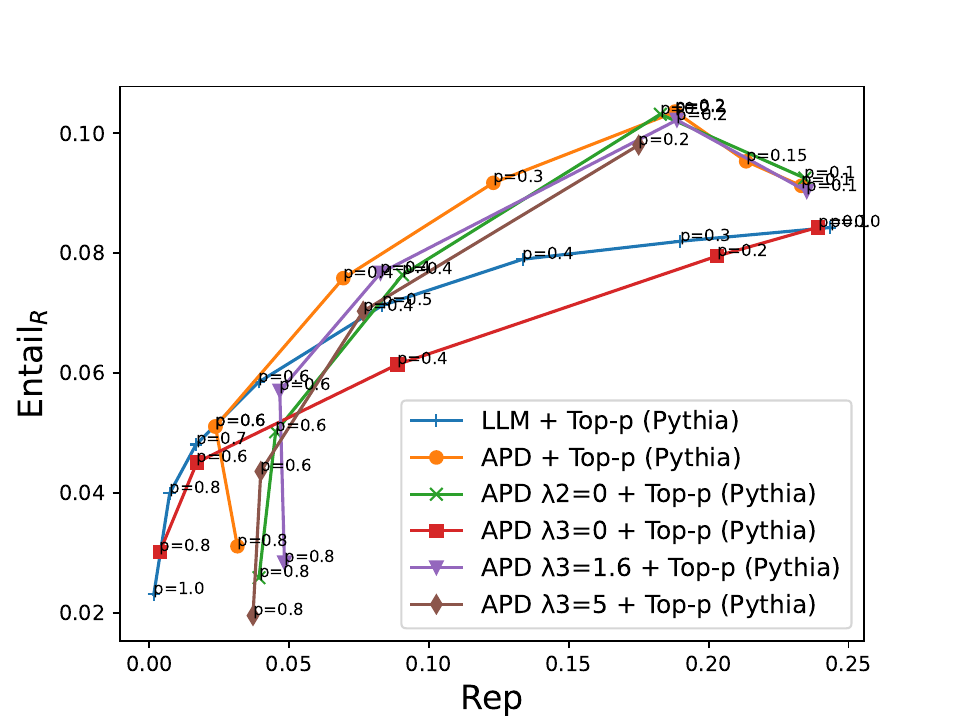}
  \caption{Loss Term Ablation (Pythia)}
\end{subfigure}%
\begin{subfigure}{.33\textwidth}
  \centering
  \includegraphics[width=1\linewidth]{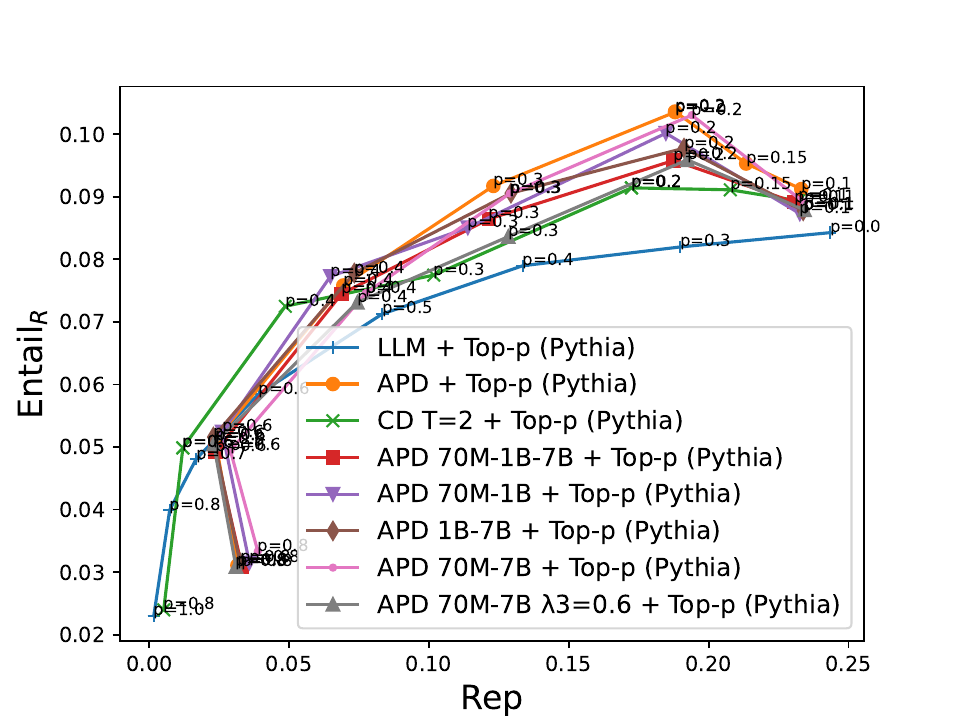}
  \caption{Mid-size LLM Ablation (Pythia)}
\end{subfigure}%
\begin{subfigure}{.33\textwidth}
  \centering
  \includegraphics[width=1\linewidth]{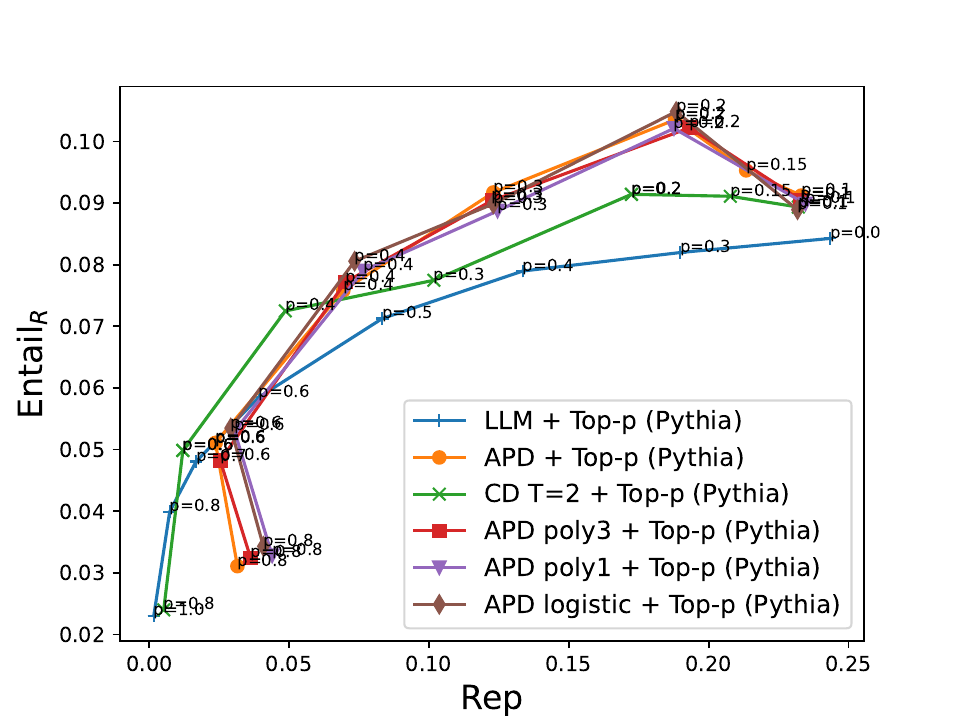}
  \caption{Function Ablation (Pythia)}
\end{subfigure}
\begin{subfigure}{.33\textwidth}
  \centering
  \includegraphics[width=1\linewidth]{figs/plot__nonfactual_rep_entail_abl_others.pdf}
  \caption{Other Ablation (Pythia)}
\end{subfigure}%
\begin{subfigure}{.33\textwidth}
  \centering
  \includegraphics[width=1\linewidth]{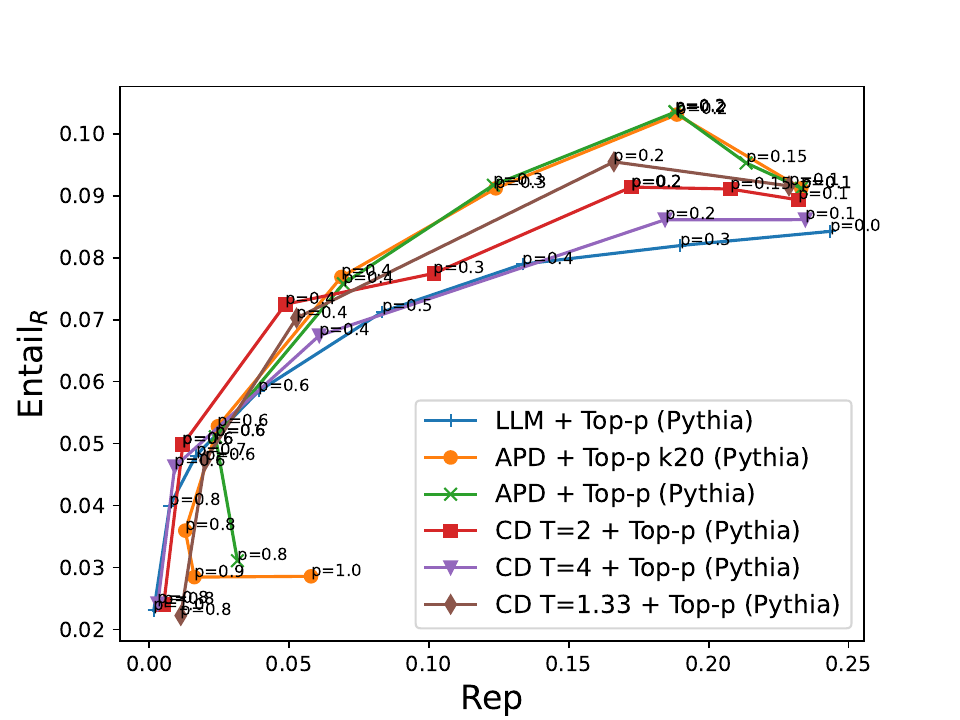}
  \caption{Temperature Tuning of CD (Pythia)}
\end{subfigure}%
\begin{subfigure}{.33\textwidth}
  \centering
  \includegraphics[width=1\linewidth]{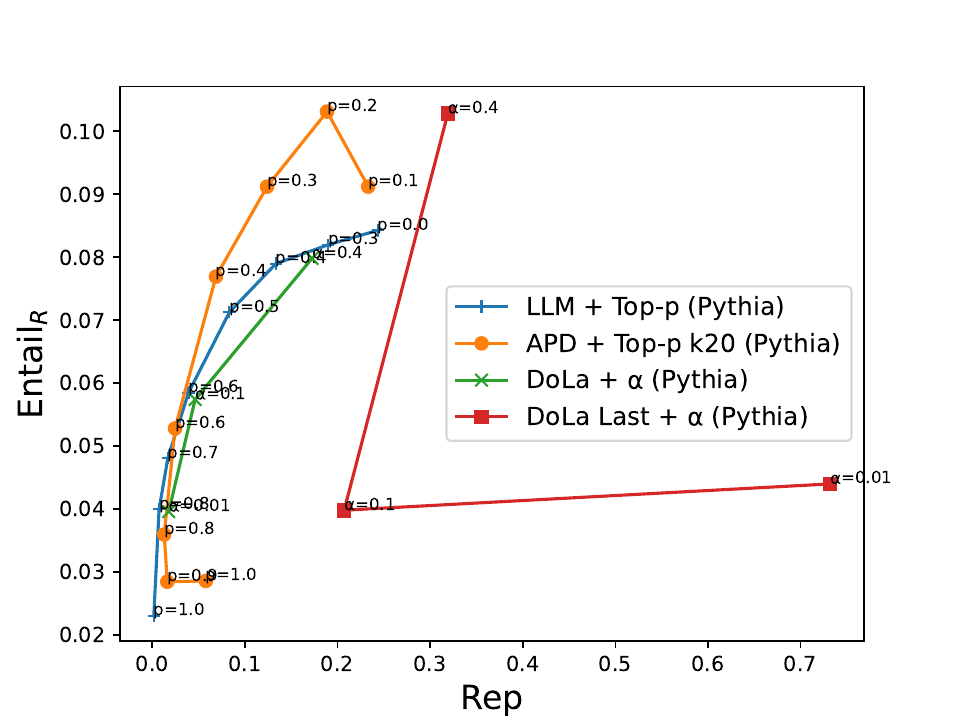}
  \caption{Layer Tuning of DoLa (Pythia)}
\end{subfigure}
\begin{subfigure}{.33\textwidth}
  \centering
  \includegraphics[width=1\linewidth]{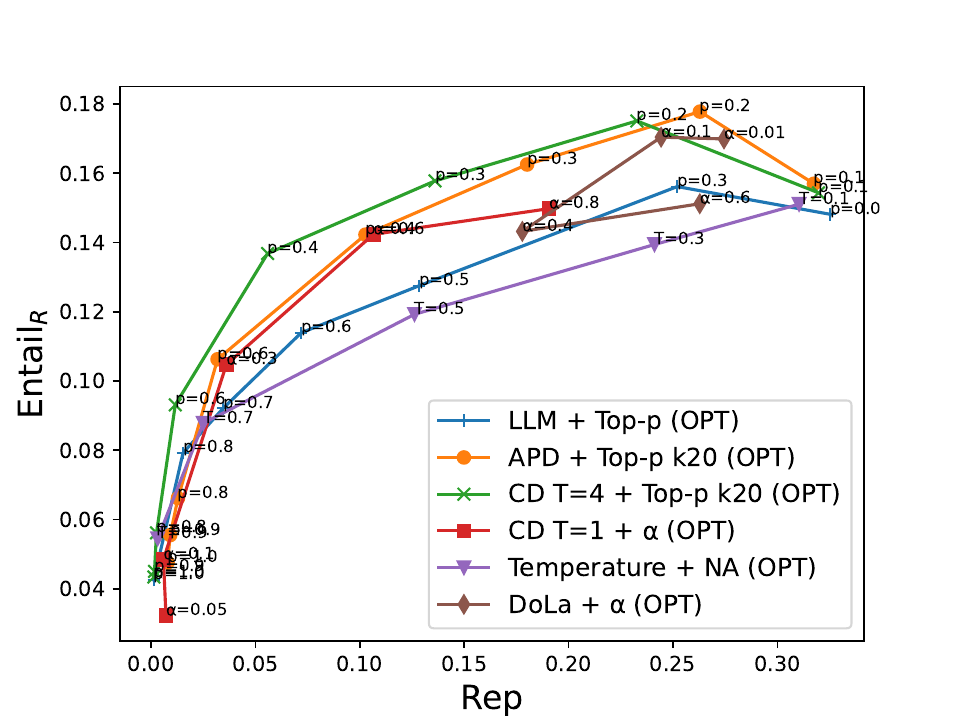}
  \caption{Ours v.s. SOTA (OPT)}
\end{subfigure}%
\begin{subfigure}{.33\textwidth}
  \centering
  \includegraphics[width=1\linewidth]{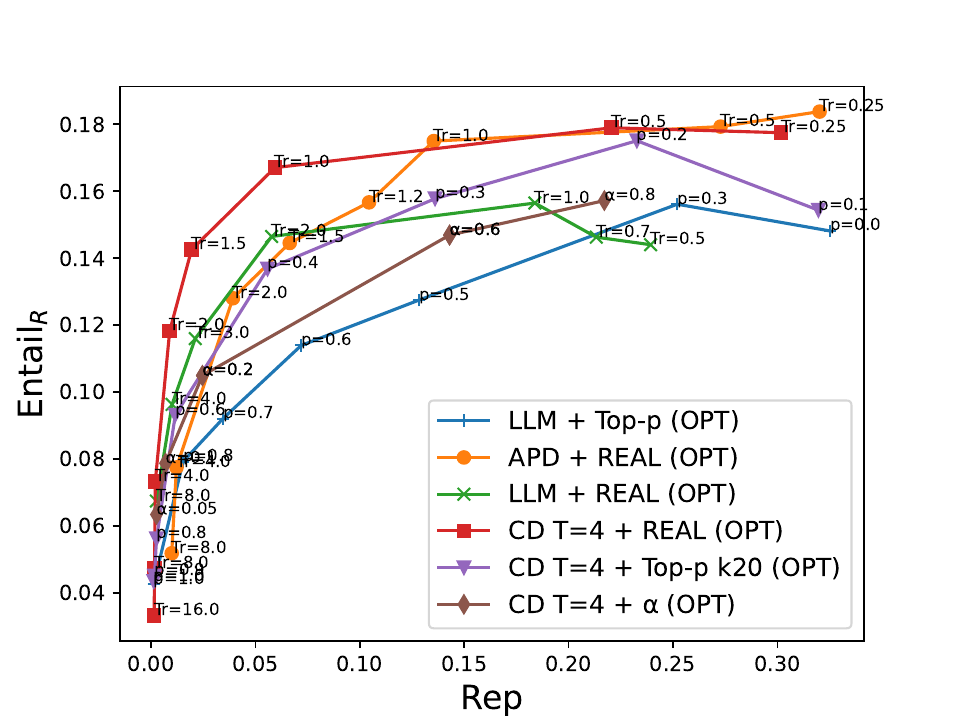}
  \caption{With REAL Sampling (OPT)}
\end{subfigure}
\begin{subfigure}{.33\textwidth}
  \centering
  \includegraphics[width=1\linewidth]{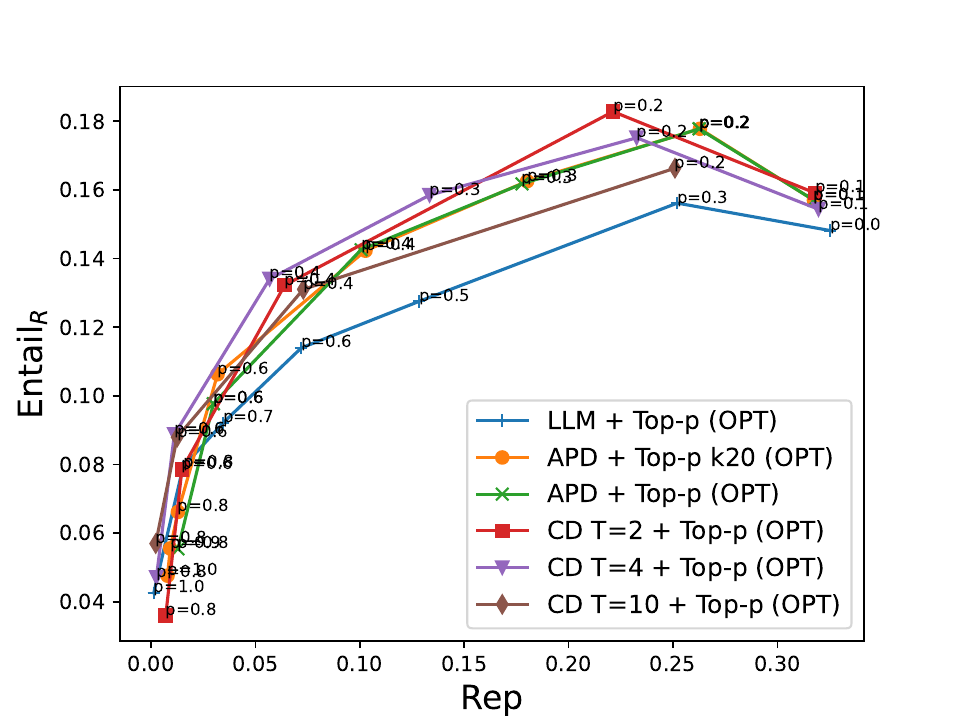}
  \caption{Temperature Tuning of CD (OPT)}
\end{subfigure}
\begin{subfigure}{.33\textwidth}
  \centering
  \includegraphics[width=1\linewidth]{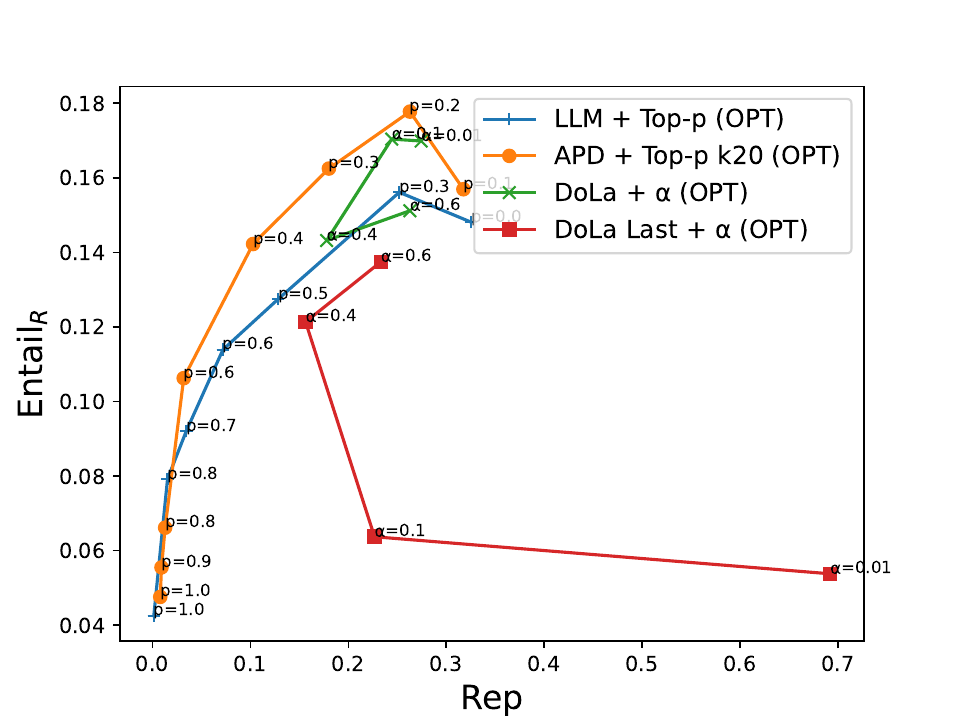}
  \caption{Layer Tuning of DoLa (OPT)}
\end{subfigure}
\begin{subfigure}{.33\textwidth}
  \centering
  \includegraphics[width=1\linewidth]{figs/plot__nonfactual_rep_entail_Qwen-4b.pdf}
  \caption{Ours v.s. CD (Qwen)}
\end{subfigure}%
\begin{subfigure}{.33\textwidth}
  \centering
  \includegraphics[width=1\linewidth]{figs/plot__nonfactual_rep_entail_Qwen-4b_T.pdf}
  \caption{Temperature Tuning of CD (Qwen)}
\end{subfigure}
\caption{\textsc{FactualityPrompts} results using nonfactual prompts. The curves closer to the upper left corner are better. The average standard error of Entail$_{R}$ is $0.0028$ and the maximal one is $0.0086$.}

\label{fig:comp_ent_rep_nonfactual}
\end{figure*}


\subsection{Asymptotic Probability Visualization}

We visualize the top 1 token probabilities of the Pythia family up to 6.9B in \Cref{fig:vis}. The input text is \textit{President of the United States Joe Biden}. We can see that probabilities of CD T=1 (red lines) are close to $0$ at every position because amateur LM could also predict the next token in this phrase correctly. If we reduce the influence of amateur LM by setting T=2 (green lines), the probability is still far away from the actual empirical probability curves (blue curves). Instead, the asymptotic probabilities (orange lines) from APD could often be closer to blue curves but sometimes farther away if APD believes the probability is steadily decreasing as the LM's size increases.

\subsection{More Results and Analyses from \textsc{FactualityPrompts}}

\textsc{FactualityPrompts} provides four main metrics: NE$_{ER}$, Entail$_{R}$, Dist-n, and Rep. \citet{REAL} combine NE$_{ER}$ and Entail$_{R}$ as Agg. Factuality and combines Dist-n and Rep as Agg. Diversity. We plot the Agg. Factuality versus Agg. Diversity in \Cref{fig:comp_agg_gen}, NE$_{ER}$ versus Dist-n in \Cref{fig:comp_ne_dist_factual} and \Cref{fig:comp_ne_dist_nonfactual}, and Entail$_{R}$ versus Rep in \Cref{fig:comp_ent_rep_factual} and \Cref{fig:comp_ent_rep_nonfactual}. We analyze the performances in \Cref{fig:comp_agg_gen} below. 

\expsec{Qwen}: 
\Cref{fig:qwen_cd} shows that \textbf{APD} is better than \textbf{CD} when $p \ge 0.4$. The improvements are smaller than Pythia and OPT probably because we can only fine-tune ALM' using Qwen-1.5 0.5B, 1.8B, and 4B\footnote{Qwen has a larger vocabulary size compared to Pythia and OPT, so 4B is the largest size we can run.}. The worse performances of \textbf{APD 1B-7B} in \Cref{fig:model_ablation} implies that smaller LM plays a crucial role in modeling the exponential decay and Qwen-1.5 does not have the LLM that is smaller than 0.5B. 


\expsec{Ablation Studies}:
The similar performance of $\lambda_3=1.6$ and the default value $\lambda_3=0.8$ in \Cref{fig:term_ablation} suggests that our performances are not very sensitive to the hyperparameters. \textbf{APD 70M-7B} improves \textbf{CD} by only using ALM and ELM probably because of a better inductive bias (i.e., exponential decay in the probability space is a better assumption than the linear decay in the logit space).

\expsec{DoLa}:
In the original DoLa paper~\citep{chuang2023dola}, they recommend to get the amateur logits from two subsets of layers: 0,2,4,6,8,10,12,14,32 or 16,18,20,22,24,26,28,30,32. We call the DoLa using the subset that contains mostly the first half of layers as \textbf{DoLa + $\alpha$} and the DoLa using the second subset as \textbf{DoLa Last + $\alpha$}. \Cref{fig:dola_temp} and \Cref{fig:opt_dola} show that selecting amateur logits from the lower layers is much better but still worse than top-$p$ sampling.

\expsec{Thresholding Methods}: 
\Cref{fig:topp}, \Cref{fig:cd}, \Cref{fig:opt}, and \Cref{fig:opt_real} indicate that from best to the worst, thresholding methods could be ranked as \textbf{REAL}, \textbf{top-$p$ $k$20}, \textbf{top-$p$}, and \textbf{$\alpha$} according to their performances in this benchmark. Hence, \textbf{APD + REAL} achieves the best results and the new state-of-the-art results according to~\citet{REAL}.

\subsection{Decay Parameterization Comparison}
In \Cref{eq:exp}, we parameterize the decay curves using an exponential function. Following \citet{REAL}, we also evaluate the logistic decay function ($\hat{p}_{w,c}(s) = \hat{P'}^{AP}_{c}(w) + \frac{a_{w,c}}{1+\exp( \max(0, b_{w,c} (s - d_{w,c}) ) )}$) and factional polynomial \citep{chang2020using} ($\hat{p}_{w,c}(s) = \hat{P'}^{AP}_{c}(w) + a_{w,c} ( \frac{d_{w,c,0.5}}{ x_c(s)^{0.5} } + \sum_{k=1}^{K} \frac{d_{w,c,k}}{ x_c(s)^{k} }  )$ and $x_{w,c}(s) = \max(1, b_{w,c} (s - d_{w,c}))$) whose parameters are all predicted by MLP. 

In \Cref{fig:func_ablation}, we report the APD using the logistic decay function (\textbf{ADP logistic}), APD using factional polynomial with $K=1$ (\textbf{ADP poly1}) and $K=3$ (\textbf{ADP poly3}). The results show that \textbf{ADP logistic}, \textbf{ADP poly1}, and \textbf{ADP} perform similarly well while \textbf{ADP poly3} performs worse. This might indicate that the probability decay curves are noisy and using fewer parameters could mitigate this problem. The worse performances of \textbf{APD on the fly} in the QA experiments also support this hypothesis.

\begin{table*}[t!]
\scalebox{0.77}{
\begin{tabular}{cl|c|cc|cccc|cc|cc}
\multicolumn{1}{l}{}    &                 & LAMBADA        & \multicolumn{2}{c|}{CQA}         & \multicolumn{4}{c|}{QASC}                                          & \multicolumn{2}{c|}{ARC}         & \multicolumn{2}{c}{SocialIQA}   \\
\multicolumn{1}{l}{}    &                 &                &                &                & \multicolumn{2}{c}{Q+Fact}      & \multicolumn{2}{c|}{Q Only}           &                &                &                &                \\
\multicolumn{1}{l}{}    &                 & ppl ($\downarrow$)            & ppl ($\downarrow$)            & acc            & ppl ($\downarrow$)            & acc            & ppl ($\downarrow$)            & acc            & ppl ($\downarrow$)            & acc            & ppl ($\downarrow$)            & acc            \\ \hline

\multirow{5}{*}{OPT} & LLM 6.7B  & \textbf{2.388} & 9.117          & 0.664          & 4.994          & 0.851          & 8.094          & 0.604          & 4.739          & 0.660          & 7.259          & 0.665          \\
 & CD  & 2.412          & \textbf{7.227} & \textbf{0.707} & 5.039          & 0.856          & 7.925          & 0.622 & 4.736          & \textbf{0.679} & \textbf{6.673} & \textbf{0.712} \\
 & APD & 2.396          & 7.335          & 0.695          & \textbf{4.985} & \textbf{0.860} & \textbf{7.911} & 0.622 & \textbf{4.701} & 0.676          & 6.691          & 0.690         \\ 
& APD on the fly & 2.414 & 8.846 & 0.664 & 5.040 & \textbf{0.860} & 8.090 & \textbf{0.630} & 4.765 & 0.678 & 7.248 & 0.705 \\ \cline{2-13}
 & LLM 13B & 2.295          & 9.308          & 0.660          & 6.458          & 0.822          & 8.319          & 0.634          & 4.757          & 0.678          & 9.139          & 0.636   \\ \hline \hline
\multirow{5}{*}{Qwen1.5} & LLM 4B & \textbf{2.430} & 5.797          & 0.671          & 3.618          & 0.888 & 6.613          & 0.672          & \textbf{4.210} & 0.716          & 7.078          & 0.590          \\
 & CD & 2.474          & 5.890          & 0.675 & 3.648 & 0.888 & 6.721          & 0.674          & 4.279 & \textbf{0.726} & 7.150          & 0.636          \\
 & APD & 2.445          & \textbf{5.535} & 0.674          & \textbf{3.596}          & 0.886          & \textbf{6.571} & \textbf{0.686} & 4.237          & 0.720          & \textbf{7.059} & \textbf{0.643} \\
 & APD on the fly & 2.459 & 5.840 & \textbf{0.676} & 3.657 & \textbf{0.889} & 6.670 & 0.671 & 4.258 & 0.725 & 7.131 & 0.613 \\ \cline{2-13}
 & LLM 7B & 2.186          & 5.899          & 0.696          & 4.083          & 0.911          & 7.306          & 0.698          & 3.977          & 0.778          & 6.776          & 0.645   \\ \hline \hline

\end{tabular}
}
\caption{Perplexity and accuracy comparison of one-shot QA using different decoding methods.  }
\label{tb:QA_ppl_opt}
\end{table*}

\begin{table}[t!]
\scalebox{0.85}{
\begin{tabular}{cl|c|cc}
    &          & MultiRC        & \multicolumn{2}{c}{SQuAD}          \\
    &          &                & Q+P            & Q              \\ \hline \hline

 \multirow{5}{*}{Pythia} & LLM 6.9B  & 3.620          & 2.360          & 5.426          \\
 & CD & 3.377          & 2.246          & 5.029          \\
 & APD & \textbf{3.349} & \textbf{2.231} & \textbf{5.018} \\
 & APD on the fly &   3.623 & 2.381 & 5.432  \\ \cline{2-5}
 & LLM 12B & 3.336          & 2.289          & 4.978      \\ \hline \hline   
 \multirow{5}{*}{OPT}    & LLM 6.7B  & 3.900          & 2.450          & 5.654          \\
 & CD & \textbf{3.775} & \textbf{2.381} & \textbf{5.403} \\
 & APD & 3.840 & 2.391 & 5.437          \\ 
 & APD on the fly & 3.908 & 2.469 & 5.657          \\ \cline{2-5}
 & LLM 13B & 3.741 & 2.310 & 5.268  \\ \hline \hline
 \multirow{5}{*}{Qwen1.5}  & LLM 4B & 4.023          & 1.853          & \textbf{5.038} \\
 & CD & \textbf{3.985} & \textbf{1.841} & 5.152          \\
 & APD & 4.057          & 1.847          & 5.098          \\
 & APD on the fly & 4.033 & 1.872 & 5.106          \\ \cline{2-5}
 & LLM 7B & 3.171          & 1.557          & 3.242  \\ \hline \hline

\end{tabular}
}
\caption{Perplexity of MRC tasks. Smaller numbers are better.}
\label{tb:MRC_both}
\end{table}

\begin{table*}[t!]
\scalebox{0.8}{
\begin{tabular}{clccccccccc}
 &                                         & LAMBADA        & CQA & \multicolumn{2}{c}{QASC}  & ARC            & SocialIQA      & MultiRC  & \multicolumn{2}{c}{SQuAD}     \\
&                                         & &  & Q+Facts & Q only  &            &       &  & Q+P    & Q only \\ \hline \hline

 \multirow{6}{*}{Pythia} & LLM 6.9B  & 0.841          & 0.483          & 0.564          & 0.491          & 0.660          & 0.475          & 0.696          & 0.820          & 0.591          \\
 & CD & 0.861          & 0.572          & 0.564          & \textbf{0.502} & 0.671          & 0.512          & \textbf{0.715} & \textbf{0.836} & \textbf{0.611} \\
 & APD & \textbf{0.867} & \textbf{0.582} & \textbf{0.612} & 0.501          & \textbf{0.673} & \textbf{0.518} & \textbf{0.715} & 0.835          & \textbf{0.611} \\
 & APD on the fly & 0.845 & 0.496 & 0.564 & 0.494 & 0.660 & 0.484 & 0.700 & 0.822 & 0.594         \\ \cline{2-11}
 & LLM 12B & 0.853          & 0.494          & 0.629          & 0.514          & 0.684          & 0.493          & 0.708          & 0.825          & 0.618          \\
 & |D| & 5000           & 3343           & 6040           & 2181           & 1611           & 4783           & 3798           & 28890          & 10491    \\ \hline \hline
\multirow{6}{*}{OPT}    & LLM 6.7B & 0.831          & 0.451          & 0.601          & 0.490          & 0.644          & 0.517          & 0.682          & 0.816          & 0.575          \\
                        & CD       & \textbf{0.833} & \textbf{0.534} & 0.599          & \textbf{0.499} & 0.648          & \textbf{0.559} & \textbf{0.688} & \textbf{0.831} & \textbf{0.594} \\
                        & APD   & 0.831          & 0.521          & \textbf{0.605} & 0.498          & \textbf{0.651} & 0.549          & 0.685          & 0.824          & 0.588          \\
                        & APD on the fly  & 0.832 & 0.470 & 0.601 & 0.494 & 0.648 & 0.527 & 0.685 & 0.819 & 0.582
                        \\ \cline{2-11}
                        & LLM 13B  & 0.841          & 0.443          & 0.525          & 0.476          & 0.636          & 0.432          & 0.692          & 0.826          & 0.601          \\
                        & |D|      & 4985           & 2380           & 5827           & 1942           & 1477           & 3843           & 3401           & 28330          & 8763          \\ \hline \hline

 \multirow{6}{*}{Qwen1.5}    & LLM 4B  & 0.825          & 0.591          & 0.705          & 0.549          & 0.655          & 0.533        & \textbf{0.685}  & 0.860          & \textbf{0.611}  \\
 & CD & 0.825          & 0.590          & 0.703          & 0.546          & 0.652          & 0.530  & \textbf{0.685}        & \textbf{0.866} & 0.606           \\
 & APD & \textbf{0.832} & \textbf{0.614} & \textbf{0.709} & \textbf{0.553} & \textbf{0.657} & \textbf{0.534}  & 0.682 &  0.862          & 0.608  \\ 
 & APD on the fly & 0.826 &0.591 & 0.705 & 0.549 & 0.656 & 0.532 & 0.684 & 0.860 & 0.610 \\ \cline{2-11}
 & LLM 7B & 0.851          & 0.586          & 0.681          & 0.520          & 0.678          & 0.546        & 0.742  & 0.909          & 0.726                    \\
 & |D| & 4925           & 5079           & 6718           & 2676           & 1856           & 8290           & 5071           & 31247          & 11690    \\ \hline \hline

\end{tabular}
}
\caption{MRR Comparison for 7 QA datasets. |D| is the dataset size after filtering out the answers whose tokens are ranked below the top 20 by ELM.}
\label{tb:QA_mrr_both}
\end{table*}

\subsection{More Results from QA Datasets}
\label{sec:qa_more}
In \Cref{tb:QA_ppl_opt}, we surprisingly find that the perplexity of \textbf{LLM 13B} is worse than the perplexity of \textbf{LLM 7B} at CQA, QASC, ARC, and SocialIQA. This suggests that some training detail differences (e.g., batch sizes or training data order) between OPT 7B and 13B affect their final performances. Since a larger model does not necessarily perform better, it is not surprising that \textbf{APD} sometimes do worse than \textbf{CD} or \textbf{LLM 7B}. For Qwen-1.5, the results are also mixed. \textbf{CD} and \textbf{APD} often perform similarly compared to \textbf{LLM 4B} probably because the size difference between ALM and ELM is too small (0.5B vs 4B). 

Besides commonsense QA datasets, we also compare \textbf{APD} and \textbf{CD} using 2 reading comprehension datasets: MultiRC~\citep{MultiRC2018} and SQuAD~\citep{rajpurkar2016squad}. In SQuAD, we test the prompts with passage (Q+P) and without passage (Q only). \Cref{tb:MRC_both} shows that in Pythia, \textbf{APD} is slightly better than \textbf{CD}, but \textbf{CD} is often slightly better than \textbf{APD} in OPT and Qwen. 

\Cref{tb:QA_mrr_both} reports the MRR (Mean Reciprocal Rank) results, which only consider the rank of the correct answer. The results are similar to perplexity and accuracy. \textbf{APD} is generally better than \textbf{CD} for Pythia and Qwen, while the results are mixed for OPT.

Finally, compared to ELM \textbf{LLM 6.9B/7B/4B}, \textbf{APD on the fly} does worse in perplexity but sometimes better in MRR and accuracy. This seems to suggest that the estimated asymptotic probabilities could improve the rank of tokens while their absolute values are degraded by the noise in the empirical probability decay curve. Fine-tuning a small ALM' and the regularization term in \Cref{eq:loss3} should be able to alleviate the problem by reducing the noise.

\begin{table*}[t!]
\scalebox{0.75}{
\begin{tabular}{c|cccc|cccc}
        & \multicolumn{4}{c|}{Human Experiments (150  continuations)}        & \multicolumn{4}{c}{Automatic Metrics (28k continuations) (\%)}          \\
                & Factuality     & Informativeness & Fluency & Overall        & Dist-2          & Rep ($\downarrow$)            & NE$_{ER}$ ($\downarrow$)          & Entail$_R$       \\ \hline
LLM + Top-$p$ ($p=1.0$)    &  1.82 & 3.67 & 3.88 &  2.41 & 40.0 &	0.2 &	47.9 &	3.2  \\
CD T=2 + Top-$p$ k20 ($p=0.8$)  & 2.00 & \textbf{4.05} & \textbf{4.12} &  \textbf{2.58} & \textbf{48.3} &	\textbf{0.5} &	40.3 &	4.2 \\
APD + Top-$p$ k20 ($p=0.8$)  & \textbf{2.07} & 3.97 & 4.11 &  \textbf{2.58} & 47.1 &	1.6 &	\textbf{39.2} &	\textbf{5.3} \\ \hline
CD T=2 + REAL ($T_r=4.0$)  & 2.16 & \textbf{4.04} & \textbf{4.18} & 2.62 & \textbf{43.3} &	\textbf{0.5} &	39.3 &	5.9 \\
APD + REAL ($T_r=4.0$)  & \textbf{2.31} & 4.02 & 4.14 & \textbf{2.73} & 42.0 &	1.6 &	\textbf{37.5} &	\textbf{7.0} \\ \hline
\end{tabular}
}
\caption{Evaluation the generated responses in \textsc{FactualityPrompts} using MTurk workers and automatic metrics. The questionaires for humans use 1 to 5 Likert scale.}
\label{tb:human_abs_scores}
\end{table*}

\section{More Experiments}
\label{sec:more_exp}

To make our results more complete, we also test APD using human evaluation and story-writing tasks.

\subsection{Human Evaluation for \textsc{FactualityPrompts}}
\label{sec:human_eval}

We adopt the Amazon Mechanical Turk (MTurk) tasks from~\citet{REAL}, which ask the workers to evaluate the factuality of the generated continuation in \textsc{FactualityPrompts} using search engines. The workers also need to label the informativeness, fluency, and overall. We collect the annotations for 150 prompts and each task is labeled by 2 workers. The Pearson correlation between the two workers for factuality, informativeness, fluency, and overall are 38.8\%, 33.0\%, 31.6\%, and 33.5\%, respectively.  

First, \Cref{tb:human_abs_scores} demonstrates that \textbf{APD} indeed improves the factuality of the continuations, which makes \textbf{APD} particularly suitable for high-stake domains such as medicine and laws. \textbf{CD T=2 + Top-$p$ k20} has a higher informativeness score compared to \textbf{APD + Top-$p$ k20} probably because \textbf{CD} tends to ignore the most obvious next token as shown in \Cref{fig:first_fig}. Overall, \textbf{APD + REAL} achieves the best performance and improves the factuality of the top-p sampling by $27\%$.

Notice that the current results might not be very stable because the sampling process involves lots of randomnesses, annotators might disagree with each other, and we can only afford 150 continuations, whose annotations cost us 875 dollars. Nevertheless, we can see that the methods that achieve higher factual scores in automatic evaluation are indeed more factual from humans' perspective.



\begin{table*}[t!]
\scalebox{0.9}{
\begin{tabular}{cc|cccc|cccc}
  &  & \multicolumn{4}{|c|}{GPT3.5 Turbo Evaluation (500 continuations)} & \multicolumn{4}{|c}{Automatic Evaluation (8k continuations)} \\
     &  & Fluency        & Coherency      & Likability     & Overall        & Dist-2         & MAUVE          & ROUGE-2        & Sim            \\ \hline \hline
 \multirow{2}{*}{Pythia}    & CD T=2   & \textbf{0.563} & \textbf{0.575} & \textbf{0.575} & \textbf{0.573} & 0.364          & \textbf{0.048} & 2.595          & \textbf{0.513} \\
                            & APD    & 0.526          & 0.546          & 0.548          & 0.546          & \textbf{0.384} & 0.044          & \textbf{2.634} & 0.510          \\ 
                           \hline \hline
 \multirow{2}{*}{OPT}      & CD T=4    & 0.520          & 0.522          & 0.524          & 0.526          & 0.301          & \textbf{0.485} & \textbf{3.115} & \textbf{0.526}          \\
                           &  APD    & \textbf{0.546} & \textbf{0.538} & \textbf{0.540} & \textbf{0.546} & \textbf{0.337} & 0.442          & 2.919          & \textbf{0.526} \\ \hline \hline
\end{tabular}
}
\caption{Short Story Writing Experiment. Every method uses top-p k20 ($p=0.8$). We report the winning rate against top-$p$ sampling ($p=0.95$) that is judged by GPT3.5 Turbo.}
\label{tb:creative_writing}
\end{table*}

\subsection{Story Writing Experiments}

In \Cref{fig:first_fig}, we show that CD tends to ignore the most obvious next token. Although this behavior could degrade the factuality as shown in the main paper, it might not be an issue in story writing because the nonfactuality of a story is often not an issue and the surprising next tokens might make the stories more interesting and likable. Nevertheless, for completeness, we compare CD with APD in terms of completing ROC stories~\citep{mostafazadeh2016corpus}, which are often used in the story generation literature~\citep{yao2019plan}. 

We use the same experiment setup and metrics of \citet{REAL}. A prompt includes 3 shot examples and the first two sentences of the final ROC story. Each decoding method generates 8 different completions of the final story. Finally, the generated completions are first compared with top-$p$ sampling ($p=0.95$) using GPT3.5 Turbo Evaluation and then compared with human reference story using MAUVE~\citep{pillutla2021mauve}, ROUGE-2 F1~\citep{lin-2004-rouge}, and similarity (Sim) from sBERT~\citep{reimers-gurevych-2019-sentence}, all-mpnet-base-v2. All the metrics use their default hyperparameters. We use the best hyperparameters of CD and APD from the \textsc{FactualityPrompts} experiments.

The results in \Cref{tb:creative_writing} indicate that APD performs comparably to CD. It is worth mentioning that our ALM' is trained only on a small subset of Wikipedia, so this experiment confirms a certain degree of generalization ability of APD.






\section{Method Details}
\label{sec:method_details}

By default, the ALM' is trained by 70m, 160m, 410m, 1B, 1.4B, 2.8B, and 6.9B de-duplicated Pythia. For OPT, the ALM' is trained by 125m, 350m, 1.3B, 2.7B, and 6.7B LMs. For Qwen-1.5, the ALM' is trained by only 3 LMs (0.5B, 1.8B, and 4B). During training, we construct the set $A_c$ by subsampling the next tokens according to their probabilities of ELM. Besides the $20$ tokens with the highest probabilities, we randomly sample $5$ tokens between top $20$ and top $100$ tokens. The smaller $N$ in OPT and Qwen allow us to sample a little bit more tokens ($N=7$ in Pythia, $N=5$ in OPT, and $N=3$ in Qwen-1.5). Hence, after the top $100$, we sample $5$ tokens for Pythia and $10$ tokens for OPT. 

From \Cref{fig:cd_temp} and \Cref{fig:opt_cd_temp}, we can see that \textbf{APD + Top-$p$ k20} is slightly better than \textbf{APD + Top-$p$} and we observe that the improvement gap increases a lot when $p=0.9$ in a preliminary study. This is because the probabilities of the top 20 tokens are always used to train ALM'. It seems to suggest that more training data could improve the performance and stability of ALM'. 

Before being inputted into the MLP, $\{p(w|c,\theta_{s_i})\}_{i=1}^N$ and $\hat{P}^{AP}_c(w)$ are normalized such that $\sum_{w \in A_c} p(w|c,\theta_{s_i}) = 1 \; \forall i$ and $\sum_{w \in A_c} \hat{P}^{AP}_c(w) = 1$. We choose to model the normalized probabilities because computing the unnormalized probabilities requires all the logits of ELM and storing the logits of all the tokens is not feasible to us.

When checking the asymptotic probability prediction, we use a four-layer MLP with hidden state size 100. We first use a dropout layer that has $0.5$ probability to mask the probabilities from $\theta_3$ to $\theta_{N-1}$. Between layers, we insert the nonlinear layer  GELU~\citep{hendrycks2016gaussian}. The final layer projects the hidden state to $3$ values and we take the exponential of these $3$ values as $a_{w,c}, b_{w,c}, d_{w,c}$ to ensure their positivity. Before the training starts, we initialize the weights and bias of the final layer to be $0$ to prevent the exponential layer output too large values.

Our ALM' is trained for 5 epochs using learning rate 1e-4 and AdamW~\citep{loshchilov2017decoupled}. The batch size is set to 64 and the warmup step is 100. We plan to release code to reveal more method details.














\section{Experiment Details}
\label{sec:exp_details}

We run all of our experiments using 5 servers. Each server has 8 32G V100. To reduce the hyperparameters in our experiments, we keep the ELM's temperature to be $1$ as in ~\citet{li2022contrastive} and input the full context to the amateur LM. 




\subsection{\textsc{FactualityPrompts} Experiment Details}

We choose to evaluate APD using \textsc{FactualityPrompts} because of the high quality of its automatic factuality metrics. The human studies in \citet{lee2022factuality} demonstrate that the retrieval-based metrics, NE$_{ER}$ and Entail$_{R}$, have strong correlations ($-0.8$) with the factuality judgments from human experts. 
Our experiment settings are the same as the setting of \citet{REAL} to make the results comparable. We use the validation/testing split from~\citet{REAL}, where there are 2k prompts in the validation set and 14k prompts in the testing set. 

We tune the hyperparameter $\lambda_3$ from the option set \{0.2,0.4,0.6,0.8,1.0,1.2\} using the validation set. The REAL sampling model for OPT is trained using the Pythia LLM family as in \citep{REAL}. The results of \textbf{LLM + Top-p}, \textbf{Temperature + NA}, \textbf{DoLa + $\alpha$}, and \textbf{CD T=1 + $\alpha$} are copied from \citep{REAL}. 

\subsection{APD on the Fly Baseline}
\label{sec:APD_no_FT}

The \textbf{APD on the fly} baseline simplifies the proposed \textbf{APD} by conducting extrapolation and estimating the asymptotic probabilities during the inference time. While we skip the step of fine-tuning ALM', the baseline is much more time-consuming because it needs to run the inference of all $N$ LMs to get $\{p(w|c,\theta_{s_i})\}_{i=1}^N$ during the testing time. 

The baseline tries to use the same loss function of \textbf{APD} except that the regularization term is not applicable because it does not fine-tune ALM'. It first reverses the increasing $\{p(w|c,\theta_{s_i})\}_{i=1}^N$ using \Cref{eq:flip} and estimates the asymptotic probability (AP), $a_{w,c}$, $b_{w,c}$, and $d_{w,c}$ in \Cref{eq:exp} using a simple gradient descent. We set $\lambda_2=10$ and $\lambda_3=0$ in \Cref{eq:loss_all}. The optimization is done by Adam~\citep{kingma2014adam} using 400 iterations and the initial learning rate 1e-2. Every parameter is clamped to 0 if its value is negative. If we encounter nan after the optimization, we divide the learning rate by 2 and redo the optimization until the nan problem is solved.

After the optimization, we flip back the asymptotic probability if necessary, normalize the probability of the top 20 tokens as $p^{ac}_{w,c}$ and output $(1-1/T) p(w|c,\theta_{s_N}) + 1/T \cdot p^{ac}_{w,c}$ using the best global weight $1/T$ in \{0.05, 0.1, 0.15, 0.2, 0.25, 0.3, 0.35, 0.4, 0.45, 0.5, 0.55, 0.6, 0.8, 1.0\}.





\subsection{QA Experiment Details}

Our method is unsupervised, so we can choose any subset of the datasets. We choose to use SQuAD validation set and the training set of all the other datasets because training datasets are usually larger and the validation set of SQuAD is large enough. To simplify our experiments, we compute the performances of CD and APD using different $1/T$ values and directly report the best performance. $1/T$ is chosen from the following options: \{0.05, 0.1, 0.15, 0.2, 0.25, 0.3, 0.35, 0.4, 0.45, 0.5, 0.55, 0.6, 0.8, 1.0, 1.2, 1.4, 1.6, 1.8, 2.0\}.

Before computing the perplexity, we renormalize the probabilities of the top $20$ tokens for all the methods. All probabilities are added by $0.01$ to prevent few extreme low probabilities influence the probability too much. When measuring accuracy, we make sure that all tokens of the correct answer and all tokens of at least one incorrect answer are within the top 20 token of ELM, and filter out the questions that do not have such the answers. After filtering, the sizes of CommonsenseQA, QASC (Q+Fact), QASC (Q only), ARC, and SocialIQA are $911$, $1138$, $956$, $899$, and $879$ for Pythia, and $491$, $982$, $845$, $780$, and $549$ for OPT.

In \Cref{tb:QA_ppl_pythia}, we use two-sample t-test on the negative log likelihoods of every answer to check the statistical significance. To run LLM 12/13B, we reduce the weight precision to fp16 during inference time in the QA experiments.

\subsection{One-shot Prompt Templates}

Template for ARC, CommonsenseQA, and QASC (Q only):
\begin{myprop}
 Question: Which kind of animals can fly? \\
 Answer: bird. \\ \\
 Question: \textbf{\{Question\}} \\
 Answer:
\end{myprop}

\noindent Template for QASC (Q+Facts):
\begin{myprop}
 Question: Which kind of animals can fly?\\
 Fact 1: a bird is an animal.\\
 Fact 2: birds can fly.\\
 Answer: bird.\\\\
 Question: \textbf{\{Question\}} \\
 Fact 1: \textbf{\{Fact 1\}}\\
 Fact 2: \textbf{\{Fact 2\}}\\
 Answer: 
\end{myprop}

\noindent Template for SocialIQA:
\begin{myprop}
 Passage: John likes to go hiking, and his wife likes to cook.\\ 
 Question: Who likes to cook?\\ 
 Answer: his wife\\ \\
 Passage: \textbf{\{Passage\}}\\
 Question: \textbf{\{Question\}}\\
 Answer:
\end{myprop}

\noindent Template for MultiRC:
\begin{myprop}
 Passage: Sent 1: John likes to go hiking, and his wife likes to cook.\\ 
 Sent 2: His wife likes to cook.\\ 
 Here is a question: Who likes to cook?\\ 
 The answer is his wife\\ \\
 Passage: \textbf{\{Passage\}}\\
 Here is a question: \textbf{\{Question\}}\\
 The answer is
\end{myprop}

\noindent Template for SQuAD (Q+P):
\begin{myprop}
 Passage: John likes to go hiking, and his wife likes to cook.\\ 
 Question: Who likes to cook?\\ 
 The answer is his wife\\ \\
 Passage: \textbf{\{Passage\}}\\
 Question: \textbf{\{Question\}}\\
 The answer is
\end{myprop}

\noindent Template for SQuAD (Q only):
\begin{myprop}
 Here is a question: What is the birthplace of Barack Obama?\\ 
 The answer is Honolulu, Hawaii. \\ \\
 Here is a question: \textbf{\{Question\}}\\
 The answer is
\end{myprop}






\end{document}